\documentclass{article} % For LaTeX2e
\usepackage{iclr2025_conference,times}

\usepackage{soul}
\usepackage{hyperref}
\usepackage{url}
\usepackage{microtype}
\usepackage{times}
\usepackage{CJKutf8}
\usepackage{xcolor}
\usepackage{latexsym}
\usepackage{tcolorbox}
\usepackage{multirow}
\usepackage{tikz}
\usepackage{capt-of}
\usepackage{graphicx}  %Required
\usepackage{pgfplots}
\pgfplotsset{compat=1.12}
\usepackage{amsmath}
\usepackage{multicol}
\usepackage{booktabs}
\usepackage{color}
\usepackage{mwe}
\usepackage{wrapfig}
\usepackage{colortbl,array,xcolor}
\usepackage{xspace}

\newcommand{\data}{$\textsc{ScholarQABench}$\xspace}
\newcommand{\model}{\textsc{OpenScholar}\xspace}

\usepackage{tikz}
\usetikzlibrary{tikzmark}
\makeatletter
\newcommand*\myfontsize{%
  \@setfontsize\myfontsize{6.7}{8}%
}
\makeatother

\renewcommand{\paragraph}[1]{\noindent \textbf{#1}}

\newcolumntype{R}[1]{>{\raggedleft\let\newline\\\arraybackslash\hspace{0pt}}m{#1}}
\usetikzlibrary{intersections}

\newcommand{\Sref}[1]{Section~\ref{#1}}
\newcommand{\Tref}[1]{Table~\ref{#1}}
\newcommand{\Fref}[1]{Figure~\ref{#1}}

\definecolor{darkgreen}{rgb}{0.0, 0.42, 0.24}
\usepackage{caption}
\usepackage{subcaption}
\usepackage{graphicx}
\usepackage{pifont}
\usepackage{titletoc}

\usepackage{amsfonts}
\usepackage{booktabs}
\usepackage{arydshln}
\usepackage{colortbl}
\usepackage{algorithm}
\usepackage[noend]{algpseudocode}
\usepackage{enumitem}
\usepackage{graphicx}
\usepackage{soul}
\usepackage{comment}
\newtoggle{comment}
\toggletrue{comment}
\definecolor{oscyan}{RGB}{23,200,221}
\definecolor{osgreen}{RGB}{5,156,5}
\DeclareRobustCommand{\hlcyan}[1]{{\sethlcolor{oscyan}\hl{#1}}}
\usepackage{colortbl,array,xcolor}

\usepackage{color-edits}

\newcommand{\mytextbox}[2]{\tikzmarknode[draw=#1,thick,inner sep=2pt]{test}{\myfontsize #2}}

\definecolor{cadmiumgreen}{rgb}{0.0, 0.42, 0.24}

\definecolor{myred}{rgb}{0.7, 0.3, 0.0}
\definecolor{myblue}{rgb}{0.2, 0.3, 0.6}
\newcommand{\acc}{\mytextbox{myred}{\textbf{\textcolor{myred}{Corr}}}}
\newcommand{\cit}{\mytextbox{myblue}{\textbf{\textcolor{myblue}{Cite}}}}
\newcommand{\expert}{\mytextbox{purple}{\textbf{\textcolor{purple}{Expert}}}}
\newcommand{\llm}{\mytextbox{osgreen}{\textbf{\textcolor{osgreen}{LLM}}}}
\newcommand{\org}{\mytextbox{osgreen}{\textbf{\textcolor{osgreen}{Org}}}}
\newcommand{\rel}{\mytextbox{osgreen}{\textbf{\textcolor{osgreen}{Rel}}}}
\newcommand{\cov}{\mytextbox{osgreen}{\textbf{\textcolor{osgreen}{Cov}}}}
\newcommand{\orgh}{\mytextbox{purple}{\textbf{\textcolor{purple}{Org}}}}
\newcommand{\relh}{\mytextbox{purple}{\textbf{\textcolor{purple}{Rel}}}}
\newcommand{\covh}{\mytextbox{purple}{\textbf{\textcolor{purple}{Cov}}}}
\newcommand{\useh}{\mytextbox{purple}{\textbf{\textcolor{purple}{Use}}}}

\newcommand{\huggingface}{\raisebox{-1.5pt}{\includegraphics[height=1.05em]{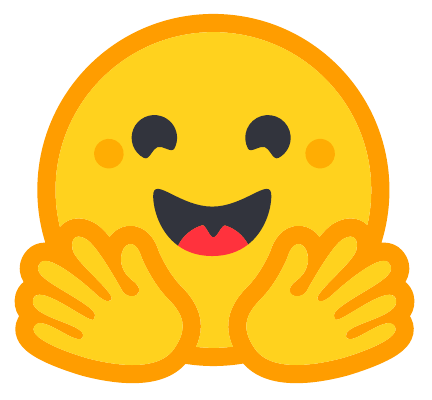}}\xspace}
\newcommand{\github}{\raisebox{-1.5pt}{\includegraphics[height=1.05em]{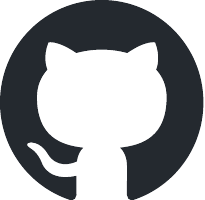}}\xspace}

\title{\textcolor{oscyan}{Open}\textcolor{white}{\hlcyan{Scholar}}: Synthesizing Scientific\\ Literature with Retrieval-augmented LMs}

\author{Akari Asai$^{1,5}$~~~Jacqueline He$^{1}$\thanks{Contributed equally (alphabetical order). All authors' contributions are detailed in the Contribution section.}~~~Rulin Shao$^{1,5*}$~Weijia Shi$^{1,2}$\\
\textbf{Amanpreet Singh$^2$~~~Joseph Chee Chang$^2$~~~Kyle Lo$^2$~~~Luca Soldaini$^2$}\\
\textbf{Sergey Feldman$^2$~~~Mike D'arcy$^2$~~~David Wadden$^2$~~~Matt Latzke$^2$}\\
\textbf{Minyang Tian$^3$~~~Pan Ji$^6$~~~Shengyan Liu$^3$~~~Hao Tong$^3$~~~Bohao Wu$^3$~~~Yanyu Xiong$^7$} \\ 
\textbf{Luke Zettlemoyer$^{1,5}$~~~Graham Neubig$^4$~~~Dan Weld$^{1,2}$~~~Doug Downey$^2$} \\
\textbf{Wen-tau Yih$^5$~~~Pang Wei Koh$^{1,2}$~~~Hannaneh Hajishirzi$^{1,2}$}\\
$^1$University of Washington~$^2$Allen Institute for AI~$^3$University of Illinois, Urbana-Champaign \\
$^4$Carnegie Mellon University~$^5$Meta~$^6$University of North Carolina, Chapel Hill~$^7$Stanford University \\
\texttt{\{akari, pangwei, hannaneh\}@cs.washington.edu} \\
}

\iclrfinalcopy % Uncomment for camera-ready version, but NOT for submission.
\begin{document}

\maketitle
\begin{abstract}

Scientific progress depends on researchers' ability to synthesize the growing body of literature. Can large language models (LMs) assist scientists in this task?
We introduce \model, a specialized retrieval-augmented LM that answers scientific queries by identifying relevant passages from 45 million open-access papers and synthesizing citation-backed responses.
To evaluate \model, we develop \data, the first large-scale multi-domain benchmark for literature search, comprising 2,967 expert-written queries and 208 long-form answers across computer science, physics, neuroscience, and biomedicine. 
On \data, \model-8B outperforms GPT-4o by 5\% and PaperQA2 by 7\% in correctness, despite being a smaller, open model. While GPT4o hallucinates citations 78–90\% of the time, \model achieves citation accuracy on par with human experts.
\model's datastore, retriever, and self-feedback inference loop also improves off-the-shelf LMs: for instance, \model-GPT4o improves GPT-4o's correctness by 12\%. 
In human evaluations, experts preferred \model-8B and \model-GPT4o responses over expert-written ones 51\% and 70\% of the time, respectively, compared to GPT4o's 32\%.
We open-source all of our code, models, datastore,  data and a public demo.

\renewcommand{\arraystretch}{1.2}
\begin{center}
\begin{tabular}{crl}
 \textbf{Demo} & \github & \href{https://openscholar.allen.ai/}{\path{openscholar.allen.ai/}}\\
  \textbf{Blog} & \github & \href{https://allenai.org/blog/openscholar} {\path{allenai.org/blog/openscholar}}\\
 \textbf{OpenScholar code} & \github & \href{https://github.com/AkariAsai/OpenScholar}{\path{github.com/AkariAsai/OpenScholar}}\\
 \textbf{ScholarBench code} & \github & \href{https://github.com/AkariAsai/ScholarBench}{\path{github.com/AkariAsai/ScholarBench}}\\
 \textbf{Checkpoints, Data, Index} & \huggingface & \href{https://huggingface.co/collections/OpenScholar/openscholar-v1-67376a89f6a80f448da411a6}{\path{OpenScholar/openscholar-v1}} \\
 \textbf{Expert Evaluation} & \github & \href{https://github.com/AkariAsai/OpenScholar_ExpertEval}{\path{AkariAsai/OpenScholar_ExpertEval}}\\
\end{tabular}
\end{center}

\end{abstract}
\begin{figure*}[h!]
\centering
    \vspace{-1.2em}
    \includegraphics[width=\textwidth]{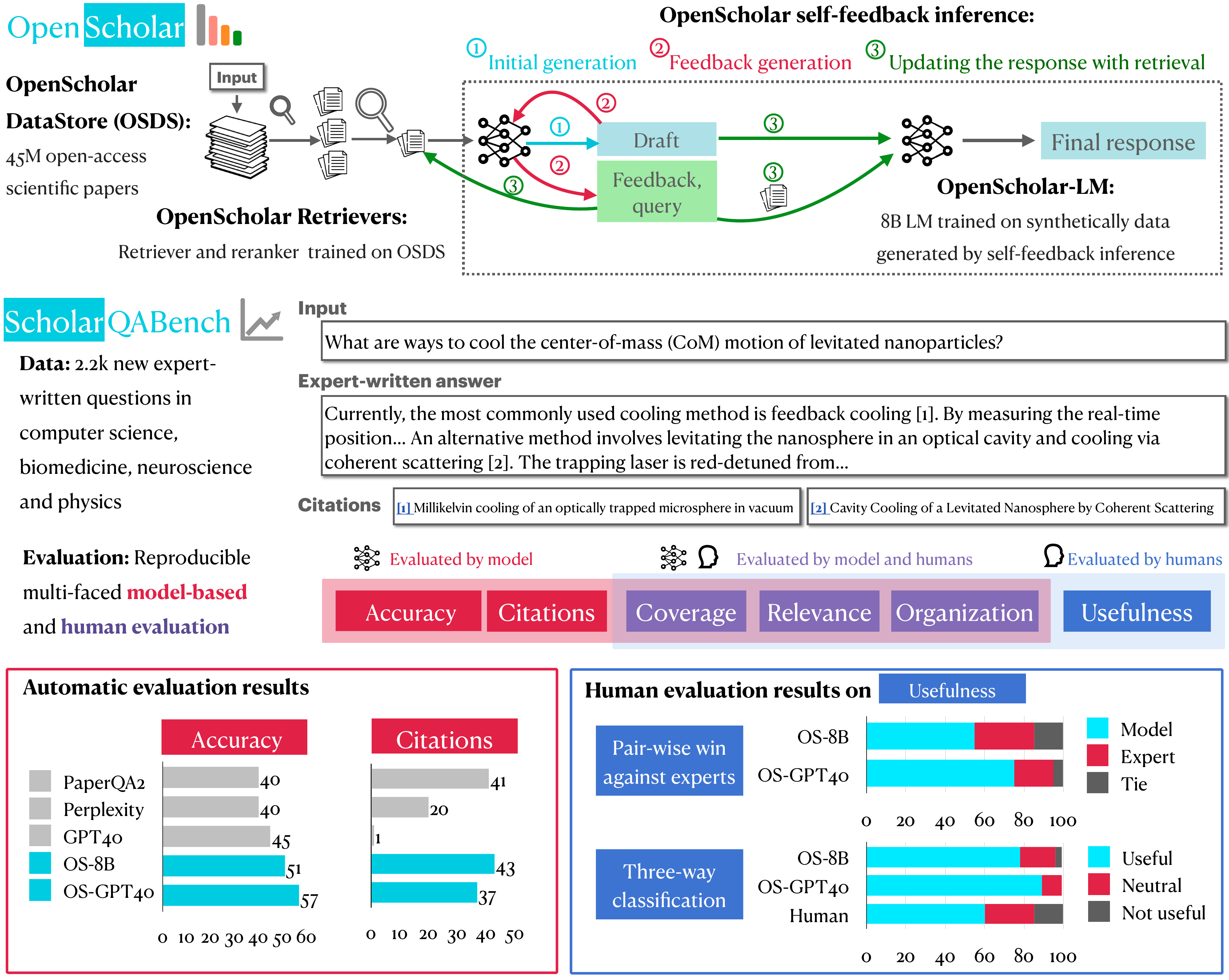}
    \caption{{\bf (Top) Overview of \model: } \model consists of a specialized datastore, retrievers and LMs and iteratively improves responses using self-feedback inference with retrieval. 
    {\bf (Middle) Overview of \data: } \data consists of 2.2k expert-written questions across multiple scientific disciplines, and we introduce automatic and human evaluation protocols for \data. 
    {\bf (Bottom) Automatic and Human Evaluation Results: } Experimental results show the effectiveness of \data, and that \model with our trained 8B or GPT4o significantly outperforms other systems, and is preferred over experts over 50\% of the time in human evaluations. 
    }
    \label{fig:overview}
\end{figure*}
\section{Introduction}
Synthesizing knowledge from scientific literature is essential for uncovering new research directions, refining methodologies, and supporting evidence-based decisions. However, the vast volume of papers published annually makes it increasingly difficult for researchers to stay informed. Effective synthesis requires precise retrieval, accurate attribution, and real-time access to current literature. 
While large language models (LLMs) show promise in assisting researchers, they face significant challenges, including hallucinations~\citep{mallen2022not,mishra2024finegrained}, reliance on outdated pre-training data~\citep{kasai2022realtime}, and a lack of transparent attribution. For instance, when tasked with citing up-to-date literature, GPT-4 fabricated citations in 78-90\% of cases across fields like computer science and biomedicine in our experiments. 

Retrieval-augmented LMs~\citep{lewis2020retrieval,guu2020retrieval}, on the other hand, can mitigate many of these issues by integrating retrieved external knowledge sources at inference-time, driving the development of systems for literature search and synthesis ~\citep{agarwal2024litllm,zheng2024openresearcher,skarlinski2024language}. 
However, many such systems rely on black-box APIs or general-purpose LLMs that are neither optimized for literature synthesis nor paired with open, domain-specific retrieval datastores (i.e., a processed corpus and corresponding retrieval index) that are specifically suited for scientific domains. 
Moreover, evaluations for scientific literature synthesis have been limited, using single-discipline and small-scale human evaluations~\citep{agarwal2024litllm,zheng2024openresearcher} or simplified tasks such as multiple-choice question answering~\citep{skarlinski2024language}. 

To address these gaps, we present \model (Figure~\ref{fig:overview}, top), a state-of-the-art retrieval-augmented LM with a specialized paper datastore and retrievers trained on scientific literature. At inference time, \model retrieves relevant passages and uses iterative self-feedback generation to refine its own output. 
We further train a new, efficient 8B LM. 
To evaluate the effectiveness of \model, we introduce \data (Figure~\ref{fig:overview}, middle), a benchmark specifically designed to enable a realistic and reproducible evaluation of open-ended scientific question answering.

\model (Section~\ref{sec:model}) uses our new \model-\textsc{DataStore} (OSDS), which contains 45 million open-access papers from Semantic Scholar, along with 237 million corresponding passage embeddings. 
To the best of our knowledge, this is the largest open-sourced datastore of scientific domains. 
\model first retrieves passages from OSDS using a retriever and reranker. Subsequently, an LM synthesizes the retrieved passages to generate responses with citations.
\model iteratively refines its outputs through natural language feedback, which improves quality and adaptively incorporates supplementary information. 
This pipeline is also used to create large-scale, high-quality training data for smaller, more efficient models. We generate synthetic queries and instructions from sampled datastore passages, feed them into \model, and use intermediate and final output to train open 8B model, \model-8B and retrieval models. 

\data (Section~\ref{sec:dataset}) is a benchmark designed to evaluate the ability of models to understand and synthesize existing research. Unlike previous benchmarks that assume that answers can be found in a single paper (e.g., scientific fact-checking; \citealp{wadden-etal-2020-fact,skarlinski2024language}), many real-world scenarios require identifying multiple relevant papers and generating long-form output with accurate citations. To address these challenges, we curated a dataset of 2,967 literature synthesis questions, along with 208 long-form responses that are written by experts and span four scientific disciplines, namely computer science, physics, biomedicine, and neuroscience. These responses were crafted by Ph.D. students and postdoctoral researchers with more than three years of experience and relevant publications in the field. On average, each response required approximately one hour to compose.  
We also introduce a multifaceted evaluation protocol that combines automated metrics and human assessments to measure citation accuracy, factual correctness, content coverage, coherence, and overall quality. 
This multifaceted approach ensures robust and reproducible evaluations, both automatic and human-driven. 

We evaluated proprietary and open models (e.g., GPT4o, Llama 3.1 8B, 70B) with and without retrieval capabilities, as well as specialized systems like PaperQA2~\citep{skarlinski2024language}, on \data (Section~\ref{sec:results}). While GPT4o demonstrated strong general performance, it struggled with citation accuracy and coverage, often producing inaccurate or non-existent citations.
\model outperformed both LM-only and retrieval-augmented pipelines, surpassing proprietary and open-source systems. Notably, using fully open-source checkpoints, \model outperformed PaperQA2~\citep{skarlinski2024language}, built on proprietary LMs, and production systems like Perplexity Pro, achieving 6\% and 10\% improvements, respectively. Additionally, \model's use of smaller, efficient retrievers significantly reduced costs.
Combining \model with GPT4o also improved correctness by 12\% over GPT4o alone.
The \model pipeline can also enhance off-the-shelf LMs. For example, when using GPT-4o as the underlying model, \model-GPT4o achieves a 12\% improvement in correctness compared to GPT-4o alone. 
 
In addition to automatic evaluations on \data, we conducted detailed expert assessments with 16 scientists from fields such as computer science, physics, and biomedicine (Section~\ref{sec:human_eval}). These experts performed pairwise and fine-grained evaluations of \model's outputs against 108 expert-written responses to literature synthesis queries in \data. 
\model, when paired with GPT-4o and our trained 8B model, consistently outperformed expert-written responses, with win rates of 70\% and 51\%, respectively. 
In contrast, GPT-4o without retrieval struggled with information coverage and was rated as less helpful than human experts, achieving only a 31\% win rate against human responses.  
This highlights that \model-generated outputs are more comprehensive, well-organized, and useful for synthesizing literature.
These findings demonstrate that \model produces high-quality outputs that are not only competitive with expert-written answers but, in some cases, exceed them, particularly in terms of coverage and organization.

\model-8B is an open retrieval-augmented LM that avoids reliance on proprietary LMs or retrieval systems, leveraging one of the largest datastores in scientific literature domains. 
We release the full \model ecosystem, including code, trained retrievers, the LM checkpoint, the datastore, the \data benchmark, expert evaluation tools, and a public demo.

\section{\model: Open Retrieval-augmented LM to Synthesizing Scientific Literature}
\label{sec:model}

\begin{figure*}[t!]
    \centering
    \includegraphics[width=\textwidth]{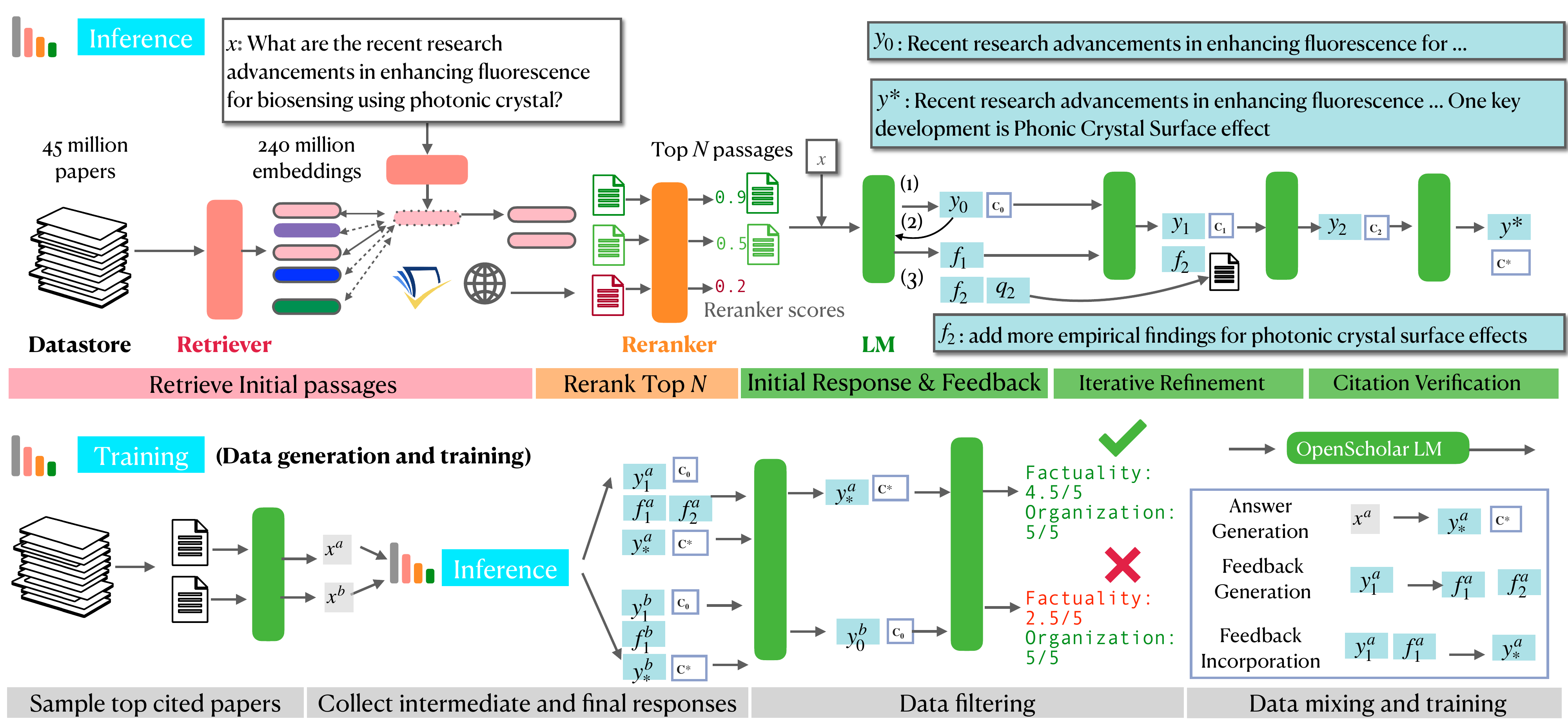}
    \caption{{\bf Detailed overview of \model inference (top) and training (bottom)}. 
    % \pw{inference}
    At inference time, given an input $x$, \model first uses a retriever to identify relevant papers from a specialized datastore (\model-Datastore), and then uses a reranker to refine and identify the top $N$ retrieved documents. The retrieved output is then passed to the LM, which generates both an (1) initial response $y_0$ and (2) self-feedback $f_1$. By incorporating its own feedback, the LM iteratively refines its output a pre-defined number of times.
    Subsequently, an LM (1) generates initial response $y_0$, (2) generates self-feedback on the initial output, and (3) incorporate feedback ($f_i$) to generates an updated response $y_1$. 
    The LM repeats the process until all feedback is incorporated. 
    To train a smaller yet competitive 8B LM, we generate high-quality training data using this inference-time pipeline followed by data filtering and mixing. 
    }
    \label{fig:overview_method_deatils}
\end{figure*}

\model (detailed in \Fref{fig:overview_method_deatils}) is a new retrieval-augmented LM designed to ensure reliable, high-quality responses to a range of information-seeking queries about scientific literature. 

\paragraph{Task formulation. }
Given a scientific query $x$, the task is to identify relevant papers, synthesize their findings, and generate a response $y$ that effectively addresses the query. This response should be accompanied by a set of citations, $\mathbf{C} = {c_1, c_2, \ldots, c_K}$, wherein each citation $c_i$ corresponds to an existing scientific paper. Each $c_i$ in $\mathbf{C}$ corresponds to specific passages from scientific literature, and should be provided as an in-line citation, linked to the relevant spans of text in $y$, following standard practice in scientific writing.
These citations allow researchers to trace the output back to the original literature, ensuring transparency and verifiability. 

\paragraph{Overview of \model.}
To ensure the retrieval of relevant papers and generate high-quality outputs, \data consists of three key components: a datastore $\mathbf{D}$, a retriever $\mathcal{R}$, and a generator LM $\mathcal{G}$. 
In standard retrieval-augmented inference pipelines, the process begins with $\mathcal{R}$, which retrieves a set of passages $\mathbf{P} = \{p_1, p_2, \ldots, p_N\}$ from $\mathbf{D}$---a large-scale corpus of previously published scientific papers—based on semantic relevance to the input query $x$. These passages serve as context for the next step. 
The generator LM $\mathcal{G}$ then takes both the retrieved passages $\mathbf{P}$ and the input query $x$ to produce the  output $y$ along with corresponding citations $\mathbf{C}$. Formally, this process can be represented as: 
\begin{equation*}
    y, \mathbf{C} = \mathcal{G}(x, \mathcal{R}(x, \mathbf{D})), 
\end{equation*}
where each $c_i$ in $\mathbf{C}$ corresponds to a specific passage from $\mathbf{P}$. 
In \model (Figure~\ref{fig:overview}), we leverage a suite of specialized components designed for scientific domains: the \model-\textsc{DataStore} $\mathbf{D}$, a \model-\textsc{Retriever}/-\textsc{Reranker}, and an LM, enabling flexible use of either off-the-shelf LMs (e.g., GPT4o) or our newly trained \model-LM. We develop self-feedback retrieval-augmented inference to improve reliability and citation accuracy. 

 \model-\textsc{DataStore} (OSDS) is a database of 45 million scientific papers, for which we build embeddings. 
We train \model-\textsc{Retriever} and \model-\textsc{Reranker} on scientific data, which passes the top $N$ passages to the generator $\mathcal{G}$ (Section~\ref{subsec:retrieval}). 
Subsequently, we use iterative self-feedback inference with retrieval: the LM first generates an initial draft $y_0$ with $\mathcal{G}$, then iteratively enhances its output through retrieval-augmented self-feedback (\Sref{subsec:inference_algorithm}).  
We use this pipeline to generate high-quality training data (\Sref{subsec:self_reflective_training}), enabling the training of specialized LMs that produce higher-quality output and more accurate citations. 

\subsection{\model Retrieval Pipeline}
\label{subsec:retrieval}

\Fref{fig:overview_method_deatils} (top left) shows our retrieval pipeline, consisting of a datastore $\mathbf{D}$, a bi-encoder retriever  $\theta_\text{bi}$, and a cross-encoder reranker $\theta_\text{cross}$. 
We first select initial candidate paragraphs using $\mathbf{D}$ and $\theta_\text{bi}$, as well as external APIs, and then refine and identify the top $N$ relevant paragraphs using $\theta_\text{cross}$.

\paragraph{Collect scientific papers to construct datastore.} 
 While prior work often uses a small subset of papers, such as arXiv papers from 2023-2024~\citep{zheng2024openresearcher}, it is important to have a diverse set of papers to improve the quality and coverage of model generation~\citep{shao2024scaling}. 
To this end, we use peS2o~\citep{soldaini2024dolma} as our retrieval source, which consists of open-access academic papers from S2ORC~\citep{lo-etal-2020-s2orc}. We  built our datastore using peS2o v3,\footnote{\url{https://huggingface.co/datasets/allenai/peS2o/tree/main/data/v3}.} which includes 45 million papers up until October 2024.\footnote{
For evaluations, we use peS2o v2 for evaluation, which consists of papers up to January 2023, as our main benchmarks and models were constructed before the v3 curation. }
 Following prior work~\citep{shao2024scaling}, we split the main text of each paper into discrete, 250-word text blocks (as determined by white space) and concatenate the paper title to each block to formulate passages in $\mathbf{D}$.
Our datastore consists of 234 million passages. 
To our knowledge, this is the largest open-sourced datastore for scientific literature.

\paragraph{Retrieve initial paragraphs.} 
We retrieve passages from three sources: (1) the peS2o datastore using our trained retriever, (2) publicly available abstract from papers returned via the Semantic Scholar API~\citep{Kinney2023TheSS} based on search keywords, and (3) publicly available texts from papers retrieved through a web search engine using the original query $x$. 
For (1), we first generate embeddings of each passage in $\mathbf{D}$ using the passage bi-encoder $\theta_\text{bi}$, which processes text chunks (e.g., queries or passages) into dense vectors~\citep{karpukhin2020dense} offline. 
Off-the-shelf retrieval models often struggle in out-of-domain scenarios~\citep{thakur2021beir}. 
To overcome this limitations, we develop $\theta_\text{bi}$ by continually pre-training Contriever~\citep{izacard2021towards} on the peS2o datastore in an unsupervised fashion to improve domain-specific retrieval performance (see Appendix ~\ref{app_sec:bi_encoder_training} for details). 
During inference, we encode the query using $\theta_\text{bi}$ and retrieve the top 100 passages through a nearest neighbor search~\citep{karpukhin2020dense}. 
For (2), we first generate keywords from the query $x$ using a generator LM. These keywords are then used to retrieve the top 10 papers for each, as ranked by citation count, via the Semantic Scholar Search API. 
This approach addresses a limitation of the Semantic Scholar API, which cannot effectively handle long, question-like search queries. 
For (3), we obtain the top 10 search results using the You.com retrieval API,\footnote{\url{https://api.you.com/}} restricting the search to academic platforms such as ArXiv and PubMed. If the papers are open-access, we extract and add their full texts to the candidate pool; otherwise, we include only their abstracts. 

\paragraph{Rerank and finalize top $N$ paragraphs.} 
After the initial stage, we have gathered over 100, or even a thousand of relevant passages per query. However, passages retrieved by the bi-encoder may include unhelpful context due to deep interactions between a query and passages, as they are encoded separately ~\citep{asai-etal-2023-task}. 
Feeding a large number of documents that might including irrelevant content to LLMs can cause efficiency and performance issues, even with state-of-the-art models~\citep{liu2023lost,xu2023recomp}. 
To overcome these challenges, we use a cross-encoder reranker~\citep{nogueira2019passage,bge_embedding}, denoted as $\theta_\text{cross}$. For each candidate paragraph, the cross-encoder reranker jointly encodes and computes the relevance score between the input query and each of the passages. We then use the relevance score to rank the passages accordingly. 
To train $\theta_\text{cross}$ for scientific domains, we fine-tune a BGE-reranker~\citep{bge_embedding} using synthetic data generated by Llama 3 70B Instruct. Specifically, we randomly generate queries based on abstracts from peS2o and retrieve the top 10 passages. Llama 3 70B Instruct then assigns relevance scores from 1 to 5 for these passages, where we consider scores of 4 or 5 as positive, and scores of 1 or 2 as negative. Passages with a score of 3 are discarded. 
More details of $\theta_\text{cross}$ training are in Appendix \ref{app_sec:raranker_training}. 
During reranking and finalization of top $N$ passages, we also implement additional meta-filtering, which includes: (1) limiting the number of passages per paper to three passages, and (2) incorporating normalized citation counts into relevance scores predicted by the cross-encoder. 

\subsection{Inference: Iterative Generation with Retrieval-augmented Self-Feedback}
\label{subsec:inference_algorithm}

In standard retrieval-augmented generation (RAG; ~\citealt{lewis2020retrieval,ram2023context}), a generator LM takes in the original input $x$ and top $N$ retrieved passages $\mathbf{P}$ and generates the output $y_0$. Although effective for tasks such as question answering~\citep{mallen2022not}, this one-step generation can lead to unsupported claims~\citep{liu2023evaluating} or incomplete output due to missing information~\citep{asai2024selfrag,jiang2023active}.  
To address these challenges, in \model, we introduce an iterative generation approach with self-feedback, which involves three steps: (1) {\bf initial response and feedback generation} to output the initial draft $y_0$ and a set of feedback on $y_0$; (2) {\bf iterative refinement with additional retrieval} to improve $y_0$ using the feedback, and (3) {\bf citation verification}. Full details are in the Appendix. 

\paragraph{Initial response and feedback generation.}
Given the input $x$ and retrieved passages $\mathbf{P}$, the generator LM first produces an initial response $y_0$  with citation markers tied to the corresponding passages in $\mathbf{P}$.
% , and uses $y_0$ as a starting point for further refinement. 
After generating $y_0$, the LM generates a set of feedback on $y_0$, $\mathbf{F} = {f_1, f_2, \ldots, f_T}$, that is aimed at improving the initial response, wherein each feedback $f_t$ is a natural language sentence that describes potential improvements. 
Although the model can generate an arbitrary number of feedback ($T$), we set a maximum limit of three feedback sentences for efficient inference. 
Unlike prior work that relies on a predefined set of feedback signals~\citep{asai2024selfrag}, our approach allows the LM to generate flexible natural language feedback on various aspects of the response, such as organization, completeness, or additional needed information. If the feedback sequence identifies missing content (e.g., ``The answer only includes empirical results on QA tasks. Add results from other task types.''), the LM also generates a retrieval query for additional retrieval using the pipeline in Section~\ref{subsec:retrieval}. 

\paragraph{Iterative refinement.} 
We then iterate over the feedback $\mathbf{F}$ to incrementally refine the output. 
If $f_k$ indicates that further retrieval is needed, the query $q_k$ is used to retrieve additional passages, which are appended to $\mathbf{P}$ before producing $y_k$.\footnote{Although we could iteratively regenerate the output each time feedback is provided, doing so introduces additional latency. Empirically, we found that feedback is often diverse, addressing different aspects of generation. As a result, sequentially incorporating feedback from the initial output remains effective.} 
The LM uses the previous output $y_{k-1}$, the retrieved passages $\mathbf{P}$, and newly retrieved passages if any, to generate an updated output $y_k$. 
This process is repeated until all feedback has been addressed, resulting in a final output $y_T$ by timestep $T$.

\paragraph{Citation verification. }
Finally, we instruct the generator LM to verify the citations in $y_t$. 
Specifically, the generator ensures that all citation-worthy statements—scientific claims requiring justification—are adequately supported by references from the retrieved passages. 
If any claims lack proper citations, the LM performs a post hoc insertion to ensure that citation-worthy statements are supported by passages. In our pipeline, we do not remove sentences that lack citation-worthy statements.

\subsection{Training: High-quality Synthetic Data Generation with Inference Pipeline}
\label{subsec:self_reflective_training}

Building powerful LMs that can effectively synthesize scientific literature is challenging due to the lack of training data for this problem. While there are some resources to train scientific LMs~\citep{wadden2024sciriff}, most tasks do not require open-retrieval settings and are single-paper tasks. As a result, most prior work in this area~\citep{skarlinski2024language} rely on proprietary LMs, which poses challenges for reproducibility and inference costs.  

We leverage our inference-time pipeline to synthetically generate high-quality training data through self-feedback, so that the resulting model can get better at generating higher-quality output without going through the self-feedback process (Figure~\ref{fig:overview_method_deatils} bottom). 

\paragraph{Question and response generations.}
Our data generation process involves three steps: first, selecting the top-cited papers from $\mathbf{D}$; second, generating information-seeking queries based on their abstracts; and third, using the \model inference-time pipeline to produce high-quality responses. 
We generate data using LLama 3.1 70B~\citep{dubey2024llama}. 
Specifically, we begin by sampling 1 million paper abstracts from the peS2o dataset and the retrieve papers' meta information such as publication year or citations. 
We then randomly select 10,000 papers that published later than 2017, and then prompt an LM to generate literature review questions or information-seeking queries based on each abstract, that may require multiple papers to answer. 
Next, we employ our \model pipeline to produce the final output $y_T$, along with intermediate generations such as feedback $\mathbf{F}$ and initial outputs. 

\paragraph{Data filtering.}
Despite its effectiveness and scalability, synthetic data may also contain issues such as hallucinations, repetitive writing, or limited instruction-following~\citep{li-etal-2024-superfiltering}. 
To address this, we introduce a two-step data filtering process: pairwise-filtering and rubric-filtering, leveraging the same LM used for data generation. 
In pair-wise filtering, we compare the quality of model outputs $y_T$ (output at the final step) and $y_0$ (initial output), and retain the output that is judged to be higher quality. 
We find that $y_0$ is preferred over $y_T$ around 20\% of the time, due to over-editing or increased redundancy after multiple iteration steps. 
We then evaluate the quality of the chosen response on a five-point scale across two aspects: {\bf organization} and {\bf factual precision and citation accuracy}. A valid model output must achieve a score of 4.5 or higher in both categories, and we discard instances whose outputs do not meet this requirement. More details are provided in the Appendix. 

\paragraph{Data mixing and training.}
From this synthetic pipeline, we generate three types of training data: answer generation $(x \rightarrow y)$, feedback generation $(y_0 \rightarrow \mathbf{F})$, and feedback incorporation $(y_{t-1}, f_t \rightarrow y_{t})$. We found that incorporating both final and intermediate outputs during training helps smaller LMs learn to generate more effective feedback. 
We further blend this synthetic training data with existing general-domain instruction-tuning data \citep{ivison2023camels} and scientific instruction-tuning data~\citep{wadden2024sciriff}, ensuring that 50\% of the training data comes from scientific domains, while the remaining 50\% is sourced from general-domain data. 
We also generate synthetic fact verification and boolean QA data based on sampled abstract data from peS2o. For this, we sort the papers based on citation count and select the top 100,000 papers. 
More details of data mixing and training are available at Appendix \ref{app_sec:training_details}. 
After data mixing, we train generator LMs on our large-scale synthetic training data. We train Llama 3.1 8B Instruct on the generated training data.

\section{\data: Realistic Literature Review Evaluation Benchmark annotated by Ph.D. Experts}
\label{sec:dataset}
\paragraph{Challenges and overview.}
Prior studies on building LLMs to synthesize scientific literature employ either small-scale, single-domain human evaluation~\citep{agarwal2024litllm,zheng2024openresearcher} or over-simplified multiple-choice QA setups~\citep{skarlinski2024language}. 
Building high-quality benchmarks for literature review has two major challenges. First, creating such datasets is resource-intensive, as it requires Ph.D.-level domain expertise and research experience, particularly when annotating realistic questions and high-quality answers. Second, even when high-quality data is available, reliably evaluating long-form natural language responses presents a significant challenge, especially in expert domains~\citep{xu2023critical,xu-etal-2024-kiwi}. This contrasts with benchmarks for other scientific processes, such as automated experimental code generation, for which clearer evaluation criteria, such as Pass@1, are more readily available~\citep{si2024can}.

To address these gaps, we introduce \data, a  benchmark that supports diverse formats of scientific literature synthesis tasks, including closed-form classification, multiple-choice, and long-form generation, as shown in Table~\ref{table:overview_dataset}.
We adopt three existing single-paper datasets, and then construct a suite of high-quality, expert annotated datasets for computer science, biomedicine,  physics, and neuroscience (Section~\ref{subsec:data_curation}). We also build a reliable automatic evaluation pipeline (Section~\ref{subsec:automatic_eval}). 
\Tref{table:overview_dataset} provides a list of tasks in \data, and Figure~\ref{fig:overview_multi_paper_qa} shows an example and an overview of the evaluation pipeline.

\begin{table}[t]
\small
    \centering
    \resizebox{\textwidth}{!}{
    \begin{tabular}{c  ccccc}
    \toprule
        \textbf{Dataset} & \textbf{Task Format} & \bf Discipline & \bf Size & \bf Evaluation & \bf Multi-paper  \\
    \midrule
    SciFact & Claim $\rightarrow$ Label & Biomedicine & 208 & \acc, \cit  \\
    \textcolor{gray}{\footnotesize{(\citealt{wadden-etal-2020-fact})}} & (True or False) &  \\\midrule
    PubMed QA & Question $\rightarrow$ Answer  & Biomedicine & 843 & \acc, \cit  \\
   \textcolor{gray}{\footnotesize{(\citealt{jin-etal-2019-pubmedqa})}} & (Yes, No) & \\ \midrule
    % BioASQ & \\
    QASA & Question $\rightarrow$ Answer  & Computer Science & 1,375 & \acc, \cit & \\
    \textcolor{gray}{\footnotesize{(\citealt{lee2023qasa})}} & (Long-form) & \\ \midrule\midrule
    \textsc{ScholarQA-CS}& Question $\rightarrow$ Answer$^\dagger$ & Computer Science & 100 & \acc,\cit  & \multirow{2}{*}{\checkmark}\\
    & (Long-form) &  & &  \\ \midrule
    \textsc{ScholarQA-Bio} & Question $\rightarrow$ Answer$^*$ & Biomedicine & 1,451 & \cit  & \multirow{2}{*}{\checkmark}\\
    & (Long-form) &  & & \\ \midrule
    \textsc{ScholarQA-Neuro} & Question $\rightarrow$ Answer$^*$ & Neuroscience & 1,308 & \cit  & \multirow{2}{*}{\checkmark}\\
    & (Long-form) &  & & \\  \midrule
    \textsc{ScholarQA-Multi} & Question $\rightarrow$ Answer & Computer Science, Physics, & 108 & \cit,  & \multirow{2}{*}{\checkmark}\\
    & (Long-form) &  Biomedicine &   & \llm, \expert \\
    \bottomrule
    \end{tabular}
    }
        \caption{{\bf Overview of \data}. The top three rows show single-paper datasets adopted from prior datasets. 
        The bottom four rows are new datasets, which we constructed by recruiting Ph.D.-level experts. 
        Answer$^*$ indicates that the dataset comes with questions only, and answer$^\dagger$ indicates that answer will be evaluated based on human-annotated rubrics. 
        The evaluation columns corresponds to the multi-faceted evaluations in Section~\ref{subsec:automatic_eval}. The ``Multi-paper'' column indicates whether the task requires multiple papers to answer. 
        \llm~and \expert~indicate the fine-grained evaluations by evaluator LLMs (i.e., Prometheus; \citealt{kim2024prometheus}) and expert humans, respectively. 
    }
    \label{table:overview_dataset}
\end{table}

\subsection{Data Curation}
\label{subsec:data_curation}
\data is designed to evaluate model capabilities in automating scientific literature review. The curation process is guided by three key factors: 
{\bf Diversity of tasks}: \data includes tasks with a range of input-output formats; 
{\bf Diversity of disciplines}: Unlike previous analyses that often focus on a single discipline such as computer science, 
\data spans four scientific disciplines; 
{\bf Inclusion of multi-paper tasks}: Unlike prior work that focuses on understanding single, pre-selected papers, all tasks require retrieving from the entire open-access collection of full text of papers (\Sref{subsec:single_paper_task}) 
 and four datasets specifically require reasoning over multiple retrieved papers (\Sref{subsec:multi_papers}).

\subsubsection{Single-Paper Tasks}
\label{subsec:single_paper_task}
For single-paper tasks, we curate and adapt existing widely-used single-paper datasets. Figure\ref{fig:single_paper_examples} shows examples of single-paper tasks; more details are in Appendix~\ref{app_sec:annotation_details}. 

\paragraph{SciFact.}~SciFact \citep{wadden-etal-2020-fact} is a dataset of 1.4K expert-written scientific claims in the biomedical domain, paired with gold evidence from existing PubMed paper abstracts annotated with labels and rationales. 
We include validation set queries labeled as either \texttt{supports} (true) or \texttt{contradicts} (false), discarding the original gold evidence, and reformulate the task as binary open-retrieval, wherein a system needs to identify relevant papers from a large collection of papers.  

\paragraph{PubMedQA.}
{PubMedQA}~\citep{jin-etal-2019-pubmedqa} has expert-annotated (yes/no/maybe) QA data on PubMed paper abstracts. 
Similarly to {SciFact}, we only keep instances with yes or no labels, and discard the original abstract passage to formulate the task as an open-retrieval setup. 

\paragraph{QASA.}
{QASA}~\citep{lee2023qasa} is a single paper QA dataset that consists of question answering pairs, requiring reasoning over scientific articles in AI and ML. 
We evaluate the model's ability to sufficiently answer a detailed question about the target paper. While the original dataset provides three subtasks (answer selection, rationale generation and answer compositions) as well as end-to-end QA, we evaluate the models' performance based on an end-to-end QA setup.

\begin{figure*}[t!]
    \centering
    \includegraphics[width=\textwidth]{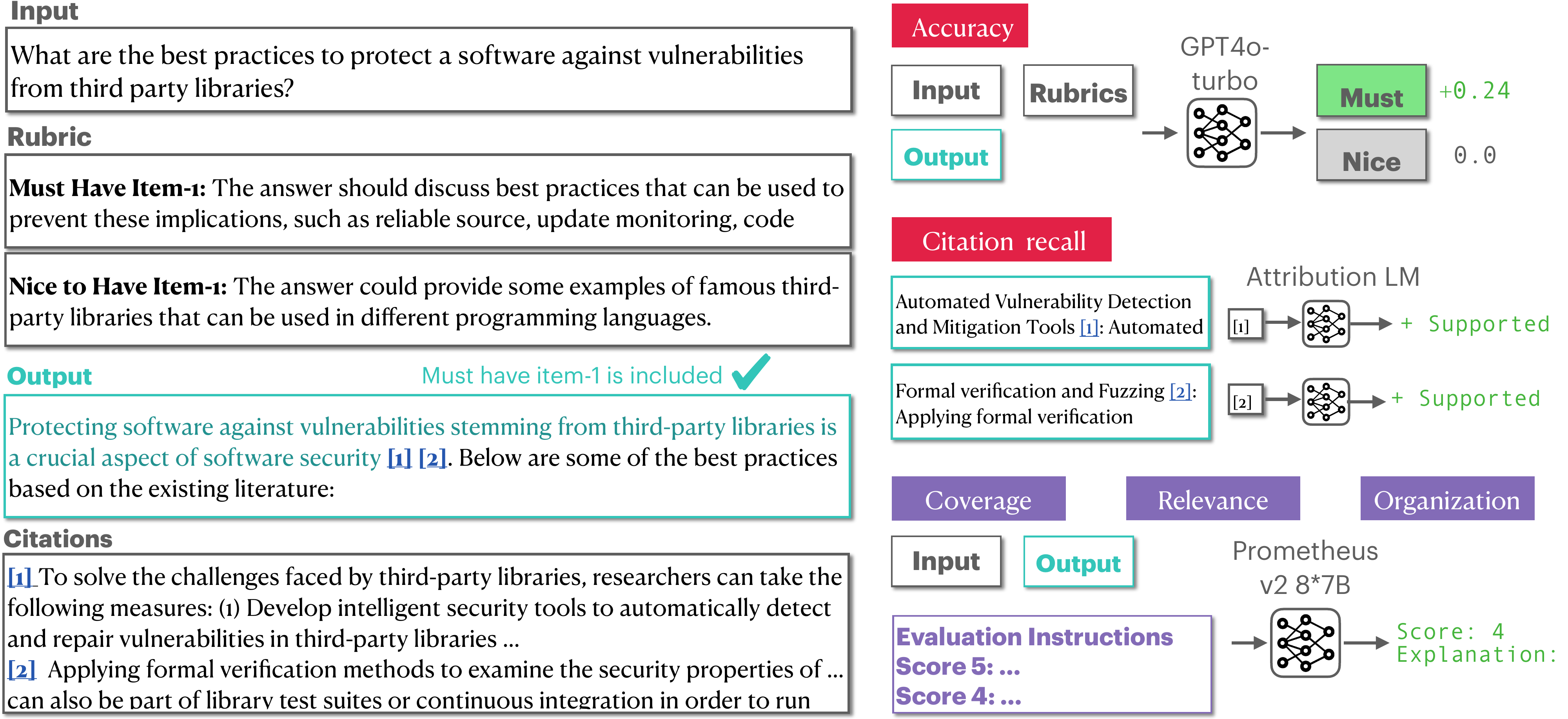}
    \caption{{\bf An \textsc{ScholarQA-CS} example and evaluation overview.} \textsc{ScholarQA-CS} consists of 100 questions and an average of 4.4 expert-written rubrics to be satisfied. Our \data evaluation pipeline evaluates aspects like correctness and citation accuracy.  
    }
    \label{fig:overview_multi_paper_qa}
\end{figure*}

\subsubsection{Multi-paper Tasks}
\label{subsec:multi_papers}
Single-paper, closed-set tasks may provide reliable evaluations. However, they may not be reflective of realistic scenarios, in which complex, open-ended questions are asked independently from existing papers, and require multi-paper retrieval and reasoning. 
Few datasets~\citep{xu-etal-2024-kiwi,malaviya2023expertqa} explore multi-paper setups with realistic queries, and most lack a reliable evaluation pipeline or human-written references. We address this gap by curating three new long-form QA datasets, annotated by experts, for these challenging settings (details in Appendix~\ref{app_sec:annotation_details}). 
Furthermore, our multi-paper tasks include four scientific disciplines.

\paragraph{\textsc{ScholarQA-CS}.} 
We collected 100 questions along with detailed answer rubrics for each question across various computer science disciplines by recruiting expert annotators holding Ph.D.s in the field (professors, postdoctoral researchers, and research scientists). 
Annotators were tasked with writing literature review questions that were expected to require multiple research papers to answer. 
The question topics span areas such as networks, algorithms, the Internet of Things, artificial intelligence, and human-computer interaction. 
Then, for each question, two other annotators searched the Web to produce a rubric listing the key ingredients for a correct answer, categorized by importance (``must-have'' and ``nice-to-have''), along with supporting quotes from sources for each ingredient. {The annotators were instructed not to use any LLM services for this initial part of the task. After the initial web search, the annotators were shown corresponding responses from four LLM services {(Claude 3.5 Sonnet, GPT-4o, Perplexity Pro and an unpublished RAG prototype based on Claude 3.5)} in a randomized order in case they wanted to revise their rubrics.} On average, each question is annotated with 4.4 key ingredients, each supported by 4.4 quotes. 

{To measure agreement, we had both annotators produce rubrics for a subset of 10 randomly sampled questions. We then compute the scores for responses from the four LLM services the annotators were exposed to using our automated approach, once for each set of annotator rubrics. Finally, we compute Pearson's correlation coefficient among the scores for each question and take the mean. Since the rubric annotation task is subjective, we compute this agreement both with and without the general criterion as part of the scores, which comes out to be 79.3 and 59.5, respectively.}
Figure~\ref{fig:overview_multi_paper_qa} shows one example, and more examples and details are available in Appendix \ref{app_sec:examples_rubric_qa}. 

\paragraph{\textsc{ScholarQA-Bio}, \textsc{ScholarQA-Neuro}.}
We further collected 2,759 expert-written literature review questions in biomedicine and neuroscience, recruiting six experts who have a Ph.D. in relevant areas and are currently research scientists and engineers.  
{The annotators were asked to choose papers from their area of expertise,} and generate complex scientific questions that biomedical scientists might reasonably ask about the scientific literature {based upon their parsing of those papers}. We collected questions from different areas, such as bioimaging, genetics, microbiology, and neuromodulation for each.  
Due to the cost of annotation, we focused solely on curating the questions. 
Full instructions and examples are available in Appendix~\ref{tab:annotator instructions_biomed} and \ref{app_sec:examples_biomed_queries}. 

\paragraph{\textsc{ScholarQA-Multi}.}
Lastly, we collected 108 literature review questions and expert-written answers with citations in four domains: computer science (AI/ML, HCI), Biomedicine (Bioimaging,  Genetics), Physics (Astrophysics, Photonics, Bio Physics). 
All annotations are conducted by Ph.D. students or post-Ph.D. scientists, who have more than three years of research experience in the corresponding areas and have multiple first-author publications. 
We asked them to come up with questions that are related to most recent literature, and to compose answers to the questions using relevant papers that they found via search. Our annotators were instructed not to use any LLM-based systems such as ChatGPT, and told to only use general search (e.g., Google Search) or paper search systems (e.g., Semantic Scholar). 
Table~\ref{table:result_human_data} show statistics of the collected questions and answers and the distribution of subjects are in Figure~\ref{fig:data_fistributions}, along with the average annotation time per subject. We show several examples in Appendix \ref{app_sec:examples_annotated}. 
On average, each annotator spent 56 minutes per instance.

\subsection{Metrics and Evaluation Protocols}
\label{subsec:automatic_eval}
We developed a multifaceted automatic evaluation pipeline to facilitate reproducible and efficient evaluations, complementing expert assessments. An overview of our evaluations is in Figure~\ref{fig:overview_multi_paper_qa}.

\paragraph{Correctness (\acc).}
Correctness evaluates the degree of overlap or matching of a model-generated answer and human-annotated reference. This metric is only applied to tasks with human-annotated reference answers. 
For short-form generation tasks given a fixed set of answer classes, namely SciFact and PubMedQA, we use accuracy as the correctness metric. 
For QASA, we use ROUGE-L as an evaluation metric, following \cite{lee2023qasa}. 
For \textsc{ScholarQA-CS}, we develop a new long-form evaluation pipeline, which employs expert-annotated rubrics. Each rubric has two criteria: general (accounting for 40\% of the score) and annotation-driven (60\%). 
General criteria cover the evaluation of length, expertise, citations, and excerpts, while annotation-driven criteria involve assessing the presence of specific key ingredients identified by annotators. GPT4o-turbo assigns scores for each criterion, and a weighted sum is computed to obtain a final score. 
More details are in appendix \ref{app_sec:details_scholar_cs}. 

\paragraph{Citation accuracy (\cit).}
Evaluating long-form responses to literature review questions requires citation accuracy: LMs should correctly attribute relevant evidence for all citation-worthy statements. In \data, all systems generate outputs with reference numbers (e.g., \texttt{[1], [2]}) linked to passages provided during inference. Following prior work~\citep{gao2023enabling,liu2023evaluating}, we measure citation precision and recall. Specifically, we check if each citation-worthy statement has appropriate citations and if the citations support the statement ({\bf Citation Recall}, \cit-r). 
For each citation, we then verify its relevance and necessity---specifically, whether the citation supports the statement and if its removal impacts the integrity of remaining citations ({\bf Citation Precision}, \cit-p). 
Finally, we compute {\bf Citation F1} (\cit-F1) as well, and use it as a primarily metric for citation accuracy. 
Citation accuracy does not require gold reference answers or rubrics, so we apply this evaluation across all tasks. 

\paragraph{Content quality and organization (\llm, \expert).}
We further define key aspects to evaluate the generated answers beyond \acc~ or \cit~ alone. Specifically, we assess {\bf Relevance} (\rel) to the question, {\bf Coverage} (\cov) of topics (e.g., diversity of discussed papers) and depth (e.g., sufficiency of details), and {\bf Organization and Writing Flow} (\org). These aspects are challenging to capture with standard metrics.  
 Since LMs can effectively follow detailed evaluation rubrics~\citep{zheng2023judging,kim2024prometheus}, we use Prometheus v2~\citep{kim2024prometheus} to assign five-scale scores based on defined rubrics and use the same schema for human evaluations. For human evaluation, we further evaluate {\bf Overall Usefulness}  (\useh).  
Full instructions for this evaluation are provided in the Appendix \ref{app_section:evaluations}. 
As prior studies show that \llm~is less reliable when gold reference answers are not available~\citep{kim2024biggen}, this evaluation is only applied to a task with human-annotated reference answer, namely \textsc{ScholarQA-Multi}. 
We analyzed the agreement between human and model assessments on fine-grained aspects (Appendix \ref{app_sec:human_evaluation}) and found that the model's evaluations often align with human rankings, showing higher correlation especially in organization and coverage.

\section{Experiments and Results}
\label{sec:results}

\subsection{Experimental Details}
\paragraph{Models.}
First, we evaluate both open-weight and proprietary LMs, including Llama 3.1 (8B, 70B) and GPT-4o (\texttt{gpt-4o-2024-05-13}). In this setup, each LM generates an answer independently, without external retrieval, and provides a list of referenced paper titles. For evaluation, we verify whether the generated paper titles exist. If they do, we retrieve their corresponding abstracts to use as citations. 
For multi-paper tasks, we further evaluate other proprietary systems: Perplexity Pro,\footnote{\url{https://www.perplexity.ai/}. We used the paid subscription version for the experiments. Note that Perplexity Search does not have an API, so we use the \texttt{selenium} toolkit and save their final prediction results from the interface. Due to this, we could not retrieve their citation information.} and PaperQA2~\citep{skarlinski2024language}, a concurrent literature review agent system that uses GPT4o for reranking, summarization, and answer generation.\footnote{We use their official codebase. As PaperQA2 does not release their retrieval corpus and requires downloading PDF files offline, we downloaded PDF files of papers suggested by our retrieval pipelines and the Semantic Scholar API. Unlike PaperQA2, we do not have access to private or license-protected papers, which may limit the effectiveness of our replication to some extent.}  
Then, we evaluate models using our \textsc{OpenScholar-DataStore} (+OSDS), where we retrieve the top $N$ passages, and concatenate and feed them together with the original input.  
Lastly, we evaluate our proposed \model, leveraging our custom inference-time pipeline using trained 8B model models ({\bf OS-8B}), as well as Llama 3.1 70B and GPT4o ({\bf OS-70B}, {\bf OS-GPT4o}).  

\paragraph{Details of \model.}
We use peS2o v2 as our default datastore $\mathbf{D}$. We analyze the effect of different datastores in Appendix \ref{app_sec:pes2o_ds_analysis}. 
For $\theta_\text{bi}$ and $\theta_\text{cross}$ in \model, we use our trained bi-encoder and cross-encoder models, which consist of 110 million and 340 million parameters, respectively. We set the maximum number of papers from web search and Semantic Scholar to 10. 
For the generator LMs, we set the temperature to 0.7 and limit the maximum token count to 3,000 for response generation and 1,000 for feedback generation, and use the \texttt{vllm} package for faster inference. We train Llama 3.1 8B for two epochs on 130k training instances for two epochs, using \texttt{torchtune}.
Further details are in Appendix~\ref{app_sec:experimental_details}. 
For all models, we set the number of passages input into the generator LM to five for single-paper tasks and ten for multi-paper tasks. No few-shot demonstrations are provided, except for SciFact and PubMed, where we include one-shot demonstrations.

\subsection{Results}
Table~\ref{table:result_multi} show scores for multiple aspects of the main baselines. 
In summary, \model achieves state-of-the performance, significantly outperforming GPT4o and their standard RAG version, as well as specialized literature review systems such as PaperQA2~\citep{skarlinski2024language} by a large margin.

\paragraph{Single-paper tasks.} 
On single-paper tasks, \model consistently outperforms other models. OS-8B and OS-70B outperforms Llama 3.1 8B and 70B with and without retrieval augmentation in terms of final \acc~and \cit~ in Table~\ref{table:result_multi}. 
OS-70B even matches or outperforms GPT4o on PubMedQA and QASA. 

\begin{table}[t!]
\centering
\small
    \resizebox{\textwidth}{!}{
    \begin{tabular}{l|cc|cc|cc||cc|cc|cc|c}
    \toprule
      & \multicolumn{6}{c||}{Single-paper performance} & \multicolumn{6}{c|}{Multi-paper performance} & Cost \\\midrule
      & \multicolumn{2}{c|}{Pub} & \multicolumn{2}{c|}{Sci} & \multicolumn{2}{c||}{QASA} & \multicolumn{2}{c|}{CS} & \multicolumn{2}{c|}{Multi} & Bio & Neu & CS \\ 
    \textbf{Model} & \acc & \cit & \acc & \cit & \acc & \cit &  \acc & \cit & \llm &  \cit & \cit & \cit  & USD / q \\  \midrule 
    Llama3-8B & 61.5 & 0.0 & 66.8 & 0.0 & 14.3 & 0.0 &  41.9 & 0.0 & 3.79 & 0.0 & 0.0 & 0.0 & 0.0001 \\ 
    +OSDS & 75.2 & 63.9 & 75.5 & 36.2 & 18.6 & 47.2 &   46.7 & 26.1 & \bf 4.22 & 25.3 & 38.0 & 36.8 & 0.0001 \\
    {\bf OS-8B} & {\bf 76.4} & {\bf 68.9} & {\bf 76.0} & {\bf 43.6} & \bf 23.0 & \bf 56.3   &  \bf 51.1 & \bf 47.9 &4.12 & \bf 42.8 & {\bf 50.8} & {\bf 56.8} & 0.003 \\\midrule
    Llama3-70B & 69.5 & 0.0 & 76.9 & 0.0 & 13.7 & 0.0   & 44.9 & 0.0 & 3.82 & 0.0 & 0.0 & 0.0 & 0.0004\\ 
    +OSDS & 77.4 & 71.1  & 78.2 & 42.5 & 22.7 & 63.6  &  48.5 & 24.5 & 4.24 & 41.4 &  53.8  & 58.1 & 0.0004 \\
    {\bf OS-70B} & \bf 79.6 & \bf 74.0 & {\bf 82.1} &\bf 47.5 & {\bf 23.4} & \bf 64.2 & {\bf 52.5} & \bf 45.9 &  4.03 & \bf 54.7  & {\bf 55.9} & {\bf 63.1} & 0.01 \\\midrule
    GPT4o & 65.8 & 0.0 & 77.8 & 0.0 & \bf 21.2 & 0.0   & 45.0 & 0.1 & 4.01 & 0.7 & 0.2 & 0.1  & 0.006  \\ 
    +OSDS & \bf 75.1 & 73.7 & 79.3 & 47.9 & 18.3 & 53.6 & 52.4 & 31.1 & 4.03  & 31.5 & 36.3  & 21.9 & 0.01 \\
{\bf OS-GPT4o} &  74.8 &  \bf 77.1  & \bf 81.3 &\bf 56.5   &18.7   &\bf 60.4 &{\bf  57.7} &  \bf 39.5  & {\bf 4.51} & \bf 37.5 & \bf{51.5} & \bf 43.5  & 0.05 \\\midrule
    PaperQA2 & -- & -- & -- & -- & -- & -- & 45.6 & {48.0} & 3.82 & { 47.2} & {56.7} & 56.0 & 0.3$\sim$2.3 \\
    Perplexity & -- & -- & -- & -- & -- & -- & 40.0 & -- & 4.15 & -- & -- & -- & 0.002$^{**}$\\
    \bottomrule
    \end{tabular}
    }
\caption{{\bf Results of \data}. CS, Multi, Bio and Neu indicate \textsc{Scholar-CS}, \textsc{ScholarQA-Multi}, \textsc{ScholarQA-Bio} and \textsc{ScholarQA-Neuro}, respectively. 
\acc~indicates correctness metrics (accuracy for PubMedQA and SciFact, ROUGE-L for QASA and Overall scores for \textsc{ScholarQA-CS}) and \cit~indicates citation F1. \llm~indicates the average score of \org~(organization), \rel~(relevance), \cov~(coverage) as predicted by Prometheus~\citep{kim2024prometheus}. $^*$PaperQA2 is based on GPT4o, and its pricing is dependent on the number of PDF files used during inference. 
For the 8B and 70B model costs, while evaluations were conducted on our local machines, we estimated costs based on Together.ai pricing.
$^{**}$We used Perplexity Pro (which requires a monthly subscription at \$20 USD) and divided this cost by 9,000, which is the maximum number of queries allowed under the Pro subscription. 
Since the Perplexity UI does not provide snippets for each citation, we were unable to evaluate its citation accuracy. 
}
\label{table:result_multi}
\end{table}%
\begin{table}
\small
    \centering
        \resizebox{\textwidth}{!}{
    \begin{tabular}{l| ccc|ccc}
    \toprule
        & \multicolumn{3}{c}{Computer Science} & \multicolumn{3}{c}{Biomedicine} \\ 
      Model  &Total \# & \# of Hallucinated  ($\downarrow)$ & Ratio ($\downarrow)$ &  Total \# &  \# of Hallucinated  ($\downarrow)$ & Ratio ($\downarrow)$  \\
    \midrule
     OS-8B &  9.65 &  0.0 & 0.0 & 6.25 &  0.0 & 0.0 \\ \hdashline
    Llama 3.1 8B &  5.20 & 4.79 & 92.1\% & 5.58 & 5.46 & 97.6\% \\
    Llama 3.1 70B & 6.14 & 4.78 & 78.1\% & 6.98 & 6.74 & 96.6\%\\
    GPT4o  & 5.74 & 4.52 & 78.7\% & 5.24  & 4.97 & 94.8\% \\
    % GPT4o & \checkmark & & & \\ 
    % % Llama3.1 70B  &  \checkmark & & & \\ 
    % Ours 8B &  \checkmark \\
    \bottomrule
    \end{tabular}
    }
\caption{{\bf Statistics of hallucinated papers in computer science and biomedicine domains}. Our analysis revealed a significant number of non-existent cited papers in predictions made by LLMs without retrieval, which is a problem not observed in \model.}\label{table:result_non_parametric}
\end{table}
\paragraph{Multi-paper tasks.} 
\model-8B, 70B, and GPT4o (OS-8B, OS-70B and OS-GPT4o) demonstrate strong performance in multi-paper tasks. Specifically, OS-GPT4o provides a 12.7 point improvement over GPT4o alone in \textsc{Scholar-CS} \acc~and a 5.3 improvement over standard RAG. 
When combined with trained OS-8B, \model significantly outperforms the pipeline that uses off-the-shelf Llama 3.1 8B, showcasing the benefits of domain-specific training. 
Furthermore, this \model-8B outperforms proprietary systems such as GPT4o, Perplexity Pro, or PaperQA2, which uses GPT4o models for passage reranking, summarization and answer generation, by a substantial margin. 
Notably, by leveraging efficient retrieval pipelines with lightweight bi-encoders, cross-encoders, and in-house models, \model-8B and \model-GPT4o achieve significantly lower costs—orders of magnitude cheaper than PaperQA2---while maintaining high performance.  

\paragraph{Limitations of parametric LMs.}
On both single-paper and multi-paper tasks, we observe that non-retrieval augmented baselines struggle and retrieval is almost always conducive to achieving better performance, and models without any retrieval often struggle to generate correct citations and show limited coverage on multi-paper tasks. 
As shown in Table~\ref{table:result_non_parametric}, the proportion of cited papers that actually exist is strikingly low. In particular, while models such as GPT4o and Llama can generate plausible reference lists, we find that 78-98\% of the cited papers are fabricated, and the issue is exacerbated in biomedical domains. Even when citations refer to real papers, the majority are not substantiated by the corresponding abstracts, resulting in near-zero citation accuracy. 

We also observe that such models also generate responses with limited coverage. 
On \textsc{Scholar}-Multi, non-retrieval models (Llama 3.1 8B, 70B, and GPT4o) consistently exhibit significantly lower average scores compared to retrieval-augmented models. This discrepancy is largely driven by much lower \cov~scores; for instance, Llama 3.1 8B achieves a \cov~score of 3.45, while Llama 3.1 8B + OSDS (a standard RAG baseline) improves it to 4.01. 
These results suggest that solely relying on models' parametric knowledge alone is particularly difficult in scientific domains, especially for smaller LMs. 

\subsection{Analysis}
\begin{figure}[t!]
\begin{subfigure}[b]{0.3\textwidth}
\centering
\footnotesize
\begin{tabular}{lcc}
\toprule
 & \multicolumn{2}{c}{\textsc{Scholar-CS}}  \\
 & \acc & \cit \\
\midrule
\textsc{OS}-8B & 51.3 & 47.9  \\ \hdashline
- training & 49.4 &  42.3 \\
- reranking & 49.6 & 28.2  \\
% -  meta filter & 49.2 & 36.7 \\
- feedback & 51.1 &  50.2  \\
-  attribution & 49.3 & 44.0  \\
\midrule
OS-\textsc{GPT4o} & 57.7  & 39.5 \\ \hdashline
-  reranking &  52.4  & 22.9 \\
-  feedback & 55.1 & 31.0  \\
-  attribution & 55.6 & 30.6  \\
\bottomrule
\end{tabular}
\caption{Ablation of different components of \model.}
\label{tab:ablatioon}
\end{subfigure}%
  \hspace{1.2cm}
  \begin{subfigure}[b]{0.28\textwidth}
        \includegraphics[width=\textwidth]{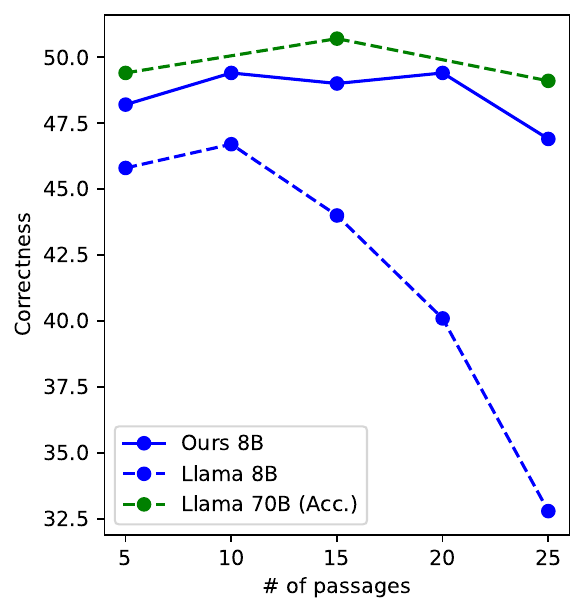}
        \caption{Top N Ablations (\acc) on \textsc{Scholar-CS}. }
        \label{fig:top_n_ablations_accuracy}
  \end{subfigure}%
  \hspace{0.3cm}
  \begin{subfigure}[b]{0.27\textwidth}
        \includegraphics[width=\textwidth]{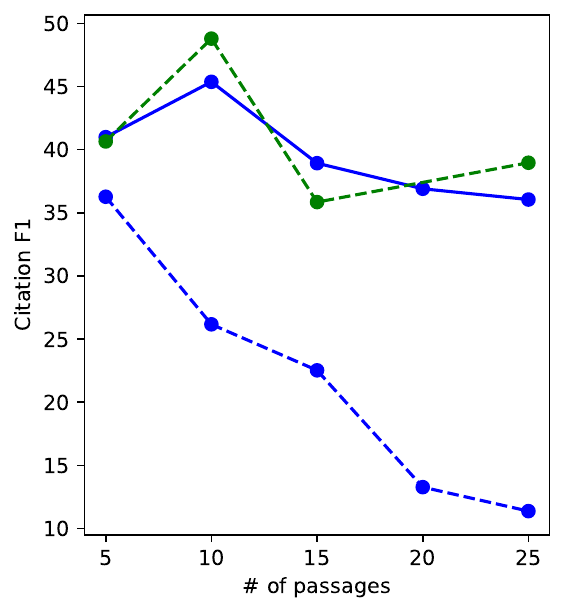}
        \caption{Top N Ablations (\cit-F1) on \textsc{Scholar-CS}. }
        \label{fig:top_n_ablations_citation}
  \end{subfigure}%
\caption{{\bf Analysis on \model:} (a) {\bf Ablation studies} for key components of \model training and inference based on different underlying LMs.  (b) {\bf Top N docs:} Analysis of the effect of varying the number of context chunks for final downstream tasks. We evaluate final model performance based on citation accuracy and correctness on multi-doc QA tasks, using \model 8B and Llama 3.1 8B. }\label{fig:analysis}
\end{figure}

\paragraph{Ablation studies.}
We conduct ablations to assess the effectiveness of individual components of \model (inference and training). Specifically, we remove each of the inference-time procedures, reranking, feedback and attribution, and for OS-8B, we ablate the training, where we use Llama3-8B without any further training.  

As shown in Figure~\ref{fig:analysis} (a), removing these components significantly impacts both the overall correctness and citation accuracy of the model outputs. Notably, removing the reranker led to substantial performance drops on both models. 
The pronounced decline in performance after removing feedback loops in  GPT4o suggests that more powerful models greatly benefit from a self-feedback cycle, consistent with ~\citet{madaan2023self}, while it gives limited performance drops in our trained 8B.  Additionally, the removal of post-hoc attribution assessments negatively affected both citation accuracy and final output correctness, highlighting the importance of ensuring that models verify their outputs. 
The significant performance gap between trained versus vanilla OS-8B suggests that further training on high-quality, domain-specific data is key to building efficient, task-specialized LMs. 
In the next analysis, we demonstrate that training has a significant impact on an LM's ability to effectively utilize more context, while maintaining citation accuracy. 

\paragraph{Number of context passages.}
We analyzed how varying the number of context passages  (top $N$) impacts model performance. 
Specifically, we experimented with Standard \texttt{RAG} and \model using our trained 8B model and Llama 3.1 8B, and evaluated both generation accuracy and citation accuracy on \textsc{Scholar-CS}. 
Figures~\ref{fig:analysis} (b)(c) show the results.
Although Llama 3.1 is trained to handle and accept a context length of up to 128K tokens, we found that its performance deteriorates after a certain context size. While increasing the top $N$ context window from 5 to 10 does improve the model’s correctness score, expanding further actually worsens both correctness and citation accuracy. This suggests that even though LMs can process large numbers of passages, they may struggle to effectively use them without specialized training, particularly for smaller models. 

In contrast, our trained 8B model maintains strong performance for up to $N=20$ passages. 
We also found larger models such as Llama 3.1 70B to be more robust to increased context length. 
In terms of citation accuracy, as shown in Figures~\ref{fig:analysis} (c), Llama 3.1 8B observes quick decline and citation F1 gets as low as 10, while our 8B LM and Llama 70B both maintain around 40 citation F1, although they also see some performance decline.

\section{Expert Evaluation}
\label{sec:human_eval}
To complement our automatic evaluations and better understand the effectiveness and limitations of \model, we conducted human evaluations. This study involved over 100 literature review questions and more than 15 participants, including Ph.D. students, research scientists, and university professors with expertise in the relevant fields. In total, we curated more than 400 fine-grained expert evaluations on human and model answers.

\subsection{Human Evaluation Design}
\paragraph{Evaluations against human experts.}
For human evaluations, we use 108 question-answer pairs from \textsc{ScholarQA-Multi}, written by experts. 
We run three models on these questions to generate answers with citations: GPT4o (without external retrieval), \model with GPT4o as the generator (OS-GPT4o), and \model with our trained 8B model (OS-8B). Expert annotators are then asked to evaluate the model-generated answers against human-written answers.

Each evaluation involves presenting a question, a model-generated answer, and a human-written answer. Expert annotators then conduct fine-grained assessments of each answer and provide pairwise preference judgments between the two. 
For fine-grained evaluations, we use the five-scale evaluation criteria described in \Sref{sec:dataset} (\covh, \orgh, \relh), with annotators scoring both model and human answers using the same rubrics. 
For usefulness (\useh), annotators assign scores on a scale from 1-5, which we convert into three classes: Not Useful (1-2), Neutral (3), and Useful (4-5). We then calculate the percentage of answers that fall into the Useful category. For pairwise preference, annotators either choose one of the answers or mark a ``tie'' if they judge both answers to be of equal quality. Optionally, experts provide explanations on why one answer is better than the other.  

\if0{
We compare the following models: 
\begin{itemize}[leftmargin=10pt]
    \item {\bf \model with GPT4o}: Answers generated by our \model using GPT4o as the generator LM. 
    \item {\bf \model with Our 8B}: Answers generated by our \model pipeline using our newly trained 8B model, which is fully open. 
\end{itemize}
}\fi

\paragraph{Expert annotators for answer writing.} 
Our expert annotators for question and answer writing are 12 Ph.D. students and post-doctoral researchers from research institutions across the United States, all of whom have at least three years of research experience and published multiple papers in journals or conferences from their fields.  
Our annotators' expert areas span computer science (Natural Language Processing, computer Vision, Human-Computer Interaction), physics (Astrophysics and Photonics / Optics), and biomedical (Neuroscience, Bioimaging) domains, and we assign our expert annotators to  questions in their expertise. On average, we paid 35-40 USD per person. 

\paragraph{Expert annotators for evaluations.} 
A total of 16 expert annotators from the three fields contributed to our evaluations, with 12 of them also participating in answer generation. All expert annotators met the same qualifications as those who composed the answers. To minimize potential biases, we ensured that annotators did not evaluate responses to their own questions by assigning evaluation tasks to different groups of experts. 
Each instance was reviewed by 1 to 3 expert annotators, depending on availability. The inter-annotator agreement was 0.68 using pairwise comparison with ties, and 0.70 using a relaxed approach, wherein ties were merged. On average, each expert spent five minutes per instance on evaluation, and received compensation ranging from 25–35 USD. 

\begin{table}[t!]
\small
    \centering
    \begin{tabular}{l |ccc|c|ccc}
    \toprule
     & \multicolumn{3}{c}{Fine-grained (1-5, Avg.)} & Overall Usefulness &\multicolumn{3}{c}{Relative to Human (\%)} \\
        &  \orgh & \covh & \relh & \useh~(\%)  & Win & Tie & Lose  \\
    \midrule
    GPT4o & 4.63 {\textcolor{osgreen}{\bf (+0.4)}} & 4.06 {\textcolor{orange}{\bf (-0.2)}} & 4.50 {\textcolor{orange}{\bf (-0.1)}} & 69.7 {\textcolor{orange}{\bf (-13.9)}} & 31.9 & 13.8 & \bf 54.2  \\
    \textsc{OS}-8B & 3.82  {\textcolor{orange}{\bf (-0.3)}} & 4.30 {\textcolor{osgreen}{\bf (+0.7)}} & 4.00 {\textcolor{orange}{\bf (-0.4)}} &72.1 {\textcolor{osgreen}{\bf (+8.7)}} & \bf 50.8 & 12.3 & 36.9 \\
    \textsc{OS}-GPT4o & {4.47} {\textcolor{osgreen}{\bf (+0.8)}} & { 4.38} {\textcolor{osgreen}{\bf (+0.9)}}  & { 4.30} {\textcolor{gray}{\bf (0.0)}}  & { 80.0} {\textcolor{osgreen}{\bf (+22.5)}}  & \bf 70.0 & 6.8 & 23.2 \\
    \bottomrule
    \end{tabular}
        \caption{{\bf Human evaluation results}. Fine-grained aspect evaluations are conducted on a five-point scale across four aspects using our detailed instructions and rubrics. Values in parentheses represent the relative performance difference; \textcolor{osgreen}{\bf (+)} indicates that the model shows higher performance, and \textcolor{orange}{\bf (-)} indicates that the human shows higher performance. 
    }
    \label{table:human_evaluation_results}
\end{table}
\subsection{Human Evaluation Results}
\begin{figure*}[t!]
    \centering
    \includegraphics[width=\textwidth]{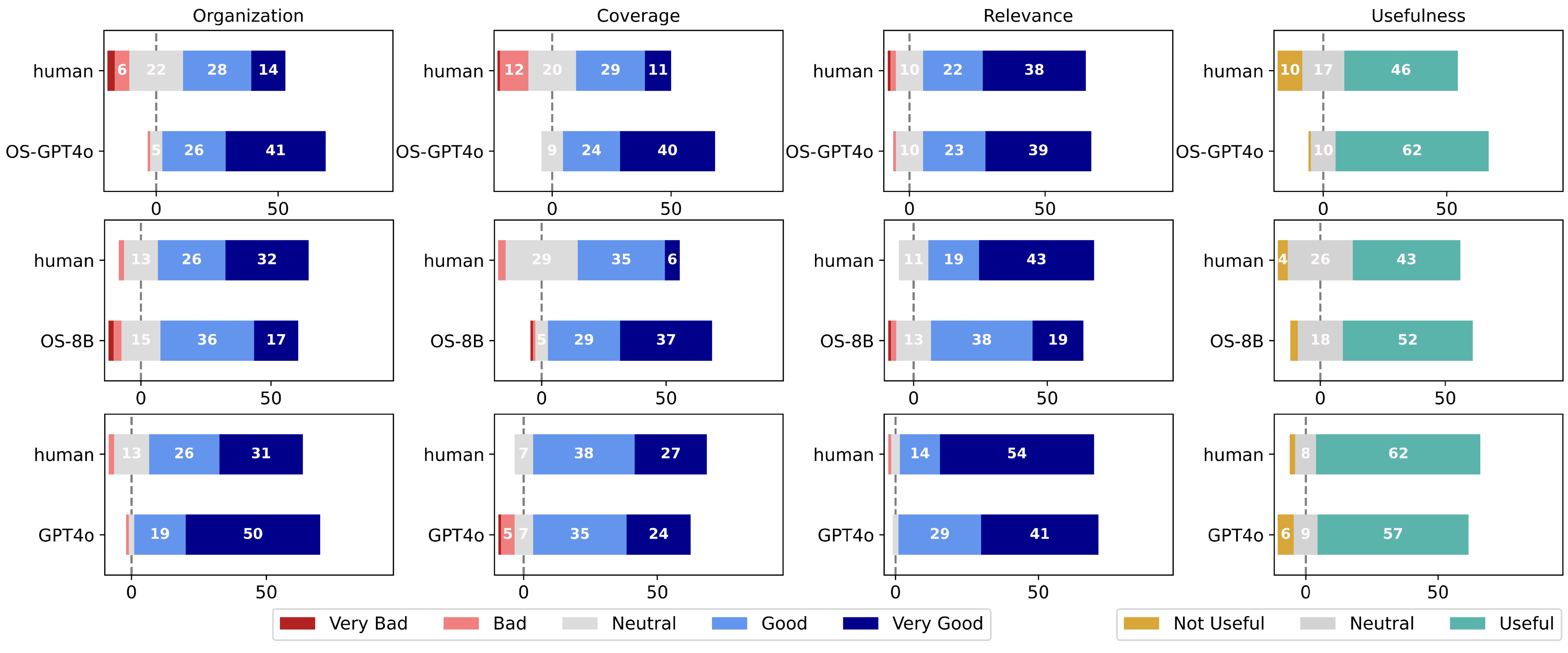}
    \vspace{-4mm}
    \caption{{\bf Fine-grained evaluation results. } Score distributions between 1) GPT4o (top), \model with 8B (middle), \model with GPT4o with Human (bottom). 
    }
    \label{fig:human_eval}
\end{figure*}

\paragraph{Results of human evaluations.}
Table~\ref{table:human_evaluation_results} presents the average scores for each evaluation aspect, alongside the relative win rates against human responses. 
Figure~\ref{fig:human_eval} illustrates the score distributions for Human, GPT4o, and \model with Llama 3 8B and GPT4o. 
Notably, both OS-GPT4o and our OS-8B versions outperform human answers in over 50\% of cases, with their advantage primarily attributed to their ability to provide a greater breadth and depth of information (coverage; \covh). In contrast, GPT4o, which lacks retrieval capabilities, demonstrates significantly limited coverage and wins in fewer than 35\% of cases, with its overall usefulness rated much lower than responses from humans and the other two models. These results highlight that even for state-of-the-art models, synthesizing and answering scientific literature review questions remains a challenging task, consistent with our findings on \data. 
Overall, \model-GPT4o and \model-8B are rated as Useful in 80\% and 72\% of the queries, respectively. 

While \model with a smaller open 8B LM already suppresses human experts, the 8B model's output is judged to be less organized or fluent than the current state-of-the-art private LM-based \model. We found that GPT4o incorporates feedback more effectively, and generally generates longer and more fluent outputs, leading to significantly higher organization scores compared to 8B based \model as well as humans.

\paragraph{Effects of length control on model responses. }
While we found model outputs to often be preferred over human outputs, one potential confounding factor is the significant difference in their output length--\model-GPTo and \model-8B are 2.4 times and 2.0 times longer than human-written answers, respectively, which may affect human judgment~\cite{dubois2024length}. 
To understand the effect of output length, we conducted a controlled experiment: for randomly sampled 50 questions, we generate shorter version of responses for \model-GPT4o, by prompting GPT4o to create summaries of the responses that fall under 300 words. As a result, we collected \model answers that average around 333 words, which is close to the average human answer length. 
We then conduct the same human evaluation and evaluate fine-grained and overall responses. 
On average, the shortened GPT4o scores 4.5 for organization, 4.6 for coverage, and 4.6 for relevance. 
The shortened \model-GPT4o responses are preferred or tied with expert answers in 75\% of the queries. 
The experimental results show that the model's superior performance is not merely due to the increased length of the \model answers. Moreover, human annotators' explanations often mention that both shortened \model and human answers could be improved by incorporating more details, implying that a 300-word restriction may limit the utility of answers.

\paragraph{Analyses on human explanations for pair-wise explanations.}  
We randomly sampled 59 instances with free-form explanations of pairwise preferences and conducted a manual analysis to identify factors that influence overall preferences. Specifically, we examined whether the explanations referenced one or more of the following four categories: organization, relevance, coverage, and citations. While the first three categories align with the fine-grained human evaluation criteria, the citation category also considers the quality of the cited papers (e.g., whether the system includes key representative papers in the field). 
Our analysis revealed that 12\%, 23\%, 29\%, and 9\% of the explanations cited organization, relevance, coverage, and citations, respectively, as key factors in pairwise decisions. This suggests that coverage plays a crucial role in how humans assess the quality of responses, with annotators largely favoring model-generated answers for their greater coverage and depth of information. However, annotators also noted that the citations provided by models could be improved, pointing out that the suggested papers were occasionally outdated or less relevant compared to more representative work. Appendix~\ref{table:explanationst} shows example explanations.

\section{Related Work}
% \akari{need to make sure to cite concurrent work on idea generations, automating data science / machine learning coding ... etc }
\paragraph{Scientific LMs.}
% Previous research has explored the potential of LLMs to accelerate scientific discovery. 
% For example, \textsc{Galactica}~\citep{taylor2022galactica}, a scientific language model trained on vast text data from diverse sources such as academic papers and knowledge bases, aimed to encapsulate extensive scientific knowledge to aid discovery. 
Scientific LMs have spanned various domains, including biomedical \citep{SciFive2021,BioBART2022,BioGPT2022}, medical \citep{singhal2023large,MeLlama2024,PMCLlama2023,Meditron2023}, biomedical \citep{BioMedGPT2023,BioMedGPT2024,BioMistral2024}, geoscience \citep{K22023}, astronomy~\citep{AstroLLaMA2023} and multidisciplinary science \citep{DARWIN2023}, with some models such as SciGLM~\citep{SciGLM2024} and UniSmart~\citep{UniSmart2024} that aim to cover diverse scientific domains in a single model. 
Recently, several works show that powerful general-purpose LLMs can also show strong capabilities in scientific tasks, such as medical question answering~\citep{ai4science2023impact,singhal2023large}, chemistry experimentation~\citep{zheng2023gpt} and applied mechanics~\citep{10.1115/1.4062773}. 
However, the language model's reliance on information memorized within its parameters leads to frequent hallucinations in its output~\citep{li-etal-2024-dawn}. 

\paragraph{LMs to assist scientists.}
Recent studies have also examined LLMs' capabilities to assist scientists in performing a range of scientific procedures, including generating novel research ideas~\citep{baek2024researchagent,Yang2023LargeLM} and automating experimental code generation~\citep{Huang2023MLAgentBenchEL,tian2024scicode}. Our work, however, focuses specifically on benchmarking and developing methods for automating literature reviews and addressing questions related to up-to-date research—tasks that are crucial to, and particularly challenging, for scientific inquiry. 
Several concurrent studies have attempted to build retrieval-augmented pipelines using proprietary LLMs and external APIs (e.g., Semantic Scholar API) for scientific literature review agents~\citep{agarwal2024litllm,skarlinski2024language,wang2024autosurvey}. While these studies and our research all explore the potential of retrieval-augmented LMs in automating literature synthesis, prior works often relied on proprietary, black-box systems and limited evaluations, which commonly entail small-scale human evaluation or simplified setups such as multiple-choice QA. 
In contrast, our work introduces a comprehensive benchmark with automated metrics, involves user studies with experts across three scientific disciplines, and develops new methodologies to train specialized open models. \model significantly outperforms previously introduced systems and shows superiority over human experts in five domains. 

% \paragraph{LLMs that accelerate science.}
% \gncomment{Also, do we only want to focus on LLMs for science? There's also pre-LLM work on various aspects, including survey generation \citep{jha-etal-2015-content}, etc. so maybe that should be fleshed out.}
\paragraph{Benchmarks for scientific literature understanding.}
Several works have developed benchmarks to evaluate models' abilities to understand scientific literature. Prior datasets, such as SciFact~\citep{wadden-etal-2020-fact}, QASPER~\citep{dasigi-etal-2021-dataset}, and QASA~\citep{lee2023qasa}, largely focus on single-paper settings, where the necessary information to answer queries is contained within a single pre-selected paper. However, in real-world scenarios, experts often need to synthesize information from multiple papers to answer questions. To address this gap, \data introduces newly annotated tasks that require reasoning across multiple papers.
There are also scientific summarization tasks, such as Multi-XScience~\citep{lu2020multi}, where models are provided with multiple papers and asked to generate summaries, typically based on the related work sections of those papers. However, in this work, we focus on scenarios where the relevant papers are not specified in advance, making the task more challenging. 
Recently, \citet{xu-etal-2024-kiwi} introduced KIWI, a dataset containing 200 questions and human-verified or edited answers generated by state-of-the-art LLMs, with a focus on the NLP domain. KIWI also provides a set of relevant papers that models must consider. While both KIWI and \data feature multi-paper, information-seeking tasks, \data includes both human-written answers and automatic evaluation pipelines. In contrast, KIWI focuses more on human evaluations, and its reference answers are primarily model-generated.

% \paragraph{General-domain long-form evaluations. }

% Moreover, we conducted human evaluations across four scientific domains, comparing system-generated responses to expert-written answers for open-ended literature review questions.

\section{Conclusion}
In order to further research on LM-based systems that can assist scientific progress, we introduce \model and \data, which can help navigate the complex, ever-growing task of scientific literature review. 
\model, a retrieval-augmented system, leverages open-checkpoint LLMs and trained retrieval models to iteratively refine scientific output, addressing challenges such as hallucinations and citation accuracy. 
\data, a novel large-scale benchmark, provides a standardized way to evaluate literature review automation across multiple scientific domains. 
In evaluations using \data, \model demonstrates substantial improvements, outperforming existing systems, including GPT-4o and the concurrent proprietary system PaperQA2. Our expert evaluation across three scientific disciplines reveals that \data, when paired with fully open-checkpoint models and open-access data stores, generates answers that are more helpful than those produced by expert annotators, who required an hour per annotation. This approach also significantly increases coverage. 
\model using our trained 8B and GPT4o  achieves a 51\% and 70\% win rate against human-generated answers. 
We open-source the \model code, data, model checkpoints, datastores, and \data, along with a public demo, to support and accelerate future research efforts. 

\section*{Limitations}
We highlight several limitations of our work in this section. It is important to note that we do not claim that LM-based systems can fully automate scientific literature synthesis. 
To further advance research in this area, we are releasing both \data and \model to the community. 

\paragraph{Limitations of \data.}
There are several limitations to \data. First, due to the cost and time required to engage expert annotators—individuals with either a Ph.D. or are currently pursuing one in relevant fields—the evaluation dataset with human-written answers is relatively small (e.g., 110 for CS-LFQA and 108 for expert-written answers). This limited dataset may introduce statistical variance and potential biases stemming from the specific expertise of the annotators. To support future research in expanding the size and scope of \data, we open-source our data and annotation pipelines.

Second, our automatic evaluation pipelines may not always perfectly capture the quality of generated content. For example, in \textsc{Scholar-CS}, we combine various components (e.g., length, excerpts, rubric items) using heuristically determined weight terms. {Further, we discovered that often annotators asked for specific kinds of ancillary information in their rubrics---background, elaborations and challenges---even though these aspects might not be strictly required to answer the question. In our experiments, we found that LLMs are proficient at generating the background aspects, which can give them an advantage over systems that directly answer a query but do not satisfy all the constraints of the rubrics. Moreover, future systems could potentially take advantage of the stylistic biases in the rubrics and be prompted to address more rubric elements in a way that does not improve answer quality.} Although we carefully analyzed the correlation between final scores and human expert evaluations, there is still room for improvement in refining which aspects should be emphasized and how these scores should be aggregated. Additionally, our evaluations of citation precision and recall are conducted at the sentence level, but we found that some sentences without direct citations are often supported by citations in adjacent sentences. As a result, our precision and recall metrics might be overly strict, potentially underestimating the true citation accuracy.  We also note that our annotations were captured at particular times (July 2024 for \textsc{Scholar-CS} and September 2024 for \textsc{Scholar-Multi}), and may not reflect subsequent scientific developments. 
Researchers who use our evaluation benchmark should ignore papers published after these dates for a fair comparison.

Lastly, \data primarily focuses on computer science, biomedicine, and physics, with no instances from social sciences or other engineering and scientific disciplines. We recognize that our findings may not fully generalize to other domains, particularly those with more restricted access to paper data.

\paragraph{Limitations of \model.}
While \model demonstrates strong performance on \data and in human evaluations, as discussed in the relevant sections, our expert annotators identified several limitations. 
Despite these issues, we believe \model remains a valuable tool for supporting human experts. 

First, as highlighted by our expert annotators, \model does not consistently retrieve the most representative or relevant papers for certain queries. Enhancing retrieval methodologies by incorporating additional information, such as citation networks or metadata like publication recency, could significantly improve its performance. 
\model outputs may contain factual inaccuracies or unsupported information, particularly in versions based on our 8B model, which has limited capacity for instruction-following and scientific knowledge.  
Future work can explore training that further improve \model-8B. 
In parallel, although it is competitive, \model-GPT4o relies on the proprietary GPT4o API, which may evolve over time, making exact reproduction of results challenging. 
\model does not use license-protected papers during inference time. 
There are ongoing discussions on how to ensure fair data use in retrieval-augmented LMs, and we leave the exploration of properly incorporating copyright-protected content to future work. 

We encourage future research to address these limitations and continue improving LM-based systems for scientific literature review.

\paragraph{Limitations of our human evaluation process.}
In our human evaluations, annotators performed fine-grained assessments on aspects such as coverage, relevance, organization, and usefulness, while other factors such as citation precision and recall were separately evaluated. As a result, when assessing usefulness or pairwise preferences, annotators may have focused more on the overall quality of writing instead of carefully evaluating factual correctness or citation accuracy. We leave more detailed human analysis on citation accuracy, validity, and factuality for future work.

Our evaluations were conducted by 16 Ph.D. students and postdoctoral professionals, and we made an effort to align their expertise with the topics being evaluated. However, since research often demands deep domain knowledge, the annotators may not have captured more nuanced differences for questions outside their immediate areas of expertise. 
Additionally, these evaluations were based on 108 questions spanning three scientific disciplines, meaning that findings may not fully generalize to other fields or domains.

\newpage
\subsubsection*{Author Contribution}
The author contributions are summarized below:  
\begin{itemize}[itemsep=1.6pt,leftmargin=12pt,topsep=1.6pt]
    \item {\bf Project Lead}: Akari Asai
    \item {\bf Project Conception}: Akari Asai, Wen-tau Yih, Pang Wei Koh, Hannaneh Hajishirzi
    \item {\bf \model Development}: Akari Asai, Weijia Shi, Rulin Shao, Jacqueline He
    \item {\bf \model Public Demo Development}: Amanpreet Singh, Joseph Chee Cheng, Akari Asai, Rulin Shao, Doug Downey, Matt Latzke
    \item {\bf peS2o Construction}: Luca Soldaini, Kyle Lo
    \item {\bf Datastore (peS2o Index) Construction}: Rulin Shao, Jacqueline He, Akari Asai
    \item {\bf Legal Discussions on Paper Licensing}: Kyle Lo, Luca Soldaini, Doug Downey, Pang Wei Koh, Amanpreet Singh, Akari Asai
    \item {\bf \model-LM Training}: Akari Asai, Weijia Shi
    \item {\bf \model-Retrievers Training and Evaluation}: Akari Asai, Jacqueline He, Rulin Shao
    \item {\bf \data Design and Conception}: Akari Asai, Pang Wei Koh, David Wadden, Doug Downey, Kyle Lo, Weijia Shi, Amanpreet Singh, Sergey Feldman, Dan Weld
    \item {\bf \data Collections (Single-paper Tasks)}: Akari Asai
    \item {\bf \data Evaluation Pipeline Design and Development}: Akari Asai
    \item {\bf \textsc{ScholarQA}-CS Collection and Evaluation}: Doug Downey, Amanpreet Singh, Sergey Feldman, Dan Weld, Mike D’arcy
     \item {\bf \textsc{ScholarQA}-Multi Collection}: Akari Asai, Minyang Tian,  Rulin Shao, Jacqueline He, Weijia Shi, Pan Ji, Shengyan Liu, Hao Tong, Bohao Wu, Yanyu Xiong
    \item {\bf \textsc{ScholarQA}-Neuro, Bio Collection}: Doug Downey
    \item {\bf Results and Codebases}: Akari Asai, Jacqueline He, Rulin Shao, Weijia Shi, Amanpreet Singh
    \item {\bf Human Evaluation Design}: Akari Asai, Pang Wei Koh, Graham Neubig
    \item {\bf Human Evaluation Interface Development and Supervision}: Akari Asai, Minyang Tian
    \item {\bf Manuscript Writing}: Akari Asai, Jacqueline He, Doug Downey, Amanpreet Singh, Kyle Lo, Pang Wei Koh
    \item {\bf \model Public Demo Testing}: Everyone
    \item {\bf Manuscript Editing}: Everyone
    \item {\bf Advisory}: Pang Wei Koh, Hannaneh Hajishirzi, Doug Downey, Wen-tau Yih, Graham Neubig, Dan Weld, Luke Zettlemoyer 
\end{itemize}

\subsubsection*{Acknowledgments}
We thank our expert annotators, for their help curating high-quality data, and Jenna Sparks at the Ai2 Annotation team for managing and supervising data collection process. 
 We thank Yizhong Wang for his help in developing the human evaluation interface; Hamish Ivison for providing an earlier version of the Tulu v3 instruction tuning data we used for \model 8B training; and Seungone Kim for his help on Prometheus evaluations. We thank Jena Hwang for analyzing limitations of our evaluation data.
 For assistance with the public demo, we thank Chloe Anastasiades, Crystal Nam, Sophie Lebrecht, Taira Anderson, and Will Smith. 
We thank Fangyuan Xu, Eunsol Choi, Aran Komatsuzaki, Sean Welleck, Xiang Yue, Tong Chen, Vijay Viswanathan, Shannon Shen and the members of H2lab and Neulab students for fruitful discussions on this project and feedback on our human evaluation experiments. 
PWK is supported by the Singapore National Research Foundation and the National AI Group in the Singapore Ministry of Digital Development and Information under the AI Visiting Professorship Programme (award number AIVP-2024-001). 
This work partially done while AA is part of the UW-Meta AI Mentorship program.

\bibliography{iclr2025_conference}
\bibliographystyle{iclr2025_conference}

\newpage
\appendix
\section*{Appendix}
\startcontents[sections]
\printcontents[sections]{l}{1}{\setcounter{tocdepth}{2}}
\newpage

\section{Released Artifacts}
We release a set of artifacts to facilitate future research:
% \akari{add all URLs}
\renewcommand{\arraystretch}{1.2}
\begin{center}
\begin{tabular}{crl}
 \textbf{Demo} & \github & \href{https://openscholar.allen.ai/}{\path{openscholar.allen.ai/}}\\
 \textbf{OpenScholar} & \github & \href{https://github.com/AkariAsai/OpenScholar}{\path{github.com/AkariAsai/OpenScholar}}\\
 \textbf{ScholarBenchQA} & \github & \href{https://github.com/AkariAsai/ScholarBench}{\path{github.com/AkariAsai/ScholarBench}}\\
 \textbf{OpenScholar-8B LM} & \huggingface & \href{https://huggingface.co/OpenScholar/OpenScholar_Llama-3.1-8B}{\path{OpenScholar/OpenScholar_Llama-3.1-8B}} \\
 \textbf{OpenScholar-Retriever} & \huggingface & \href{https://huggingface.co/OpenScholar/OpenScholar_Retriever}{\path{OpenScholar/OpenScholar_Retriever}} \\
 \textbf{OpenScholar-Reranker} & \huggingface & \href{https://huggingface.co/OpenScholar/OpenScholar_Reranker}{\path{OpenScholar/OpenScholar_Reranker}} \\
  \textbf{OpenScholar-DataStore-V2} & \huggingface & \href{https://huggingface.co/datasets/OpenScholar/OpenScholar-DataStore-V2}{\path{OpenScholar/OpenScholar-DataStore-V2}} \\
  \textbf{OpenScholar-DataStore-V3} & \huggingface & \href{https://huggingface.co/datasets/OpenScholar/OpenScholar-DataStore-V3}{\path{OpenScholar/OpenScholar-DataStore-V3}} \\
  \textbf{ScholarBench (Data)} & \huggingface & \href{https://huggingface.co/OpenScholar/ScholarBench}{\path{OpenScholar/ScholarBench}} \\
    \textbf{OpenScholar (Training Data)} & \huggingface & \href{https://huggingface.co/datasets/OpenScholar/OS_Train_Data}{\path{OpenScholar/OS_Train_Data}} \\
 \textbf{Expert Evaluation} & \github & \href{https://github.com/AkariAsai/OpenScholar_ExpertEval}{\path{AkariAsai/OpenScholar_ExpertEval}}\\
 % \textbf{ScholarBench} & \github & \href{https://github.com/XXX/XXX}{\path{path/to/bla}} & \github & \href{https://github.com/XXX/XXX}{\path{path/to/bla}}\\
 % \textbf{Index} & \github & \href{https://github.com/XXX/XXX}{\path{path/to/bla}} & \github & \href{https://github.com/XXX/XXX}{\path{path/to/bla}}\\
 % \textbf{Demo} & \github & \href{https://github.com/XXX/XXX}{\path{path/to/bla}} & \github & \href{https://github.com/XXX/XXX}{\path{path/to/bla}}\\
\end{tabular}
\end{center}

\section{More Details on \data}

\subsection{Goal of \data}
There are two key principles to \data: it should serve as a realistic benchmark for literature review, and as a reproducible, mutifaceted evaluation pipeline. 

\paragraph{Realistic benchmark for literature review (\Sref{subsec:data_curation}).} 
\data integrates tasks from two key sources: (i) curated and adapted existing datasets related to literature synthesis tasks and annotated by scientists, and (ii) four new datasets, annotated by Ph.D. experts from four scientific domains that reflect realistic literature review scenarios, such as information synthesis from multiple papers. 
The tasks in \data require different output formats and disciplines. 

For multi-paper tasks, we instruct our expert annotators to formulate information-seeking questions—questions they are genuinely interested in finding answers to, rather than questions they already know the answers to or that could be answered using small text chunks from a single paper~\citep{asai-choi-2021-challenges, choi-etal-2018-quac}. 
We found this approach crucial for curating realistic questions that real-world scientists might ask. These questions are typically more detailed, contextualized (e.g., ``I am planning to generate synthetic training data using GPT4o and filter noisy data using GPT4o, but I'm concerned that GPT4o is not filtering models in this case, as it may favor its own generation.''), and require nuanced, long-form answers rather than simple yes/no or multiple-choice responses. 
We collected human-written answers or rubrics to ensure reliable evaluation, rather than relying on answers generated by state-of-the-art proprietary LMs, as done in \citet{xu-etal-2024-kiwi, malaviya2023expertqa}. While these proprietary models are powerful, they still exhibit limitations, such as hallucinations from lacking domain knowledge, biases, and rapid information changes, making them unsuitable for consistent evaluation with newer models. Additionally, using model-generated answers as references can unfairly favor models from the same family, introducing possible evaluation biases~\cite{panickssery2024llm}. To avoid these issues, we curated expert-written answers for \textsc{ScholarQA-CS} and \textsc{ScholarQA-Multi}.   

\paragraph{Reproducible multi-face evaluation pipelines (\Sref{subsec:automatic_eval}).}  
Due to low correlation between conventional similarity-based metrics such as ROUGE~\citep{xu2023critical,malaviya2023expertqa} and human judgments, evaluations of long-form generation tasks in expert domains have typically relied on small- to medium-scale expert annotations~\citep{zheng2024openresearcher,singhal2023large,si2024can}. While expert human evaluation is valuable (as we detail in \Sref{sec:human_eval}), it requires significant costs for hiring annotators and is hard to reproduce. 
To address these limitations, we introduce automated evaluation pipelines that comprehensively assess the quality of long-form generation outputs from important aspects such as citation correctness or coverage. 

\subsection{Data Curation Details}
\label{app_sec:annotation_details}

\subsubsection{Details of Modification of Single-paper Tasks }
\paragraph{SciFact.}
\textsc{SciFact}~\citep{wadden-etal-2020-fact} is a dataset of 1.4K expert-written scientific claims in the biomedical domain, paired with evidence-based abstracts annotated with labels and rationales. The original task involves three subtasks---paragraph selection, sentence selection, and label prediction---based on a collection of 5,000 abstracts. However, we reformulate this as an open-retrieval label prediction task, where the model is given only a query and must predict the label from a larger corpus of 40 million passages. 
We exclude queries labeled as \texttt{not enough information} and retain only instances labeled as either \texttt{supports} (true) or \texttt{contradicts} (false).

\paragraph{PubMedQA.}
We leverage \textsc{PubMedQA}~\citep{jin-etal-2019-pubmedqa}, which has 1k expert-annotated (yes/no/maybe) QA data on PubMed paper abstracts. 
Similarly to \textsc{SciFact}, we keep instances with yes or no labels, and discard the original abstract passage to formulate the task as an open setup. 

\paragraph{QASA.}
\textsc{QASA}~\citep{lee2023qasa} is a single paper QA dataset consisting of 1,798
novel question answering pairs that require reasoning over scientific articles in AI and ML. We evaluate the model's ability to sufficiently answer a detailed question about the target paper. While they provide three subtasks (answer selection, rational generation and answer compositions) and the end-to-end full-stack QA, we evaluate the models' performance based on full-stack QA.

\subsubsection{Details of Data Collections of Multi-paper Tasks }

\paragraph{Recruiting annotators.} 
For data curation, we recruit expert annotators through UpWork and inter-institution channels, ensuring that they meet the following criteria: (1) they hold a Ph.D. or are enrolled in a relevant Ph.D. program, (2) they have over three years of research experience in the field, and (3) they have published papers in the target areas. In total, we recruited over 20 annotators, including Ph.D. students, postdoctoral fellows, professors, and research scientists, across various multi-paper subsets in the target domains. 

\paragraph{Annotation instructions.} 
{\autoref{tab:annotator instructions_cs} outlines our annotations instructions to collect rubrics for \textsc{ScholarQA-CS}}, and 
Table~\ref{tab:annotator instructions_biomed} shows annotation instructions given to the annotators for \textsc{ScholarQA-Bio} and \textsc{ScholarQA-Neuro}.
Table~\ref{tab:answer_annotation_instructions} shows the instructions given to annotators for \textsc{ScholarQA-Multi}.

\begin{table*}[h!]
\begin{tcolorbox}
{\bf Instructions}\\ 

This project is aimed at building a dataset that can be used to evaluate how well large language models can answer scientific questions.  Specifically, for a set of complex scientific questions that require multiple documents to answer, you’re asked to provide the points that are important to include in an answer, along with supporting text from the scientific literature and other sources.  We will release your output to the community in the form of a dataset that allows researchers to assess how well AI systems can answer complex scientific questions.
\\
\\
For each question, you are given two documents - 
\begin{itemize}
    \item A key ingredients document that you will fill-in with components that you feel are necessary to include in a correct answer to the question.
Key ingredients are short (1-4 sentences) statements of an important item to include in a correct answer, such as ``near the beginning, the answer should should briefly define state space models and transformers and detail their technical differences.''  
    \item (do not read right away) A sources document that you will use for reference, giving the output for the question from different online services
\end{itemize}
Perform the following steps:

\begin{enumerate}
    \item First, BEFORE reading the existing text in the sources document, based on your knowledge of the question and about 40 minutes of literature review using Google Scholar or Semantic Scholar (but NOT any online large language model services like ChatGPT or similar), compose a set of key ingredients of the answer and supporting quotes from the documents.  Starting from the corresponding key ingredients file for the question, add your key ingredient to the list along with zero or more snippets of text from the documents under ``supporting quotes.''  Each snippet should be less than 150 words.  After each quote, paste a link from the source (if you quote a document multiple times, paste a link after each quote).  Your key ingredients should typically have supporting snippets, but may not always if you wish to include an important fact that you know but cannot find support for.
    \item Then, in a total of about 20-50 additional minutes, read through the sources document  for the question (you may have to skim over portions, as some source documents are long), and complete the set of key ingredients in the answer based on the combination of what you’d listed in step 1, and what you learn from the sources.  You should delete or change ingredients you added after step one, if appropriate.  
    
    Note: you can copy and paste text directly from the sources document as Supporting Quotes, and for those quotes you do NOT need to include a link, just paste the relevant portion in quotation marks.
\end{enumerate}

In your key ingredients, you do not need to cover general stylistic points (e.g., that the answer should be well-organized and include citations), those characteristics are already captured in a general rubric we’ve written.  Instead, focus on the specific content needed in the answer, organized as bullet points.  When capturing the ingredients, follow the template of the document to separate them into \textbf{Most important} and \textbf{nice-to-have}.  The ``most important'' items should be ones that, if they were missing, would lead the answer to be misleading or incomplete.  The ``nice to have'' should include other helpful elements, but ones that could potentially be omitted in a short answer.

To help you understand the task, an example of a completed ingredients file, and the corresponding sources file is provided.

\end{tcolorbox}
\caption{\label{tab:annotator instructions_cs} {\bf \textsc{ScholarQA-CS} annotation instructions}: Instructions given to our expert annotators for producing key ingredients for \textsc{ScholarQA-CS} questions.}
\end{table*}

\begin{table*}[h!]
\begin{tcolorbox}
{\bf Instructions}\\ 

In this task, you are asked to generate complex scientific questions that biomedical scholars might reasonably ask about scientific literature.  These should be broad literature review questions that require multiple papers to answer, e.g.: ``What are the various molecular mechanisms through which chronic inflammation contributes to the development of colon cancer?''  This question requires multiple papers to answer, as there are a variety of different studies on such molecular mechanisms.  By contrast, an example of a question we are NOT looking for would be ``What was the efficacy of Pembrolizumab in reducing tumor size in patients with advanced melanoma based on the S1801 clinical trial?'', since a single paper (the results of the clinical trial) holds the answer.

To help ensure we obtain a wide variety of questions, we'd like you to use existing papers as inspiration for your questions.  You can think of it as trying to guess what kinds of literature review questions the authors of the paper may have had while they were performing the research in the paper.  Specifically, follow the following steps:

1. Pick a paper of interest in your domain.  This can be any paper, but review or survey articles can be an especially useful source of inspiration as they cover broad topics encompassing many previous papers.  Also, previous work or future work sections of papers can be good sources. \\
2. Write up to 5-10 literature review questions that are relevant to the paper.  These should be:
(a) Questions that generally require multiple documents to answer; (b) Questions about the published literature—the kinds of questions that a scientist could attempt to answer by searching and reading papers (i.e., they don’t necessarily require new experiments)
\\
We have provided a Google Sheet with three columns, ``question,'' ``inspiring paper,'' and ``estimated number of papers needed to answer (1/10/100/1000+),'' for you to list each of your questions and roughly how many papers you think a person would need to examine in order to answer it accurately.
\\

While it is important to choose papers on topics you are familiar with, please also strive to choose a diverse set of papers on a variety of subtopics, and do not write more than 10 questions about any single paper.  Paste each of your questions and the Semantic Scholar link for the inspiring paper in the spreadsheet (in the unlikely event that the link cannot be found, please paste the bibliographic information for the paper in APA format).

As a concrete example, consider this paper:

\tcblower
Bian, Shuhui, Yicheng Wang, Yuan Zhou, Wendong Wang, Limei Guo, Lu Wen, Wei Fu, Xiaoxia Zhou and Fuchou Tang. ``Integrative single-cell multiomics analyses dissect molecular signatures of intratumoral heterogeneities and differentiation states of human gastric cancer.'' National Science Review 10 (2023): n. pag.

Some questions that you might write for this paper are the following (these particular examples were not written by a domain expert and may have imperfections, but give the idea):

What are the known biomarkers associated with cancer progression, differentiation states, and immune evasion in tumors? 

How have other studies integrated multiomics data to reveal molecular signatures and regulatory mechanisms in cancer?

What are the current capabilities and limitations of single-cell multiomics sequencing in cancer research?

What are the methodologies and best practices for multi-regional sampling in solid tumors?

What are the existing transcriptomic and epigenomic signatures associated with gastric cancer?

\tcblower
You do not need to make any attempt to determine the answers to your questions, or even try to verify whether such answers exist.  It is okay if some of your questions turn out to be unanswerable (for example, for Q1 above, it might be the case that there are no known biomarkers; or for Q2, perhaps no such studies exist).  We want you to choose the most realistic literature review questions you can, and sometimes realistic questions will not have answers in the current literature.
\end{tcolorbox}
\caption{\label{tab:annotator instructions_biomed} {\bf \textsc{ScholarQA-Bio}, \textsc{Neuro} annotation instructions}:Instructions given to our expert annotators for authoring questions in biomedicine. The example questions in this passage were composed with assistance from GPT. }
\end{table*}

\begin{table*}[h!]
\begin{tcolorbox}
{\bf Instructions}\\ 
You will be given a question and your task is to annotate your answers with appropriate citations. You are allowed to use tools like Google Search, Google Scholar, or Semantic Scholar to find sources. However, you may not use any language model-powered systems such as ChatGPT, GPT-4, Claude, Perplexity, or you.com.

{\bf Steps to Follow:}\\
1. Start the Stopwatch: Before you begin reading the question, start a stopwatch to track how long it takes you to complete the task.

2. Answer the Question: After reading the question, write your answer. Ensure that every sentence that requires a citation is followed by a citation number (e.g., [1]).

3. Track Time: Once you have finished writing your answer, stop the stopwatch and note the exact duration it took to complete the task.

{\bf Answer Format:}\\
{\bf Output}: Write your answer under \texttt{Output}. Ensure that all citation-worthy sentences include at least one citation, indicated as [citation number] (e.g., [1]).

{\bf Citations}: For each citation you use (denoted as [i] in your answer), you need to fill out the following information:

\texttt{citation\_i\_title}:  
Provide the title of the paper.

\texttt{citation\_i\_corpus\_id}: Semantic Scholar Corpus ID for the citation.   

\texttt{citation\_i\_text}: Copy the relevant paragraph from the paper that supports your answer. In this context, a ``paragraph'' refers to a set of sentences separated by a new line in the paper. Below, we show an example of paragraphs. If there’s no new line and one paragraph gets really long e.g., over 0.5 pages, then please copy and paste the core sentences including the useful information.

\texttt{citation\_i\_url}: Provide the URL link to the paper (e.g., arXiv abstract page, Semantic Scholar page, PubMed abstract page, etc.).
\end{tcolorbox}
\caption{\label{tab:answer_annotation_instructions} {\bf \textsc{ScholarQAMulti} annotation instructions}: Instructions given to our expert annotators to annotate question and answers. }
\end{table*}

\begin{figure}[t!]
\begin{subfigure}[b]{0.42\textwidth}
    \includegraphics[width=\textwidth]{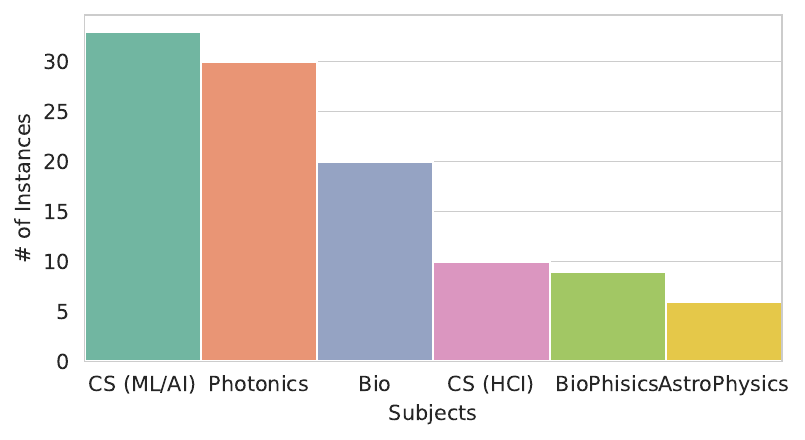}
  \caption{Subject distributions of expert-annotated answers}\label{fig:data_fistributions}
\end{subfigure}%
\hspace{0.3cm}
  \begin{subfigure}[b]{0.55\textwidth}
        \includegraphics[width=\textwidth]{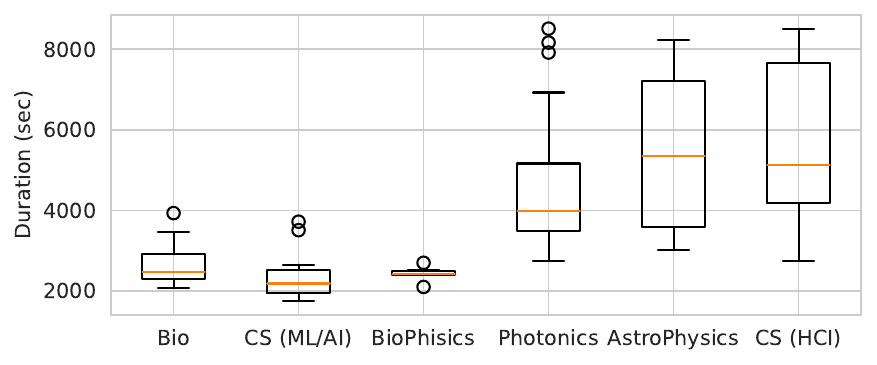}
        \caption{Average per-instance annotation time for each subject (seconds). }
        \label{fig:annotation_time}
  \end{subfigure}%
\caption{{\bf Analysis of human-written answers:} (a) shows the distribution of instances per subject, and (b) shows the average annotation time per instance per subject.  
}\label{fig:analysis_human}
\end{figure}

\paragraph{Statistics of human-written questions and answers in \textsc{ScholarQA-Multi}.}
Figure~\ref{fig:data_fistributions} shows the subject distribution, and Figure~\ref{fig:annotation_time} depicts the average time in seconds that experts spend on annotating each answer across different subjects. 
Annotators on average spend at least 30 minutes per answer, but those from some subjects, such as Physics, took over an hour (e.g., approximately 5,000 seconds) to complete a single annotation. This demonstrates that the task is highly time-consuming and challenging even for domain experts.

\subsection{Evaluation Metrics Details}
\label{app_section:evaluations}

\subsubsection{\textsc{ScholarQA-CS} \acc~Evaluation}
\label{app_sec:details_scholar_cs}
For \textsc{ScholarQA-CS}, we employ expert-annotated rubrics to evaluate the generated answers. 
Each rubric has two criteria---general (accounting for 40\% of the score) and annotation-driven (60\%). 
The general criterion covers the evaluation of length, expertise, citations, and excerpts, while the annotation criterion involves assessing the extent to which each specific key ingredient (and associated quotes) identified by annotators is present in the answer, on a scale of 0 to 1.
{Each must-have ingredient is considered twice as important as a nice-to-have. However, there is no distinction between individual ingredients of the same type.}
Using LLM-as-a-judge~\citep{zheng2023judging}, a score is assigned for each criterion, and a weighted sum is computed across the criterion to obtain a final score.  This score emphasizes the presence of citations and excerpts, such that systems lacking either tend to underperform even when they score relatively high on the remaining criteria, as seen in our experiments. 

\begin{table}[t!]
\small
    \centering
    \begin{tabular}{p{2cm} |p{9cm} | l }
    \toprule
        \textbf{Aspect} & \textbf{Definition}  & \textbf{Instructions}\\
    \midrule
    Organization  & Evaluate if the output is well-organized and logically structured.  & Table~\ref{tab:rubrics_organizations}\\
    Coverage & Evaluate if the output provides sufficient coverage and amount of information. & Table~\ref{tab:rubrics_coverage} \\
    Relevance & Evaluate if the response stay on topic and maintain a clear focus to provide a useful response to the question &  Table~\ref{tab:rubrics_relevance} \\ 
    Overall Usefulness  & Evaluate if the output contains useful information to fulfill the information needs. & Table~\ref{tab:rubrics_overall_usefulness} \\ 
    \bottomrule
    \end{tabular}
        \caption{{\bf Evaluation protocols for writing quality}. We define three fine-grained aspects to be evaluated by both human experts and LLMs. In addition to these, experts are also asked to assess whether the answers are useful.
    }
    \label{table:evaluation_protocols}
\end{table}

\subsubsection{Content Quality and Organization}
An overview of the evaluation aspects is in \Tref{table:evaluation_protocols}.  
We use Relevance, Coverage and Organization for automatic evaluation. We found that existing evaluator LMs struggle with evaluating overall usefulness, and tend to be over-optimistic. 
%\jh{Unfinished sentence}

\paragraph{Evaluation instructions and rubrics.} 
We show annotator rubrics for organization, coverage, relevance and overall usefulness in Tables~\ref{tab:rubrics_organizations},\ref{tab:rubrics_coverage}, \ref{tab:rubrics_relevance}, and \ref{tab:rubrics_overall_usefulness}, respectively.  

\paragraph{Prometheus configuration.} 
For evaluations, we combine Prometheus BGB~\citep{kim2024biggen} (\texttt{prometheus-eval/prometheus-bgb-8x7b-v2.0}) and Prometheus v2~\citep{kim2024prometheus} (\texttt{https://huggingface.co/prometheus-eval/prometheus-8x7b-v2.0}). 
We found that Prometheus BGB generally works well, while on \rel~it sometimes gives scores that are much higher than human assessments, especially for GPT4o. Consequently, we use Prometheus BGB for \org~ and \cov~, and Prometheus v2 for relevance. 
We use human-written answers as gold references, which are shown to improve Prometheus' correlation with human evaluation. We use these models with \texttt{vllm}. 
Following the default setup, we set the maximum number of new tokens to 512, top-$p$ to 0.95, and temperature to 0.01.

\subsubsection{Citation Accuracy}
To evaluate citation accuracy, we use \texttt{osunlp/attrscore-flan-t5-xl}~\citep{yue2023automatic}, which is FLAN-T5-XL trained on a mixture of attribution tasks. We follow citation precision and recall formulation from \cite{gao2023enabling} and compute citation precision and recall at the sentence level. We discard sentences under 50 characters, as these sentences are often paragraph or subsection headers that do not require citations. 
We use the original citation evaluation instructions from \citet{yue2023automatic}, as shown in Table~\ref{tab:citation_prompt}.  

\begin{table*}[h!]
\begin{tcolorbox}[colback=blue!5!white,colframe=blue!75!black]
As an Attribution Validator, your task is to verify whether a given reference can support the given claim. A claim can be either a plain sentence or a question followed by its answer. Specifically, your response should clearly indicate the relationship: Attributable, Contradictory or Extrapolatory. A contradictory error occurs when you can infer that the answer contradicts the fact presented in the context, while an extrapolatory error means that you cannot infer the correctness of the answer based on the information provided in the context.

{\bf Claim:} 

{\bf Reference: }
\end{tcolorbox}
\caption{\label{tab:citation_prompt}{\bf Evaluation instruction for citation evaluation.} We prompt an attribution LM if the claim is supported by the provided reference or not. The instruction is adapted from ~\citet{yue2023automatic}.  }
\end{table*}

\section{More Details on \model}
\label{app_sec:experimental_details}

\subsection{Training a scientific bi-encoder $\theta_{\rm bi}$}
\label{app_sec:bi_encoder_training}

For $\theta_\text{bi}$, we follow the unsupervised training methodology from \cite{izacard2021towards}, and continually pre-train the Contriever bi-encoder on a mixture of peS2o version 2, CCNews, and Proofpile2~\citep{azerbayev2024llemma} data for 500k steps, using a batch size of 4,096 and a learning rate of 0.00005. We initialize the model checkpoint from Contriever. 
\subsection{Training a scientific cross-encoder $\theta_{\rm cross}$}
\label{app_sec:raranker_training}
Paragraphs that score above 3 are labeled as positive, while those that score below 3 are labeled negative. Finally, we select the top $N$ passages based on this process, passing the top paragraphs $\mathbf{P}$ to the generator LM.  
For $\theta_\text{cross}$, we fine-tune the BGE-large reranker for 5 epochs on our newly created training data for five epochs, using a learning rate of 6e-5. 

\subsection{Training Details of Generators $\mathcal{G}$}
\label{app_sec:training_details}
\begin{figure*}[t!]
    \centering
    \includegraphics[width=0.5\textwidth]{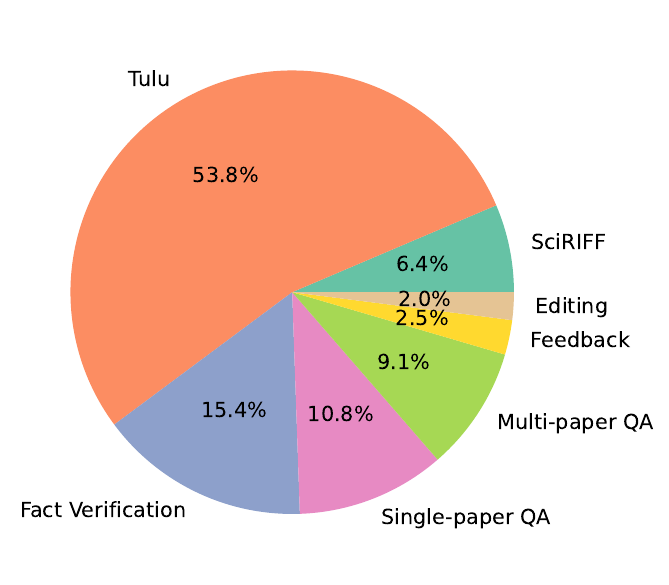}
    \caption{{\bf Generator training data distribution.} We mix diverse training data to train our 8B LM.  
    }
\label{figu:distribution_training_data}
\end{figure*}

\paragraph{Training data statistics.}
Figure~\ref{figu:distribution_training_data} shows the training data distribution. ``Tulu'' indicates general-domain instruction-tuning data from \cite{ivison2023camels} and SciRIFF~\citep{wadden2024sciriff} indicates task-specific data from SciRIFF. For both datasets, we ensure that we do not include any data used for evaluation, namely PubMedQA, SciFact and QASA. In total, this leads to 130,135 training instances. 
% \begin{wrapfigure}{r}{0.5\textwidth}
%   \begin{center}
%     \includegraphics[width=0.5\textwidth]{figs/data_distribution.pdf}
%   \end{center}
%   \caption{Generator training data distribution.}\label{figu:distribution_training_data}
% \end{wrapfigure}

\paragraph{Training hyperparameters.}
For OS-8B, we initialize model checkpoints from Llama 3.1 8B  Instruct~\citep{dubey2024llama}. We use \texttt{torchtune}~\footnote{\url{https://github.com/pytorch/torchtune}} for training. 

We train both models for two epochs using learning rates of 5e-6 and 1e-4, respectively, with a maximum context length of 10k and batch size of 1 and gradient accumulation step of 2 with bf16. 
We use AdamW~\citep{loshchilov2018decoupled} as an optimizer.

\begin{table*}[h!]
\begin{tcolorbox}[colback=blue!5!white,colframe=blue!75!black]
{\bf Instructions}\\ 
Evaluate if the answer is well-organized and logically structured. An acceptable response should: 1. be clearly structured, grouping related points together for a logical flow., and 2. be coherent, without any contradictions or unnecessary repetition.
\tcblower
{\bf Score 1: Poor Organization}\\
- The response is disorganized, with no clear structure. \\
- Points are scattered without any logical order, making it difficult to follow. \\
- The text lacks coherence, and there are contradictions or irrelevant repetitions throughout. 

{\bf Score 2: Basic Organization}\\
- The response has some organization, but the structure is inconsistent. 
- There are occasional lapses in coherence, with minor contradictions or repetitive statements that disrupt the overall clarity.

{\bf Score 3: Moderate Organization}\\
- The response is generally well-organized, with a clear structure that is mostly maintained. 
- Points are grouped logically, though there may be some minor lapses in flow or coherence. The answers have several clear paragraphs, each of which clearly discuss some core points. 
- The text is mostly clear, with only occasional repetition or slight contradictions.

{\bf Score 4: Strong Organization}\\
- The response is well-organized, with a clear and logical structure that is followed consistently. 
- Points are effectively grouped, and the flow is smooth. 
- There may be minor lapses in coherence, but overall, the response is clear and easy to follow, with minimal repetition or contradictions.

{\bf Score 5: Exceptional Organization}\\
- The response is exceptionally well-organized, with a flawless logical structure. 
- Points are grouped perfectly, and the flow between them is seamless. 
- The text is coherent throughout, with no contradictions or unnecessary repetition, making the argument clear and compelling.
\end{tcolorbox}
\caption{\label{tab:rubrics_organizations} {\bf Evaluation rubrics for organization.} The evaluation instructions and five-point rubric for assessing organization.  }
\end{table*}

\begin{table*}[h!]
\begin{tcolorbox}[colback=blue!5!white,colframe=blue!75!black]
{\bf Instructions}\\ 
Evaluate if the output provides sufficient coverage and amount of information. The output should: 1. (coverage) Offer a comprehensive review of the area of interest, citing a diverse range of representative papers and discussing a variety of sources, not just a few (1-2 papers) 2. (depth) Provide enough relevant information to understand each discussion point.  
\tcblower
{\bf Score 1: }\\
- {\bf Severely Lacking Coverage}: The output lacks coverage of several core lines of research or focuses predominantly on a single line of work, missing a holistic view of the area.
- {\bf Greatly Limited Depth of Information}: The output either lacks essential details needed to fully understand the topic (e.g., definitions of methods, relationships between methods). 

{\bf Score 2:}\\
- {\bf Partial Coverage}: The output covers some key aspects of the area but misses significant lines of research or focuses too narrowly on a few sources. It lacks a well-rounded view and fails to adequately represent the diversity of work in the field.\\
- {\bf Limited Amount of Information}: The response provides some relevant information but leaves out important details that would help in understanding the topic fully.

{\bf Score 3: }\\
- {\bf Acceptable Coverage / Overview}: The output discusses several representative works and provides a satisfactory overview of the area. However, including more papers or discussion points could enhance the answer significantly. It addresses the core aspects of the question but may miss some details.\\
- {\bf Acceptable Amount of Relevant Information}: The output provides a reasonable amount of relevant information, though it may lack some helpful details. 

{\bf Score 4: }\\
- {\bf Good Coverage}: The output offers good coverage of the area, discussing a variety of representative papers and sources. While it provides a broad overview, it may miss a few minor areas or additional papers that could enhance the comprehensiveness.\\
- {\bf Mostly Sufficient Amount of Information}: The response includes most of the necessary and relevant information to understand the topic. It avoids excessive irrelevant details, but a few points might benefit from deeper exploration or more specific examples.

{\bf Score 5: }\\
- {\bf Comprehensive Coverage}: The answer covers a diverse range of papers and viewpoints, offering a thorough overview of the area. It includes additional important discussion points not explicitly mentioned in the original question.\\
- {\bf Necessary and Sufficient Amount of Information}: The response provides all necessary and sufficient information. 
\end{tcolorbox}
\caption{\label{tab:rubrics_coverage} {\bf Evaluation rubrics for coverage.} The evaluation instructions and five-point rubric for assessing coverage.}
\end{table*}

\begin{table*}[h!]
\begin{tcolorbox}[colback=blue!5!white,colframe=blue!75!black]
{\bf Instructions}\\ 
Evaluate if the response stays on topic and maintain a clear focus to provide a useful response to the question. Specifically, the output should: 1. fully address the core points of the original question and satisfy your information needs if they are factual 2. not contain a lot of minor information that is not related to the original question. 
\tcblower
{\bf Score 1: Off-topic}\\
- The content significantly deviates from the original question, making it difficult to discern its relevance or distract the user. 

{\bf Score 2: Frequently off-topic with limited focus}\\
- The response addresses the question to some extent but often strays off-topic. 
- There are several sections with irrelevant information or tangential points that do not contribute to answering the main question. 
- These distractions make it difficult to maintain a clear focus and reduce the overall usefulness of the response.

{\bf Score: 3: Somewhat on topic but with several digressions or irrelevant information}\\
- While the response still centers around the original question, there are frequent deviations that distract from the main question or redundant information.

{\bf Score 4: Mostly on-topic with minor deviations}\\
- The response stays largely on-topic, with a clear focus on addressing the question. However, there may be a few minor digressions or slightly irrelevant details that momentarily detract from the main focus.
- These deviations are infrequent and do not significantly undermine the overall clarity or usefulness of the response.
.

{\bf Score 5: Focused and entirely on topic }\\
- The response remains tightly centered on the subject matter with enough depth and coverage of the core information, with every piece of information contributing directly to a comprehensive understanding of the topic.
\end{tcolorbox}
\caption{\label{tab:rubrics_relevance} {\bf Evaluation rubrics for relevance.} The evaluation instructions and five-point rubric for assessing relevance. }
\end{table*}

\begin{table*}[h!]
\begin{tcolorbox}
{\bf Instructions}\\ 
Do you think the provided answers are overall helpful and assist your literature review? In particular:
\tcblower
{\bf Score 1: Unhelpful}\\
- The response does not answer the question or provide rather confusing information. 
- It does not serve as a useful starting point for learning or writing about the area or understanding the literature.

{\bf Score 2: Better than searching by myself from scratch but limited utility  }\\
- The response provides at least one helpful paper that I can take a closer look at. 
- However, overall the discussion in the answer is somewhat irrelevant and does not provide helpful information. 

{\bf Score 3: Provides some useful discussions and papers, although I need to read individual papers by myself}\\
- The response is generally helpful and provides at least 2-3 helpful papers that I wasn’t aware of. I can start looking at some of the suggested papers. 
- It provides some good overview of the areas and multiple viewpoints, which I can dive into. 
- Yet, I may need to verify some details or consult additional core research papers to ensure completeness. 

{\bf Score 4: Useful. I may try to verify some details, but overall the output gives me a great summary of the area of my interest.}\\
- The answer gives me a great set of papers highly related to my question. 
- The answer provides a great overview and detailed information.
- I may need to check some minor details from 1-2 specific papers included in the answers, but overall the provided answer is useful with little need of additional editing. 

{\bf Score 5: Super Useful. I can fully trust the answer}\\
- The response provides a comprehensive overview of the area, and sufficiently answers my question. 
- I don’t think I need to search additional papers or check details by myself. 

\end{tcolorbox}
\caption{\label{tab:rubrics_overall_usefulness} {\bf Evaluation rubrics for overall helpfulness.} The evaluation instructions and five-point rubric for assessing overall usefulness. }
\end{table*}

\begin{figure*}[t!]
    \centering
    \includegraphics[width=0.98\textwidth]{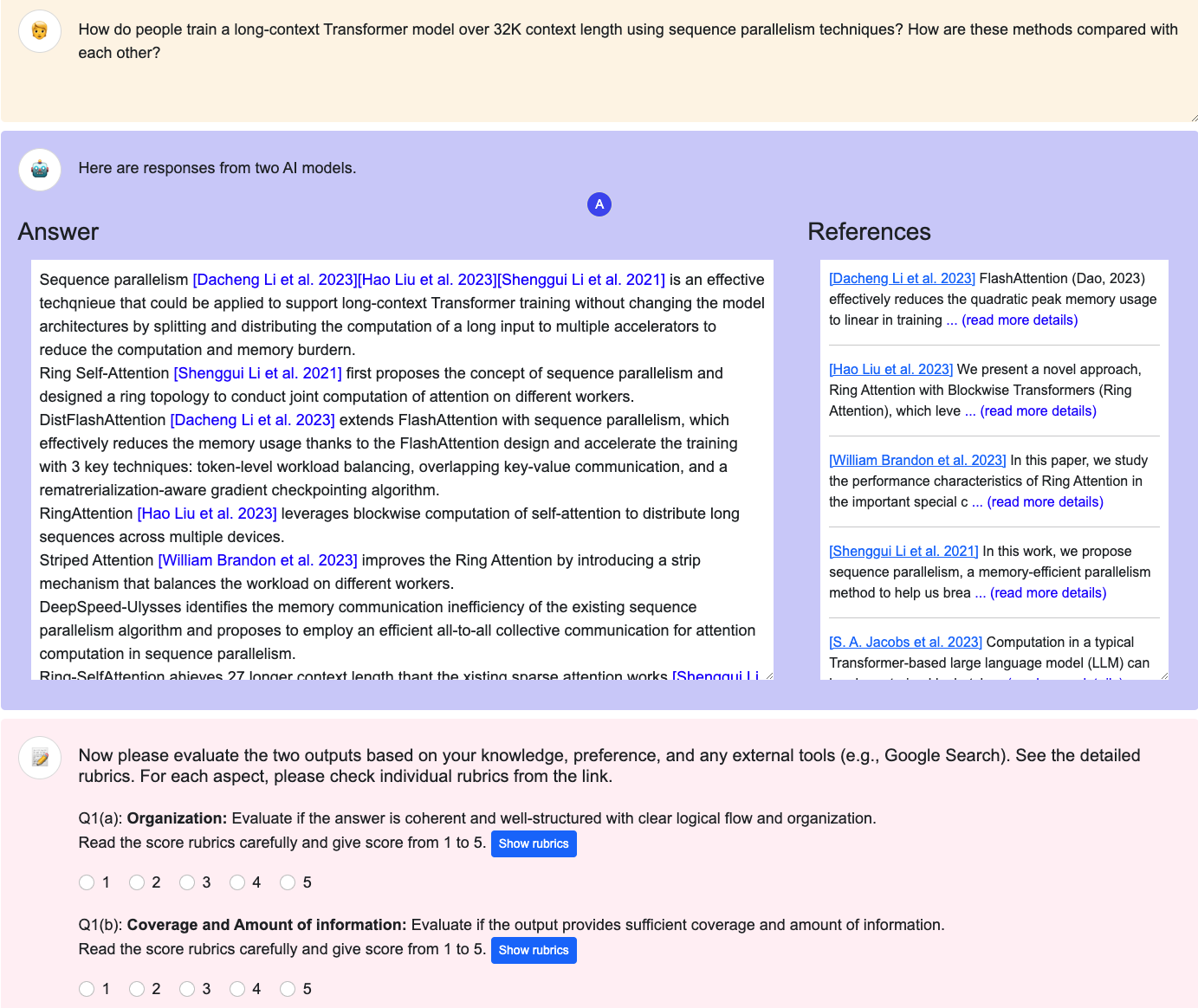}
    \caption{Human evaluation annotation interface.}
    \label{fig:annotation_interface_1}
\end{figure*}

\begin{figure*}[t!]
    \centering
    \includegraphics[width=0.98\textwidth]{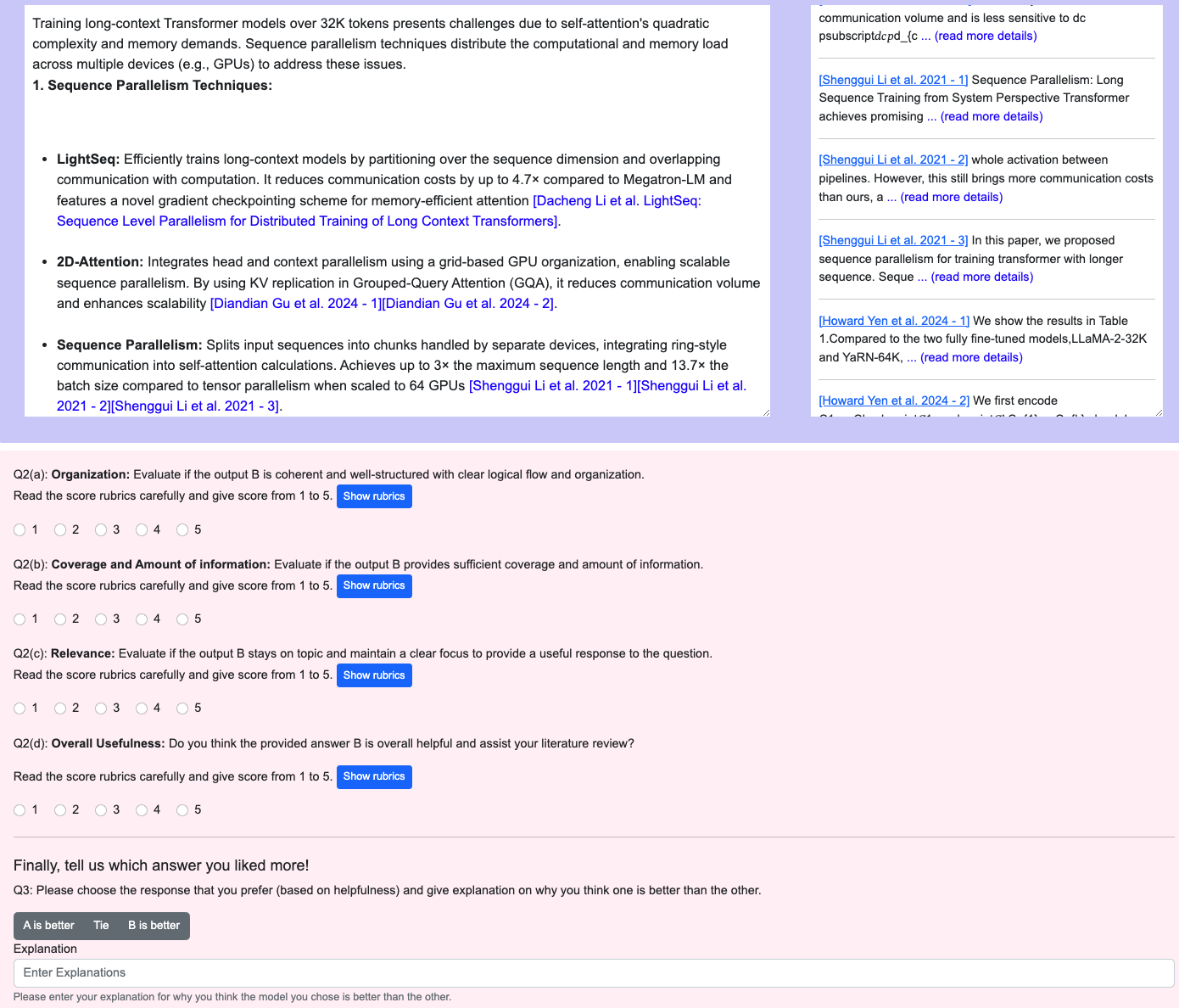}
    \caption{Human evaluation annotation interface.}
    \label{fig:annotation_interface_2}
\end{figure*}

% \section{More Results}
% \subsection{Full Results on \data}

\section{More Analysis}
\subsection{Comparison of peS2o v2 and v3}
\label{app_sec:pes2o_ds_analysis}
\begin{figure*}[t!]
    \centering
    \includegraphics[width=0.95\textwidth]{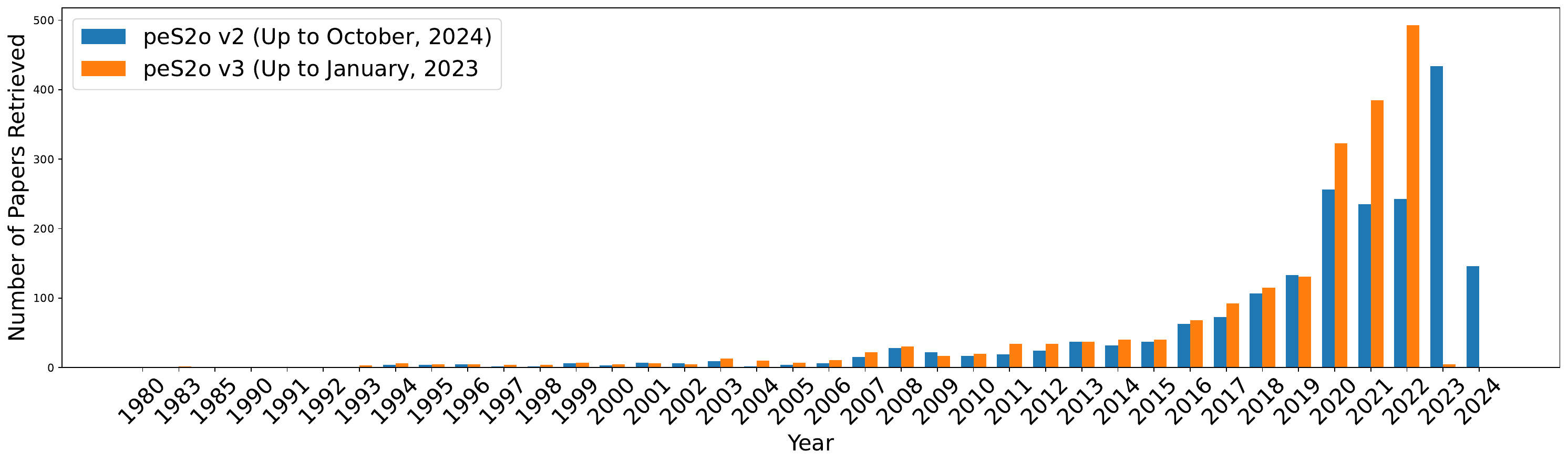}
    \caption{{\bf Distributions of the paper publication years of top 20 retrieved papers for \textsc{ScholarQA-CS}. } This figure shows that by updating the datastore from peS2o v2 to the more recent peS2o v3, which includes papers up till October 2024, our dense retrieval model can successfully retrieve more recent papers. 
    }
    \label{fig:cited_papers_pes2o}
\end{figure*}

% \subsection{Datastore Analysis}
% \label{app_sec:pes2o_ds_analysis} 
In this section, we conduct a brief analysis on datastores used in \model. 
Figure~\ref{fig:cited_papers_pes2o} shows the distributions of top 20 retrieved papers from the two datastores, peS2o v2 and v3 for the same \textsc{ScholarQA-CS} queries. Note that our retrieval models are trained on peS2o v2 data. Although our model is not directly trained on the new version 3 datastore, we found that it can constantly retrieve relevant papers from the newer datastore at test time, resulting in many papers from 2023 - 2024.

\subsection{More analysis on Expert Evaluation}
\label{app_sec:human_evaluation}

\paragraph{Automatic evaluations of human and model-generated answers.} 
\begin{table}[h]
\small
    \centering
    \begin{tabular}{l |cccccccc}
    \toprule
        &  Length & Inf. time (min) &   \# of Citations & Prec. & Rec. & Org. & Cov. & Rel.   \\
    \midrule
 Human & 289.5 & 56.8 & 6.0 & 44.4  & 41.5  & -- & -- & --   \\ \hdashline
\textsc{OS}-GPT4o & 716.3 & 1.5 & 9.8 & 38.0 & 37.1&  4.63 & 4.50 & 4.23  \\ 
\textsc{OS}-8B & 578.6 &  0.9 & 21.6 & 42.5 & 43.2  &   3.92 & 4.44 & 4.02  \\\hdashline
    GPT4o & 281.3 & 0.5& 3.8 &  0.8 & 0.6 &  4.06 & 3.94 & 4.21 \\
    PaperQA2 & 166.9 & 0.7 & 3.14  & 53.1 & 42.2 & 3.67 &  3.46 & 4.20 \\ 
    \bottomrule
    \end{tabular}
        \caption{{\bf Human-written answer stats}. Models tend to generate longer responses, citing more papers than humans. For reference, we run GPT4o without retrieval on the human evaluation queries. 
    }
    \label{table:result_human_data}
\end{table}
Table~\ref{table:result_human_data} provides basic statistics for human and \model-generated responses, detailing the average length, the number of cited articles, and the scores for each evaluation metric. We observed that human answers are generally shorter and reference fewer papers compared to model-generated outputs. 
While a concurrent study normalizes human-written answers using an LM to control for confounding factors~\citep{si2024can}, we retain the original human and model answers for our human evaluation and conduct extensive analysis to understand how these differences influence human preferences. 

We ran our automated evaluation pipelines, which include assessments of citation precision, recall, and writing quality, on both human- and model-generated answers. Our findings reveal that OS-8B frequently matches or even surpasses expert-level citation precision and recall, whereas OS-GPT4o performs slightly worse in terms of citation accuracy. Although PaperQA2 demonstrates higher citation precision compared to OS-8B or human experts, its answers are often brief and cite fewer papers, leading to limited coverage.

\paragraph{Qualitative analyses on experts' explanations of pair-wise preference. }
Table~\ref{table:explanationst} shows experts' explanations on pair-wise preferences.

\begin{table}[t]
\small
    \centering
        \resizebox{\textwidth}{!}{
    \begin{tabular}{ll|p{12cm}}
    \toprule
        \textbf{Category} & \textbf{\%} & \textbf{Examples} \\ \midrule
    Organization & 12\% &  Lots of redundancy and repeat in B \\
    & & Although B has lots of overlaps, but B covers more aspects in grating behavior \\\midrule
    Relevance & 23\%  & Response spends half of the text talking about the performance comparison between fine-tuning and RAG, instead of the question ``fine-tuning on RAG''. \\ 
    & & A contains some unnecessary details that don't help understand the topic. \\ 
    & & Response A has a poor organization and a little deviation from the topic when it talks about privacy and safety of RLHF. \\ 
    & & Although response A shows a smaller coverage, it is slightly better than B due to detailed description of the techniques in generating reasoning data. \\ 
    & & B focuses on numerical methods and gives more approachable models.  \\\midrule
    Coverage &  29\% & A is clearly more helpful as it provides a clean organization and cover more papers. \\
    & & Although response A shows a smaller coverage, it is slightly better than B due to detailed description of the techniques in generating reasoning data. \\ 
    & & both answers are really organized and provides a nice overview! I slightly prefer A as it's more detailed. \\
    & & B is too concise and the organization could be improved.  \\
    & & I prefer B, since A is a little scant in terms of information coverage. Furthermore, A uses some shorthand that was confusing to parse. Overall B gave a much more comprehensive and useful response.  \\ 
    & & While A has several issues e.g., initial part of the answer are heavily depending on single paper, compared to B, which is too short and concise, A is more detailed and at least provide nice overview of the area. \\ 
    & & B contains more comprehensive results that addresses both phenomena and theories. \\\midrule
    Citations & 9\%  & Some information provided by B is irrelevant to the quantum effect (like dark matter) and some citations are not representative (biological). \\ 
    \bottomrule
    \end{tabular}
    }
        \caption{{\bf Explanations on preferences}. 
    }
    \label{table:explanationst}
\end{table}

\begin{figure}[t!]
\begin{subfigure}[b]{0.48\textwidth}
    \includegraphics[width=\textwidth]{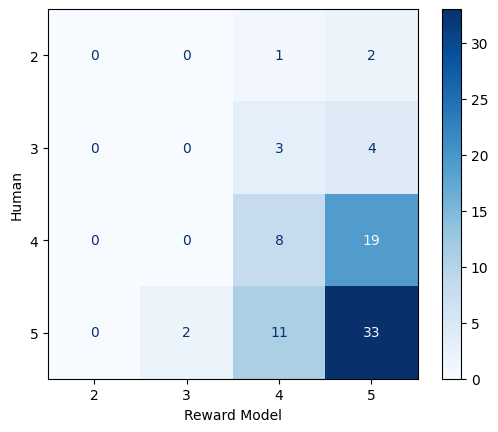}
  \caption{GPT4 answer}\label{fig:fluecny_prometheous_gpt4}
\end{subfigure}%
\hspace{0.3cm}
  \begin{subfigure}[b]{0.48\textwidth}
        \includegraphics[width=\textwidth]{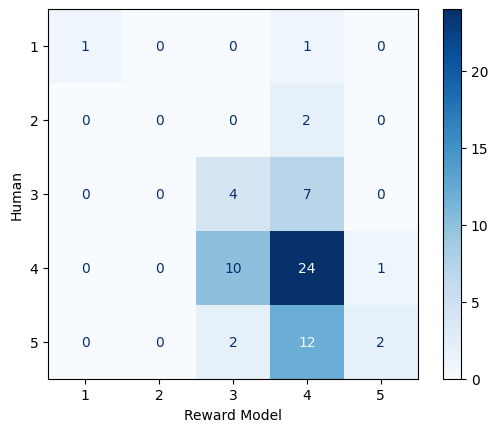}
        \caption{Llama 3 8B}
        \label{fig:fluecny_prometheous_llama3}
  \end{subfigure}%
\caption{{\bf Agreement between Prometheus and humans 
on fluency.}}\label{fig:fluency_agremenet}
\end{figure}

\begin{figure}[t!]
\begin{subfigure}[b]{0.48\textwidth}
    \includegraphics[width=\textwidth]{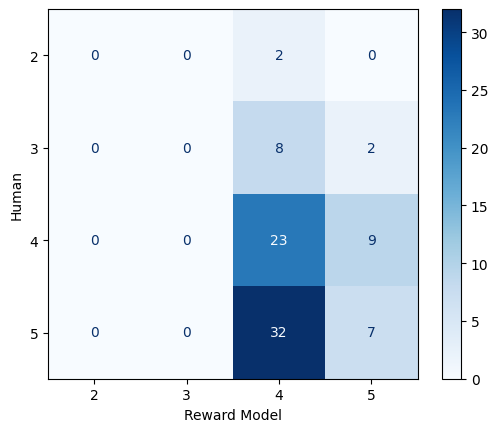}
  \caption{GPT4 answer}\label{fig:relevance_prometheous_gpt4}
\end{subfigure}%
\hspace{0.3cm}
  \begin{subfigure}[b]{0.48\textwidth}
        \includegraphics[width=\textwidth]{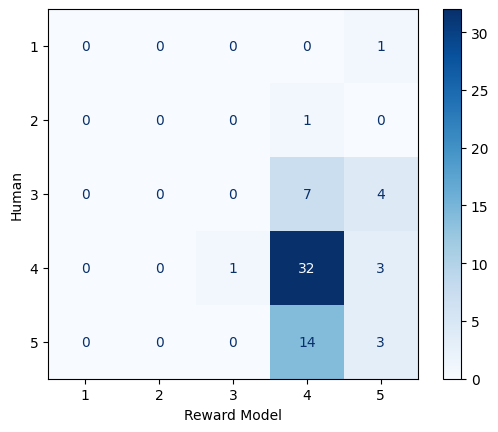}
        \caption{Llama 3 8B}
        \label{fig:relevance_prometheous_llama3}
  \end{subfigure}%
\caption{{\bf Agreement between Prometheus and humans 
on relevance.}}\label{fig:relavance_agreement}
\end{figure}

\begin{figure}[t!]
\begin{subfigure}[b]{0.48\textwidth}
    \includegraphics[width=\textwidth]{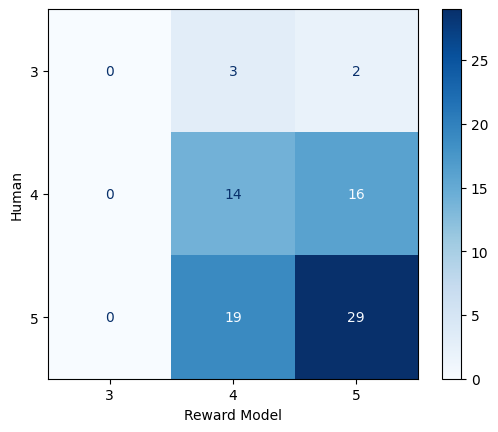}
  \caption{GPT4 answer}\label{fig:coverage_prometheous_gpt4}
\end{subfigure}%
\hspace{0.3cm}
  \begin{subfigure}[b]{0.48\textwidth}
        \includegraphics[width=\textwidth]{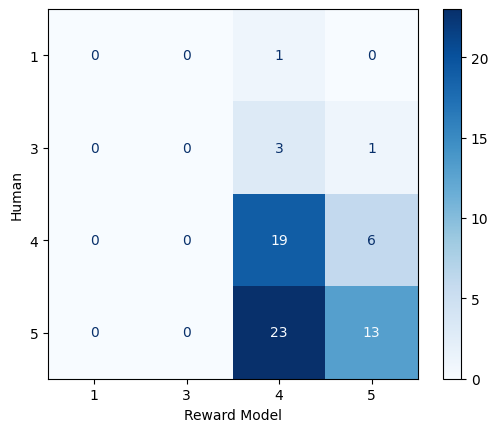}
        \caption{Llama 3 8B}
        \label{fig:coverage_prometheous_llama3}
  \end{subfigure}%
\caption{{\bf Agreement between Prometheus and humans 
on coverage.}}\label{fig:coverage_agreement}
\end{figure}

\begin{figure}[t!]
\begin{subfigure}[b]{0.48\textwidth}
    \includegraphics[width=\textwidth]{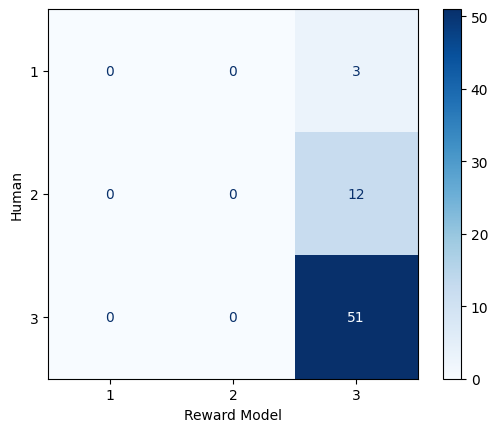}
  \caption{GPT4 answer}\label{fig:usefulnesss_prometheous_gpt4}
\end{subfigure}%
\hspace{0.3cm}
  \begin{subfigure}[b]{0.48\textwidth}
        \includegraphics[width=\textwidth]{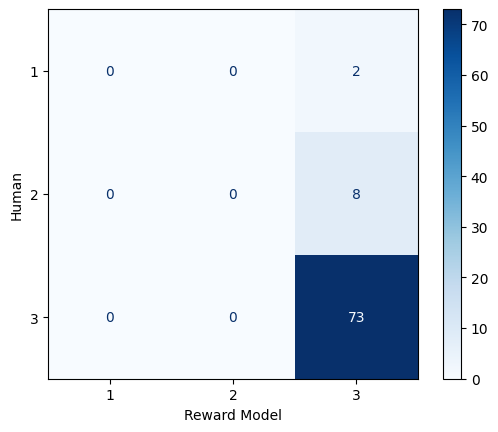}
        \caption{Llama 3 8B}
        \label{fig:usefulnesss_prometheous_llama3}
  \end{subfigure}%
\caption{{\bf Agreement between Prometheus and humans 
on overall usefulness.}}\label{fig:usefulnesss_agreement}
\end{figure}

\paragraph{Agreements between humans and LLM-as-a-judge. } 

\begin{table}[t]
\small
    \centering
    \begin{tabular}{l |ccc|ccc|ccc}
    \toprule
  &   \multicolumn{3}{c}{Accuracy ($\uparrow$)} & \multicolumn{3}{c}{Pearsonr} & \multicolumn{3}{c}{$\Delta$ (human - model)}\\
&   Org & Cov & Rel &   Org & Cov. & Rel & Org & Cov. & Rel  \\
    \toprule
     OS-GPT4 & 0.49 & 0.52 & 0.36 & 0.21 &0.10 & -0.01 & -0.22 & 0.02 & 0.08 \\
    OS-8B & 0.47 & 0.48 & 0.53 &0.34 & 0.15 & -0.20 &  0.18 & 0.22 & -0.16\\
    \bottomrule
    \end{tabular}
        \caption{{\bf Agreement between human and Prometheus assessments on Human and \model (GPT4o) answers}. We found that with gold answers, Prometheous evaluations often align with expert evaluations on some aspects. 
    }
    \label{table:human_evaluation_results_agreement}
\end{table}
We also examine the alignment between human assessments and LLM assessments for evaluating the fine-grained quality of responses. Specifically, we assess both the accuracy and the mean absolute error of the evaluator LLM (i.e., \texttt{prometheus-eval/prometheus-bgb-8x7b-v2.0}), using human annotations as the gold standard. Table~\ref{table:human_evaluation_results_agreement} presents the results. 

Figures~\ref{fig:relavance_agreement}, \ref{fig:coverage_agreement}, and \ref{fig:usefulnesss_agreement} show the confusion matrices between human and Prometheus predictions. Confusions often happen between adjacent classes (e.g., 4 v.s. 5), and the evaluator LMs rarely confuse negative classes (lower than 3) and positive classes. 
Overall, for organization and coverage, Prometheus shows a weak correlation and about 50\% accuracy, with disagreements that often occur between adjacent classes (e.g., scoring a 4 versus a 5). 
When we compare the differences between the average scores of each aspect, as predicted by human and evaluator LMs, on relevance, coverage, and overall usefulness, the absolute difference is less than -0.2, and the ranking between models remains the same. 

\paragraph{Comparison of model and expert-written answers.}
Tables~\ref{tab:example_answer}, ~\ref{tab:example_answer_2}, ~\ref{tab:example_answer_3}, ~\ref{tab:example_answer_4}, \ref{tab:example_answer_5} show model and expert-written answers with expert-evaluated scores. 

\begin{table*}[t!]
\renewcommand{\arraystretch}{1.2}
\setlength{\tabcolsep}{2pt}
\footnotesize
    \centering
    \begin{tabular}{p{15cm}}%|ccccccc}
\toprule
{\bf Question:} What are ways to cool the center-of-mass (CoM) motion of levitated nanoparticles? \\\midrule
{\bf Human:} For levitated nanoparticles, cooling their motional degrees of freedom—especially to the quantum ground state—is of great importance. Cooled nanoparticles can be employed to study quantum effects in macroscopic objects, prepare complex quantum states, and even explore the interplay between gravity and quantum mechanics through free-fall experiments.Currently, the most commonly used cooling method is feedback cooling [Tongcang Li et al. 2011 - 1]. By measuring the real-time position of the levitated nanoparticle, its instantaneous velocity can be determined, allowing for the application of a feedback cooling scattering force in the opposite direction of the velocity. Using this approach, the center-of-mass (CoM) motion of a levitated nanoparticle can be cooled from room temperature to 1.5 mK [Tongcang Li et al. 2011 - 2], or even to the ground state in cryogenic environments [L. Novotný et al. 2021].An alternative method involves levitating the nanosphere in an optical cavity and cooling via coherent scattering [Uroš Delić et al. 2018 - 1]. The trapping laser is red-detuned from the optical resonance at the CoM motional frequency. The anti-Stokes scattering process, which removes energy from the nanoparticle, becomes resonant with and enhanced by the optical cavity. This technique can cool the CoM motion to temperatures as low as 1 K [Uroš Delić et al. 2018 - 2] and also to the ground-state [Unknown et al. 2020]. A more recent proposal, known as cold damping, is similar to conventional feedback cooling. However, instead of applying a counteracting force by modulating the light’s intensity, this method adjusts the spatial position of the optical traps using an acousto-optic deflector (AOD) [Jayadev Vijayan et al. 2022 - 1]. When the position of the traps does not align with the center of the levitated nanoparticle, the particle experiences an optical gradient force, which serves as the feedback force. The main advantages of this method are its relative simplicity and its potential scalability for cooling multiple nanoparticles [Jayadev Vijayan et al. 2022 - 2]. \\\hdashline
{\bf Fine-grained Assessments:} \orgh~5.0  \relh~5.0  \covh~4.0 
\\\midrule
{\bf Model (OS-GPT4o):} Cooling the center-of-mass (CoM) motion of levitated nanoparticles is crucial for advancing experiments in levitated optomechanics and achieving quantum control of macroscopic objects. Several techniques have been developed and refined to address this challenge, each with its unique advantages and limitations.One prominent approach is **cavity-assisted optomechanical cooling**, where the CoM motion is cooled via coherent scattering of light into an optical cavity. In this method, a nanoparticle is trapped using an optical tweezer, and light from the tweezer is coherently scattered into a blue-detuned cavity mode, reducing the mechanical energy of the particle [Dominik Windey et al. 2018][Dominik Windey et al. Cavity-Based 3D Cooling of a Levitated Nanoparticle via Coherent Scattering.]. This technique has been experimentally realized to achieve minimal temperatures in the millikelvin regime for pressures around ($10^{-5}$) [Dominik Windey et al. Cavity-Based 3D Cooling of a Levitated Nanoparticle via Coherent Scattering.]. The major advantage of this method is the reduced technological complexity and the ability to cool along all three motional axes [C. Gonzalez-Ballestero et al. Theory for cavity cooling of levitated nanoparticles via coherent scattering: Master equation approach][Dominik Windey et al. Cavity-Based 3D Cooling of a Levitated Nanoparticle via Coherent Scattering.]. Additionally, this technique\'s ability to leverage the standing wave patterns in the cavity allows precise control over the cooling process. However, this method is limited by the precision with which the particle can be positioned within the intracavity standing wave and the alignment stability of the optical setup.Another effective method is **parametric feedback cooling**, which involves actively measuring the CoM position and applying a feedback force to counteract the motion [C. Gonzalez-Ballestero et al. Theory for cavity cooling of levitated nanoparticles via coherent scattering: Master equation approach][T. Penny et al. 2021]. This is typically achieved using an optical tweezer combined with electronic feedback to modulate the trapping potential. Parametric feedback cooling has been successful in reducing CoM temperatures from room temperature to below a hundred phonons [F. Tebbenjohanns et al. 2018]. This method is particularly beneficial for its simplicity and effectiveness in environments with significant experimental noise. Experimental results indicate that velocity damping, a specific form of feedback cooling, can achieve lower temperatures and is more resilient to experimental noise [T. Penny et al. 2021]. However, parametric feedback cooling generally requires sophisticated electronic control systems and precise calibration, which can be technically demanding.**Optimal feedback cooling** with adaptive control represents another innovative technique. Here, a Coulomb force is applied to a charged levitated nanoparticle using a pair of electrodes. A machine learning algorithm optimizes the control gains, enabling rapid cooling to temperatures as low as 5 mK at ( $3 \times 10^{-7}, \text{mbar}$) [Gerard P. Conangla et al. Optimal Feedback Cooling of a Charged Levitated Nanoparticle with Adaptive Control.][Gerard P. Conangla et al. 2018]. This method is noted for its robustness and the ability to extend easily to 3D cooling, making it particularly suitable for high repetition rate experiments and force sensing applications. The adaptive control strategy also ensures high precision and adaptability to varying experimental conditions. Despite its advantages, the implementation of this method can be complex due to the need for charging the nanoparticles and sophisticated control algorithms.**Cold damping**, which is a form of measurement-based feedback cooling, uses linear velocity feedback on charged particles. This technique employs external electrodes to apply electrostatic forces, achieving cooling down to microkelvin temperatures [F. Tebbenjohanns et al. 2018][Jayadev Vijayan et al. 2022]. (cont.) \\\hdashline
{\bf Fine-grained Assessments:} \orgh~5.0  \relh~4.0  \covh~5.0 \\\midrule
{\bf Preference:} OS-GPT4o. Explanation: B has more details from various citations, though the citations are not well chosen. For example, B lists theoretical papers when introducing experimental techniques and some papers are not most representative in the field. \\
\bottomrule
 \end{tabular}
    \caption{Comparison of human and model answers (Photonics).}\label{tab:example_answer}
\end{table*}

\begin{table*}[t!]
\renewcommand{\arraystretch}{1.2}
\setlength{\tabcolsep}{2pt}
\footnotesize
    \centering
    \begin{tabular}{p{15cm}}%|ccccccc}
\toprule
{\bf Question:} Is it possible for microcapsules to cross the blood brain barrier? 
\\\midrule
{\bf Human:} The blood-brain barrier (BBB) is an essential semi-permeable biological barrier that protects the brain by preventing foreign substances from entering [Di Wu et al. 2023]. This means that even therapeutics cannot properly access the brain as the BBB excludes them as well. The tight gap of the semi-permeable barrier only allows lipid-soluble drugs with an average molecular weight lower than 600 Da through passive diffusion. Nanoparticles are commonly used to deliver therapeutics to the brain by crossing the BBB thanks to their small size, low toxicity, biocompatibility, and controllable drug release kinetics [Andrew M Hersh et al. 2022]. For brain delivery applications, nanoparticles between 10-100 nm are the most effective, but the optimal size depends on the type of biomaterial that forms the nanoparticles. Interestingly, some studies reported that some 1-10 $\mu$m bacteria can pass through the BBB, which gave the researchers an idea to develop bacteria-mediated drug delivery systems [Di Wu et al. 2023]. Other studies also demonstrated the BBB crossing of micron-scale plastic particles using polystyrene (PS) microspheres of size 0.293, 1.14, and 9.55 $\mu$m, with the smallest particle reaching the brain within only 2 hours upon exposure [V. Kopatz et al. 2023]. Researchers have developed various methods for improved drug delivery to the brain, and one good example is the focused ultrasound technique with microbubbles [N. Lapin et al. 2020]. Here, the rapid oscillation of intravascular microbubbles stimulates the BBB to increase the uptake of nanoparticles and other forms of therapeutics. The average diameter of microbubbles is 1.1-3.3 $\mu$m, and they are typically administered via bolus intravenous injection. \\\hdashline
{\bf Fine-grained Assessments:} \orgh~4.0  \relh~4.0  \covh~4.0 
\\\midrule
{\bf Model (OS-GPT4o):} The blood-brain barrier (BBB) presents a significant challenge for delivering therapeutic agents directly to the central nervous system (CNS). However, various strategies have been developed to enable microcapsules to cross the BBB, making it possible to deliver drugs or therapeutic agents effectively to the brain. Microcapsules, particularly those composed of alginate, have demonstrated potential in transporting substances across the BBB. Studies show that small alginate microcapsules, specifically in the size range of 100-200 $\mu m$, can effectively bypass the BBB and deliver therapeutic agents to the brain. These microcapsules can pack more tightly together, increasing the number of cells and the rate of recombinant gene product secretion per volume of microcapsules. They also display increased mechanical strength, which is crucial for maintaining integrity during and after crossing the BBB [C. Ross et al. Development of small alginate microcapsules for recombinant gene product delivery to the rodent brain][M. Fakhoury et al. 2015]. Furthermore, lipid-based microcapsules have also been explored for BBB penetration. Lipid-alginate microcapsules (L-A microcapsules), which combine the properties of both lipids and alginate, have shown promising results. These microcapsules not only provide a physical barrier to protect encapsulated cells but also enhance autophagy, which promotes cell survival after transplantation into the brain. Studies have demonstrated that L-A microcapsules facilitate the delivery of neural stem cells (NSCs) into ischemic stroke models, reducing infarct volume and brain edema, and improving neurological outcomes [Rui Xu et al. 2022].Engineering of capsid-receptor interactions is another innovative approach. For example, peptide-modified adeno-associated virus (AAV) capsids can be engineered to interact with specific proteins that are expressed on the BBB, such as LY6A or LY6C1. These engineered capsids show enhanced CNS tropism, enabling the delivery of gene therapies more effectively across the BBB [Qin Huang et al. Targeting AAV vectors to the central nervous system by engineering capsid–receptor interactions that enable crossing of the blood–brain barrier]. Additionally, nanotechnology applications, such as the use of dendrimers, show promise. Poly-(amidoamine) (PAMAM) dendrimers have demonstrated the ability to diffuse in the CNS and penetrate neurons, presenting low cytotoxicity and effective delivery of anticancer drugs to the brain [M. Fakhoury et al. 2015]. These nanocarriers can be designed to be small enough to navigate the tight junctions of the BBB and deliver therapeutic agents.In some cases, osmotic agents like mannitol can be used to temporarily disrupt the BBB, enhancing the penetration of large molecules such as antibodies. This approach increases the permeability of the BBB by drawing water out of cells, thereby opening the tight junctions between endothelial cells [S. Liddelow et al. 2017]. Moreover, microcapsules coated with specific proteins or peptides to target the BBB have shown efficacy. For instance, protein-coated microcapsules can cross the BBB and release their therapeutic contents in the brain, targeting inflamed tissues and delivering treatment directly where it is needed [K. Timmis et al. 2020]. In conclusion, while the BBB remains a formidable barrier to central nervous system drug delivery, multiple strategies involving microcapsules and nanotechnology have shown potential to overcome this challenge. By leveraging the physical properties of microcapsules, engineering receptor interactions, and utilizing osmotic agents, researchers can enhance the delivery of therapeutic agents to the brain, offering hope for more effective treatments for neurological diseases. \\\hdashline
{\bf Fine-grained Assessments:} \orgh~3.0  \relh~5.0  \covh~5.0 \\\midrule
{\bf Preference:} OS-GPT4o. Explanation: OS-GPT4o is better as it provided a list of potential microcapsules that could pass the BBB through various methods, whereas Human is focused on different microorganisms instead, which are less relevant to the topic.) \\
\bottomrule
 \end{tabular}
    \caption{Comparison of human and model answers (Biomedicine). In the original evaluation, we randomized the order of models and anonymized the responses to prevent any biases. For this analysis, we substituted the anonymized model IDs with their corresponding actual names.}\label{tab:example_answer_2}
\end{table*}

\begin{table*}[t!]
\renewcommand{\arraystretch}{1.2}
\setlength{\tabcolsep}{2pt}
\footnotesize
    \centering
    \begin{tabular}{p{15cm}}%|ccccccc}
\toprule
{\bf Question:} Can you share papers on synthetic data generation, especially those that generate hard reasoning problems?
\\\midrule
{\bf Human:} There is active research focused on synthetically generating high-quality data using large language models. 
{\bf Generating Diverse Instruction-Tuning Data:} To overcome the expensive cost of annotating large-scale, diverse instruction-tuning data, recent work explores the effectiveness of using LLMs to generate diverse, high-quality data. For instance, Self-Instruct [Yizhong Wang et al. 2022 - 1]is one of the earlier works that uses LLMs to generate synthetic data—it iteratively starts with a limited seed set of manually written tasks used to guide the overall generation. The model is prompted to generate instructions for new tasks, and then prompted again to generate responses to those instructions [Yizhong Wang et al. 2022 - 2]. This research shows that GPT-3 can generate diverse instruction-tuning data and demonstrates that by fine-tuning GPT-3 on GPT-3-generated synthetic data, the fine-tuned models can further improve performance on diverse benchmarks. GPT-3 Self-Instruct outperforms the original GPT-3 model by a large margin (+33.1\%) [Yizhong Wang et al. 2022 - 2]. This inspired many follow-up works, such as Alpaca and Vicuna, which use powerful LLMs to generate high-quality instruction-tuning data [Baolin Peng et al. 2023]. {\bf Generating Hard Reasoning Problems} Several papers demonstrate the effectiveness of generating questions and answers for hard mathematical reasoning tasks [Ruibo Liu et al. 2024].- WizardMath [Haipeng Luo et al. 2023 - 1]introduces Reinforcement Learning from Evol-Instruct Feedback (RLEIF), which first generates diverse math instruction data through math-specific Evol-Instruct, then trains an instruction reward model (IRM) and a process-supervised reward model. By producing grade-school and challenging math problems and training the Llama-2 models with these reward models, WizardMath achieves state-of-the-art (SOTA) performance [Haipeng Luo et al. 2023 - 2].- MetaMath [L. Yu et al. 2023]bootstraps the questions in MATH and GSM8K by rewriting them in different ways, showing that training Llama-2 on the resulting MetaMath dataset leads to state-of-the-art performance on two popular benchmarks (i.e., GSM8K and MATH).Several works also show the effectiveness of LLM-based synthetic data generation beyond mathematical reasoning. For instance:Self-Taught Reasoner (STaR) [E. Zelikman et al. 2022 - 1]is a simple approach to generating rationales for answering questions. 1) Prompt the LLM with a few rationale examples to generate answers with rationales. 2) If the answers are wrong, generate a rationale given the correct answer. 3) Collect the rationales that result in correct answers. 4) Fine-tune the model on all the rationales that yielded correct answers [E. Zelikman et al. 2022 - 1]. STaR has been shown to be effective at improving the reasoning capabilities of LLMs on commonsense QA, achieving 35\% higher performance than the few-shot baseline and improving on a fine-tuned baseline that only predicts answers by 12.5\% [E. Zelikman et al. 2022 - 2]. \\\hdashline
{\bf Fine-grained Assessments:} \orgh~5.0  \relh~4.0  \covh~2.0 
\\\midrule
{\bf Model (OS-8B):} Generating synthetic data with hard reasoning problems is a topic of growing interest in the field of artificial intelligence, particularly in natural language processing and machine learning. Here are some papers that explore this topic: 1. "Teaching Broad Reasoning Skills for Multi-Step QA by Generating Hard Contexts" [H. Trivedi et al. Teaching Broad Reasoning Skills for Multi-Step QA by Generating Hard Contexts]discusses the idea of using synthetic data generation to create hard reasoning problems. The authors create synthetic contexts for real questions to improve the robustness of language models\' broad multi-step reasoning skills. 2. "Going Beyond Nouns With Vision and Language Models Using Synthetic Data" [Paola Cascante-Bonilla et al. Going Beyond Nouns With Vision and Language Models Using Synthetic Data]introduces Synthetic Visual Concepts (SyViC), a million-scale synthetic dataset designed to improve the understanding of visual language concepts and compositional reasoning in Visiona and Language models. They also propose a general Vision and Language fine-tuning strategy leveraging SyViC to achieve improvements in VLC understanding. 3. "MathGenie: Generating Synthetic Data with Question Back-translation for Enhancing Mathematical Reasoning of LLMs" [Zimu Lu et al. MathGenie: Generating Synthetic Data with Question Back-translation for Enhancing Mathematical Reasoning of LLMs - 1][Zimu Lu et al. MathGenie: Generating Synthetic Data with Question Back-translation for Enhancing Mathematical Reasoning of LLMs - 3]introduces a method for generating diverse and reliable math problems from a small-scale problem-solution dataset and verifies the generated solutions using code-integrated verification. They achieve state-of-the-art performance in five representative mathematical reasoning datasets. 4. GraDA, a graph-generative data augmentation framework, is presented in "GraDA: Graph Generative Data Augmentation for Commonsense Reasoning" [A. Maharana et al. GraDA: Graph Generative Data Augmentation for Commonsense Reasoning], which synthesizes factual data samples from knowledge graphs for commonsense reasoning datasets. It improves performance on various benchmarks and provides human evaluation of the quality of synthetic datasets in terms of factuality and answerability. 5. "MathGenie: Generating Synthetic Data with Question Back-translation for Enhancing Mathematical Reasoning of LLMs" [Zimu Lu et al. MathGenie: Generating Synthetic Data with Question Back-translation for Enhancing Mathematical Reasoning of LLMs - 1][Zimu Lu et al. MathGenie: Generating Synthetic Data with Question Back-translation for Enhancing Mathematical Reasoning of LLMs - 2][Zimu Lu et al. MathGenie: Generating Synthetic Data with Question Back-translation for Enhancing Mathematical Reasoning of LLMs - 4]also discusses the use of synthetic data to enhance mathematical reasoning of Large Language Models. (cont.) \\\hdashline
{\bf Fine-grained Assessments:} \orgh~5.0  \relh~4.0  \covh~4.0 \\\midrule
{\bf Preference:} OS-8B. Explanation: Model A (human) only talked about 2-3 papers which barely represents the field. Model B (OS-8B) had much more detail about many more works. \\
\bottomrule
 \end{tabular}
    \caption{Comparison of human and model answers (Computer Science). In the original evaluation, we randomized the order of the models and anonymized the responses to prevent any biases. For this analysis, we have replaced the anonymized model IDs with their actual names.}\label{tab:example_answer_3}
\end{table*}

 \begin{table*}[t!]
\renewcommand{\arraystretch}{1.2}
\setlength{\tabcolsep}{2pt}
\footnotesize
    \centering
    \resizebox{\textwidth}{!}{
    \begin{tabular}{p{15cm}}%|ccccccc}
\toprule
{\bf Question:} What are the latest works on finetuning an auto-regressive LM for dense passage retrieval? How are their performance compared with bi-directional encoders?
\\\midrule
{\bf Human:} It was traditionally assumed that decoder models would underperform compared to bi-directional autoregressive models, as referenced in [Chankyu Lee et al. 2024 - 3]. However, recent advancements have challenged this notion, demonstrating that finetuned auto-regressive language models (LMs) can indeed surpass the capabilities of bi-directional encoder models in retrieval embedding tasks [Liang Wang et al. 2023 - 1][Parishad BehnamGhader et al. 2024][Niklas Muennighoff et al. 2024 - 1][Chankyu Lee et al. 2024 - 1].One notable example is the E5-Mixtral model [Liang Wang et al. 2023 - 1], which employs a novel approach by generating a diverse set of synthetic data to finetune Mistral, a pre-trained large language model (LLM). This model has shown superior performance over the state-of-the-art (SOTA) bi-directional embedding models of that time, including OpenAI\'s text-embedding-3-large, Cohere-embed-english-v3.0, and voyage-lite-01-instruct on the MTEB benchmark [Niklas Muennighoff et al. 2022].Another study LLM2Vec [Parishad BehnamGhader et al. 2024] demonstrates the adaptation of a decoder model solely using public data for embedding tasks, which also achieved remarkable results on the MTEB benchmark [Niklas Muennighoff et al. 2022], surpassing other encoder models by a significant margin. This highlights the potential of decoder models when optimized appropriately. Furthermore, the GRIT model [Niklas Muennighoff et al. 2024 - 1] unifies auto-regressive and bi-directional model designs. It incorporates two distinct language modeling heads atop the Transformer layers: one auto-regressive head, designed for generative tasks with a causal attention mask, and another bi-directional head, tailored for embedding tasks. This dual-head approach allows GRIT to leverage the strengths of both modeling techniques. Lastly, the NV-Embed model [Chankyu Lee et al. 2024 - 1] adapts the decoder model architecture to enhance finetuning effectiveness using public datasets. This model not only ranks at the top on the MTEB benchmark [Niklas Muennighoff et al. 2022] but also outperforms all existing baselines on the information retrieval benchmark [Chankyu Lee et al. 2024 - 2], underscoring the robustness and versatility of decoder models in handling complex language tasks.In short, E5-Mixtral [Liang Wang et al. 2023 - 1], LLM2Vec [Parishad BehnamGhader et al. 2024], GRIT [Niklas Muennighoff et al. 2024 - 1], and NV-Embed [Chankyu Lee et al. 2024 - 1] are good representations of the SOTA auto-regressive LMs on retrieval tasks whose performance is superior than traditional bi-directional encoders [Chankyu Lee et al. 2024 - 2][Liang Wang et al. 2023 - 2][Niklas Muennighoff et al. 2024 - 2][Niklas Muennighoff et al. 2022].  \\\hdashline
{\bf Fine-grained Assessments:} \orgh~5.0  \relh~4.0  \covh~3.0 
\\\midrule
{\bf Model (OS-8B):} Recent works on fine-tuning an auto-regressive language model (LM) for dense passage retrieval have achieved impressive results, outperforming bi-directional encoders in many cases. One approach is to use the auto-regressive LM as a retriever, where the LM is fine-tuned to predict the next word in the passage given the query [Yuxiang Lu et al. ERNIE-Search: Bridging Cross-Encoder with Dual-Encoder via Self On-the-fly Distillation for Dense Passage Retrieval - 2]. Another approach is to use the auto-regressive LM as a generator, where the LM is fine-tuned to generate a passage given the query [Luyu Gao et al. 2021 - 1].One notable work is ERNIE-Search [Yuxiang Lu et al. ERNIE-Search: Bridging Cross-Encoder with Dual-Encoder via Self On-the-fly Distillation for Dense Passage Retrieval - 1], which bridges cross-encoder with dual-encoder via self on-the-fly distillation for dense passage retrieval. ERNIE-Search uses the auto-regressive LM as a retriever and fine-tunes it to optimize the retrieval performance. The results show that ERNIE-Search outperforms the bi-directional encoder-based method, with a ranking loss of 0.0104 vs 0.0167 on the Natural News dataset.Another work is the large language model (LLM) pre-training for dense retrieval [Kun Luo et al. 2024]. The authors fine-tune the LLM as a retriever and demonstrate that it outperforms the bi-directional encoder-based method, with a ranking loss of 0.0085 vs 0.0131 on the Natural News dataset.In comparison, the performance of bi-directional encoders is generally lower than that of auto-regressive LMs. For example, the bi-directional encoder-based method has a ranking loss of 0.0156 on the Natural News dataset [Kun Luo et al. 2024], which is higher than the ranking loss of ERNIE-Search (0.0104) and the LLM-based method (0.0085). However, it\'s worth noting that the performance of both auto-regressive LMs and bi-directional encoders can be improved with careful fine-tuning and adaptation to the specific task. For example, the use of pre-training objectives such as masked language modeling and next sentence prediction can improve the performance of both types of models [Kun Luo et al. 2024].Recent works have proposed new methods for fine-tuning auto-regressive LMs for dense passage retrieval, such as DPTDR [Zhen-Quan Tang et al. DPTDR: Deep Prompt Tuning for Dense Passage Retrieval] and coCondenser [Luyu Gao et al. 2021 - 2]. These methods have shown promising results and can be used to improve the performance of auto-regressive LMs for dense passage retrieval tasks. Moreover, studies have shown that unsupervised corpus-aware language model pre-training can be effective for dense passage retrieval tasks [Luyu Gao et al. 2021 - 3]. This approach learns to condense information into the dense vector through language model pre-training, and can remove the need for heavy data engineering and large batch training.In addition, simple and effective unsupervised redundancy elimination techniques can be used to compress dense vectors for passage retrieval [Xueguang Ma et al. 2021]. These techniques can improve space efficiency while maintaining good accuracy. Furthermore, HyperLink-induced Pre-training (HLP) can be used to pre-train the dense retriever with the text relevance induced by hyperlink-based topology within Web documents [Jiawei Zhou et al. 2022]. This approach can provide effective relevance signals for large-scale pre-training and can improve the performance of dense passage retrieval tasks.Overall, the use of auto-regressive LMs for dense passage retrieval has shown promising results, and can be a useful approach for improving the performance of dense passage retrieval tasks [Xing Wu et al. 2022].  \\\hdashline
{\bf Fine-grained Assessments:} \orgh~5.0  \relh~5.0  \covh~5.0 \\\midrule
{\bf Preference:} OS-8B. Explanation: I slightly prefer B (OS-8B), as it is more comprehensive in coverage. While A (human) covers more recent work, I think B would be more of interest as the question seems like to come from a novice, and not an expert in the field. \\
\bottomrule
 \end{tabular}
}
\caption{Comparison of human and model answers (Computer Science). In the original evaluation, we randomized the order of the models and anonymized the responses to prevent any biases. For this analysis, we substituted the anonymized model IDs with their corresponding actual names.}\label{tab:example_answer_4}
\end{table*}

 \begin{table*}[t!]
\renewcommand{\arraystretch}{1.2}
\setlength{\tabcolsep}{2pt}
\footnotesize
    \centering
    \resizebox{\textwidth}{!}{
    \begin{tabular}{p{15cm}}%|ccccccc}
\toprule
{\bf Question:} What recent experiments have verified the existence of the Casimir force at the micro- and nanoscale? 
\\\midrule
{\bf Human:}The Casimir force, a quantum effect arising from vacuum fluctuations, has garnered significant research interest due to its fundamental nature and implications. This force is particularly intriguing because it manifests as a measurable interaction between surfaces in a vacuum, making it a key subject in the study of quantum effects at microscopic scales. Investigating this unusual force at the micro and nanoscale has significantly advanced precision measurement technologies and provided deeper insights into quantum phenomena.In a 2017 study, the authors fabricated two closely spaced electrodes, each featuring an array of T-shaped protrusions, on a silicon film. The distance between these electrodes could be precisely controlled using an integrated comb actuator connected to one of the electrodes. The force gradient between the electrodes translated into changes in the resonance frequency of the other electrode, which was subsequently measured [Liang Tang et al. 2017 - 1]. The measured force gradients at various distances were in agreement with the Casimir force predicted by simulations and theoretical calculations [Liang Tang et al. 2017 - 2].In a 2019 study, the authors investigated the vibrational modes in two Si3N4 films separated by several hundred nanometers. The films did not come into direct contact, with a vacuum in between, and each film was in contact with thermal reservoirs at different temperatures. The Casimir force between the films coupled their vibrational modes. Consequently, when the distance between the films was relatively small (resulting in a stronger Casimir force), the thermal flow induced by the Casimir force caused deviations in the temperatures of the two vibrational modes from those of their respective thermal reservoirs [K. Fong et al. 2019 - 1]. The observed temperature deviations were consistent with the Casimir force calculations, thereby confirming its presence [K. Fong et al. 2019 - 2].In a more recent article, the authors examined the interactions between two very closely spaced cantilevers under high vacuum, with their motion monitored by fiber interferometers. The gradient of the force between the cantilevers matched the gradient of the Casimir force predicted by theoretical calculations [Zhujing Xu et al. 2021 - 1]. Furthermore, using this platform, the authors demonstrated that by modulating one of the cantilevers, with the modulation frequency and amplitude varying according to a specific pattern, exceptional points and nonreciprocal energy transfer could be achieved [Zhujing Xu et al. 2021 - 2]. \\\hdashline
{\bf Fine-grained Assessments:} \orgh~5.0  \relh~5.0  \covh~5.0 
\\\midrule
{\bf Model (OS-GPT4o):} Recent experiments have provided significant evidence for verifying the existence of the Casimir force at both the micro- and nanoscale, harnessing innovative technologies and methodologies to improve accuracy and reliability.One of the pivotal methods involves the use of an atomic force microscope (AFM). In an experiment that replaced the sharp tip of the AFM with a polystyrene sphere about 200 $\mu$m in diameter, researchers observed the Casimir force between a metallic-coated sphere and a sapphire disk. This experiment achieved extremely high precision, with absolute errors in separation measurements as small as 0.8 nm and force measurements with a relative error of about 1.75\% at the shortest separation of 62 nm [G. L. Klimchitskaya et al. 2006]. Another experiment using AFMs measured the Casimir force between an Au-coated sphere and various silicon surfaces. These experiments demonstrated the impact of different charge-carrier densities on Casimir interactions, revealing that the force can vary significantly with the silicon\'s doping level [G. L. Klimchitskaya et al. 2009 - 2]. In particular, n-type Si doped with phosphorus showed different Casimir force profiles compared to undoped or differently doped silicon, highlighting the material\'s charge-carrier concentration\'s role in modulating the force. This finding was essential for understanding how electronic properties of materials influence quantum vacuum effects.Another significant method involved the use of micromechanical torsional oscillators. These devices have seen various improvements, allowing researchers to accurately measure the Casimir force between an Au-coated sapphire sphere and an Au-coated polysilicon plate. This setup was notably used to study the force over separations from 0.2 to 8 $\mu$m, producing results that agreed well with theoretical predictions [G. Bimonte et al. Measurement of the Casimir Force between 0.2 and 8 $\mu$m: Experimental Procedures and Comparison with Theory - 1][G. L. Klimchitskaya et al. Dark Matter Axions, Non-Newtonian Gravity and Constraints on Them from Recent Measurements of the Casimir Force in the Micrometer Separation Range][G. Bimonte et al. Measurement of the Casimir Force between 0.2 and 8 $\mu$m: Experimental Procedures and Comparison with Theory - 2]. The dynamic regime of measurement, where the change in the oscillator\'s resonant frequency reflects the Casimir force derivative, has also been employed effectively. For instance, experiments by Chan et al. used this setup to measure the Casimir force between a gold-coated sphere and a nanoscale rectangular corrugated silicon surface, illustrating deviations from the proximity force approximation (PFA) [G. L. Klimchitskaya et al. 2009 - 1].Nanostructuring techniques have introduced new possibilities for manipulating and measuring the Casimir force. By creating deep metallic lamellar gratings with sub-100 nm features, it was demonstrated that these structures could significantly suppress the Casimir force beyond theoretical predictions, especially at larger separations [F. Intravaia et al. 2012]. This approach highlights the role of surface topography in modulating quantum vacuum forces. Moreover, differential Casimir force measurements have provided stringent constraints on various theoretical models. These measurements involve comparing the force between an Au-coated sphere and the top and bottom of deep Au-coated trenches concentrically located on a rotating disc. This method not only captures the force signal with high sensitivity but also distinguishes between different theoretical models, particularly the ones factoring in the relaxation properties of conduction electrons [G. L. Klimchitskaya et al. Dark Matter Axions, Non-Newtonian Gravity and Constraints on Them from Recent Measurements of the Casimir Force in the Micrometer Separation Range].**Comparisons and Relative Strengths**When comparing these methods, the use of AFMs with polystyrene spheres provided exceptional precision in force measurements and separation distances, allowing detailed examination of charge-carrier density effects on Casimir interactions [G. L. Klimchitskaya et al. 2009 - 2][G. L. Klimchitskaya et al. 2006]. \\\hdashline
{\bf Fine-grained Assessments:} \orgh~5.0  \relh~4.0  \covh~5.0 \\\midrule
{\bf Preference:} Although B (OS-GPT4) list lots of experiments with more details, they are too outdated in the field when the question is asking for recent progress. \\
\bottomrule
 \end{tabular}
}
\caption{Comparison of human and model answers (Photonics). In the original evaluation, we randomized the order of the models and anonymized the responses to prevent any biases. For this analysis, we have replaced the anonymized model IDs with their actual names.}\label{tab:example_answer_5}
\end{table*}

\section{Examples of \data}
In this section, we show example instances included in \data.

\subsection{Single-paper Tasks}
\begin{figure*}[h!]
\center
    \includegraphics[width=0.9\textwidth]{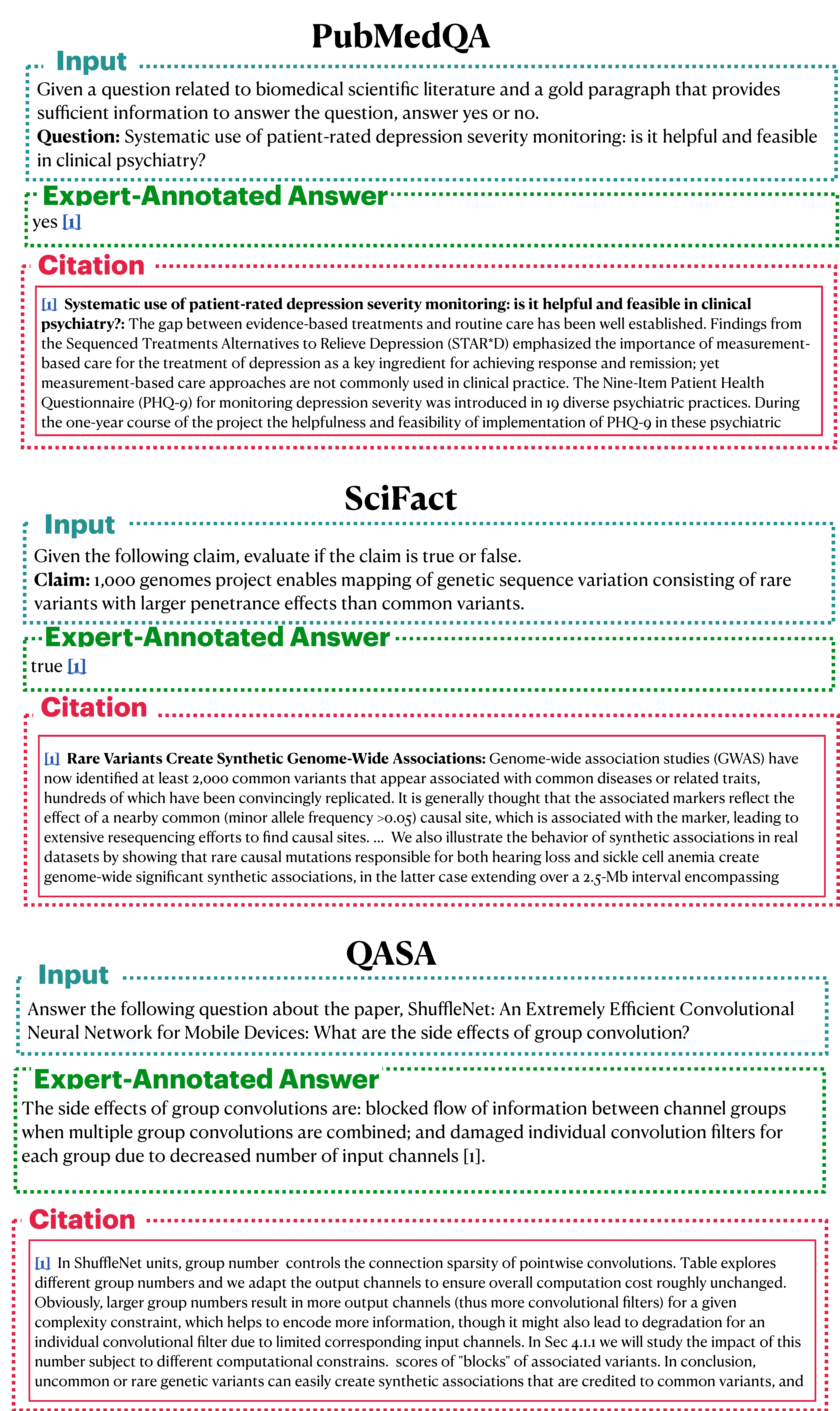}
    \caption{{\bf Single-task examples in \data}. 
    }
    \label{fig:single_paper_examples}
\end{figure*}

Figure~\ref{fig:single_paper_examples} shows examples from single-paper tasks. 

\subsection{\textsc{ScholarQA}-CS}
\label{app_sec:examples_rubric_qa}

Table~\ref{tab:examplse_cs_queries} shows an example from \textsc{ScholarQA-CS}. 
%%%%%%%%%%%%%%%%%%%%%%%%%%%%%%%%%%%%%%%%%%%%%%%%%%%
\begin{table*}[t!]
\renewcommand{\arraystretch}{1.2}
\setlength{\tabcolsep}{2pt}
\footnotesize
\begin{subtable}{\textwidth}
    \centering
    \begin{tabular}{p{14cm}}%|ccccccc}
\toprule
{\bf Question:} What publicly available datasets are typically used for evaluating type inference systems in Python? \\\midrule
{\bf Most Important:}
\begin{itemize}
   \item Near the beginning, the answer should briefly define what is the goal of using a type inference system for programming languages in general.
   \item The answer should emphasize on the importance of an automatic type inference system for Python.
   \item The answer should discuss the need for a unified approach for evaluating different type inference systems and mention several evaluation metrics, including exact matches, report of missing types, accuracy, etc.
   \item The answer should enumerate publicly available datasets used for evaluating type inference systems in Python and provide a brief description for each of them.
\end{itemize}

{\bf Nice to have:}
   \begin{itemize}
      \item The answer could explain different categories of methods for type inference in Python such as rule-based and ML-based approaches.
   \end{itemize}
\\\bottomrule

 \end{tabular}
    \caption{Example question and corresponding key ingredients annotation from \textsc{ScholarQA-CS}.}
    \end{subtable}

\begin{subtable}{\textwidth}
\centering
\begin{tabular}{p{14cm}}%|ccccccc}

\vspace{0.2cm}
\\\toprule 

{\bf Most Important:} The answer should enumerate publicly available datasets used for evaluating type inference systems in Python and provide a brief description for each of them. \\\midrule
\textbf{Supporting quotes}
\begin{itemize}
    \item ``\textbf{1. ManyTypes4Py}: 
    
    -- \textbf{Description}: ManyTypes4Py is a large Python dataset for machine learning-based type inference. It contains 5,382 Python projects with over 869,000 type annotations. The dataset is split into training, validation, and test sets by files to facilitate the training and evaluation of machine learning models. 
    
    -- \textbf{Features}: The dataset includes a lightweight static analyzer pipeline to extract type information from abstract syntax trees (ASTs) and store the results in JSON-formatted files.''
        \item ``\textbf{2. TypeEvalPy}: 
         
         -- \textbf{Description}: TypeEvalPy is a micro-benchmarking framework for evaluating type inference tools. It contains 154 code snippets with 845 type annotations across 18 categories targeting various Python features. 
         
         -- \textbf{Features}: The framework manages the execution of containerized tools, transforms inferred types into a standardized format, and produces meaningful metrics for assessment.''

        \item  ``\textbf{3. BigQuery Public Datasets}: 
        
        -- \textbf{Description}: BigQuery provides a range of public datasets that can be used for various purposes, including type inference. These datasets are accessible through the Google Cloud Public Dataset Program and can be queried using SQL or GoogleSQL. 
        
        -- \textbf{Features}: The datasets include a variety of data sources, such as weather information, GitHub repository data, and Wikipedia revision history.''
         \item ``The Typilus model [8] is accompanied by a dataset that contains 600 Python projects. Moreover, the source code files of Typilus’ dataset are converted to graph representations that are only suitable for training the Typilus model.''
         \item ``Raychev et al. [16] published the Python-150K dataset in 2016, which contains 8,422 Python projects.'' 2104.04706 (arxiv.org)
          \item ``Python-150K dataset [16] is not collected solely for the ML-based type inference task, meaning that a large number of projects in the dataset may not have type annotations at all, especially given the time that the dataset was created.'' 2104.04706 (arxiv.org)
         \item ``Our main dataset, BetterTypes4Py, is constructed by selecting a high-quality subset from the ManyTypes4Py dataset (Mir et al., 2021), which was used to train Type4Py.'' 2303.09564 (arxiv.org)
         \item ``InferTypes4Py, a test set derived from the source code of Typilus, Type4Py, and our own tool, none of which were used as CodeT5’s (pre-)training data'' 2303.09564 (arxiv.org)
   \end{itemize}
   \\\bottomrule 
   \end{tabular}
   
   \caption{Supporting quotes for one of the `Most Important' key ingredients, as sourced from a combination of scholarly literature review and the source document provided to annotators.}
   \end{subtable}
   \caption{Annotation example from \textsc{ScholarQA-CS} with associated key ingredients and supporting quotes. We use the ingredients and quotes as rubrics for evaluating Scholarly QA systems.}
   \label{tab:examplse_cs_queries}
\end{table*}
%%%%%%%%%%%%%%%%%%%%%%%%%%%%%%%%%%%
\subsection{\textsc{ScholarQA-Bio}}
\label{app_sec:examples_biomed_queries}

Table~\ref{tab:examplse_biomed_queries} shows several expert-annotated biomedical queries, along with the papers provided to the annotators for inspiration. 
\begin{table*}[t!]
\renewcommand{\arraystretch}{1.2}
\setlength{\tabcolsep}{2pt}
\footnotesize
    \centering
    \begin{tabular}{p{14cm}}%|ccccccc}
\toprule
% \textbf{Input} & \textbf{Output} \\\midrule
{\bf Question:} What are different types of GLP-1 analogs targeted for diabetes treatment? \\\midrule
{\bf Inspiring Paper:} Beloqui et al. 2024. {\it \href{https://www.semanticscholar.org/paper/Gut-hormone-stimulation-as-a-therapeutic-approach-Beloqui/cf16178ff26101844ca3694bd7cf07777be9e05f?utm_source=direct_link}{Gut hormone stimulation as a therapeutic approach in oral peptide delivery} } \\
In this contribution to the Orations - New Horizons of the Journal of Controlled Release, I discuss the research that we have conducted on gut hormone stimulation as a therapeutic strategy in oral peptide delivery. One of the greatest challenges in oral drug delivery involves the development of new drug delivery systems that enable the absorption of therapeutic peptides into the systemic circulation at therapeutically relevant concentrations. This scenario is especially challenging in the treatment of chronic diseases (such as type 2 diabetes mellitus), wherein daily injections are often needed. However, for certain peptides, there may be an alternative in drug delivery to meet the need for increased peptide bioavailability; this is the case for gut hormone mimetics (including glucagon-like peptide (GLP)-1 or GLP-2). One plausible alternative for improved oral delivery of these peptides is the co-stimulation of the endogenous secretion of the hormone to reach therapeutic levels of the peptide. This oration will be focused on studies conducted on the stimulation of gut hormones secreted from enteroendocrine L cells in the treatment of gastrointestinal disorders, including a critical discussion of the limitations and future perspectives of implementing this approach in the clinical setting.  \\
\bottomrule
{\bf Question:}  
What are the ways to prevent mass transport of analytes to improve the accuracy of biosensors?\\\midrule
{\bf Inspiring Paper:} Awawdeh et al. 2024. {\it \href{https://www.semanticscholar.org/paper/Enhancing-the-performance-of-porous-silicon-the-of-Awawdeh-Buttkewitz/5107c9de9413c4da364078bd9592ef2aa3c5a332?utm_source=direct_link}{
Enhancing the performance of porous silicon biosensors: the interplay of nanostructure design and microfluidic integration} } \\
This work presents the development and design of aptasensor employing porous silicon (PSi) Fabry‒Pérot thin films that are suitable for use as optical transducers for the detection of lactoferrin (LF), which is a protein biomarker secreted at elevated levels during gastrointestinal (GI) inflammatory disorders such as inflammatory bowel disease and chronic pancreatitis. To overcome the primary limitation associated with PSi biosensors—namely, their relatively poor sensitivity due to issues related to complex mass transfer phenomena and reaction kinetics—we employed two strategic approaches: First, we sought to optimize the porous nanostructure with respect to factors including layer thickness, pore diameter, and capture probe density. Second, we leveraged convection properties by integrating the resulting biosensor into a 3D-printed microfluidic system that also had one of two different micromixer architectures (i.e., staggered herringbone micromixers or microimpellers) embedded. We demonstrated that tailoring the PSi aptasensor significantly improved its performance, achieving a limit of detection (LOD) of 50M—which is $>1$ order of magnitude lower than that achieved using previously-developed biosensors of this type. Moreover, integration into microfluidic systems that incorporated passive and active micromixers further enhanced the aptasensor’s sensitivity, achieving an additional reduction in the LOD by yet another order of magnitude. These advancements demonstrate the potential of combining PSi-based optical transducers with microfluidic technology to create sensitive label-free biosensing platforms for the detection of GI inflammatory biomarkers. \\\bottomrule
{\bf Question: } How does CRISPR/Cas9 compare to other gene-editing technologies in terms of efficiency and precision?\\\midrule
{\bf Inspiring Paper:} Sajeesh et al. 2006.  \href{https://www.semanticscholar.org/paper/CRISPR-Cas9-gene-editing\%3A-Research-technologies\%2C-Memi-Ntokou/36d936c75cbc7bd8b32a03c868256f9b260888ba}{\textit{CRISPR/Cas9 gene-editing: Research technologies, clinical applications and ethical considerations}} \\
Gene therapy carries the potential to treat more than 10,000 human monogenic diseases and benefit an even greater number of complex polygenic conditions. The repurposing of CRISPR/Cas9, an ancient bacterial immune defense system, into a gene-editing technology has armed researchers with a revolutionary tool for gene therapy. However, as the breadth of research and clinical applications of this technology continues to expand, outstanding technical challenges and ethical considerations will need to be addressed before clinical applications become commonplace. Here, we review CRISPR/Cas9 technology and discuss its benefits and limitations in research and the clinical context, as well as ethical considerations surrounding the use of CRISPR gene editing. \\
\bottomrule
 \end{tabular}
    \caption{{\bf Scholar-Bio query examples}. The ``Inspiring Paper'' refers to the paper used to help the expert annotator formulate a question.}\label{tab:examplse_biomed_queries}
\end{table*}

\subsection{\textsc{ScholarQA-Multi}}
\label{app_sec:examples_annotated}

\begin{figure*}[h!]
\center
    \includegraphics[width=0.98\textwidth]{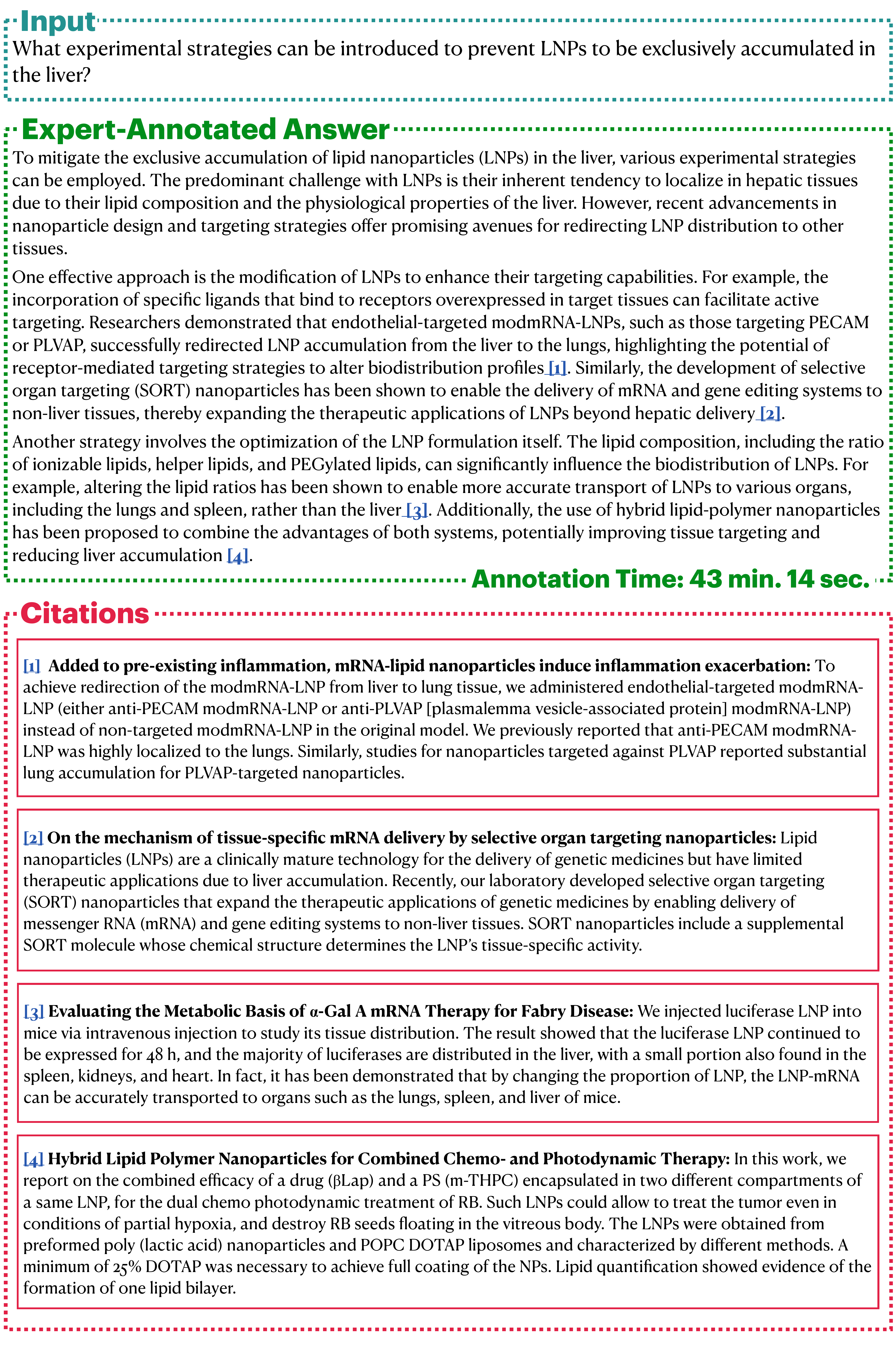}
    \caption{Examples of ScholarBench (Bio).
    }
    \label{fig:examplse_biomed_qa_1}
\end{figure*}

\begin{figure*}[h!]
\center
    \includegraphics[width=0.9\textwidth]{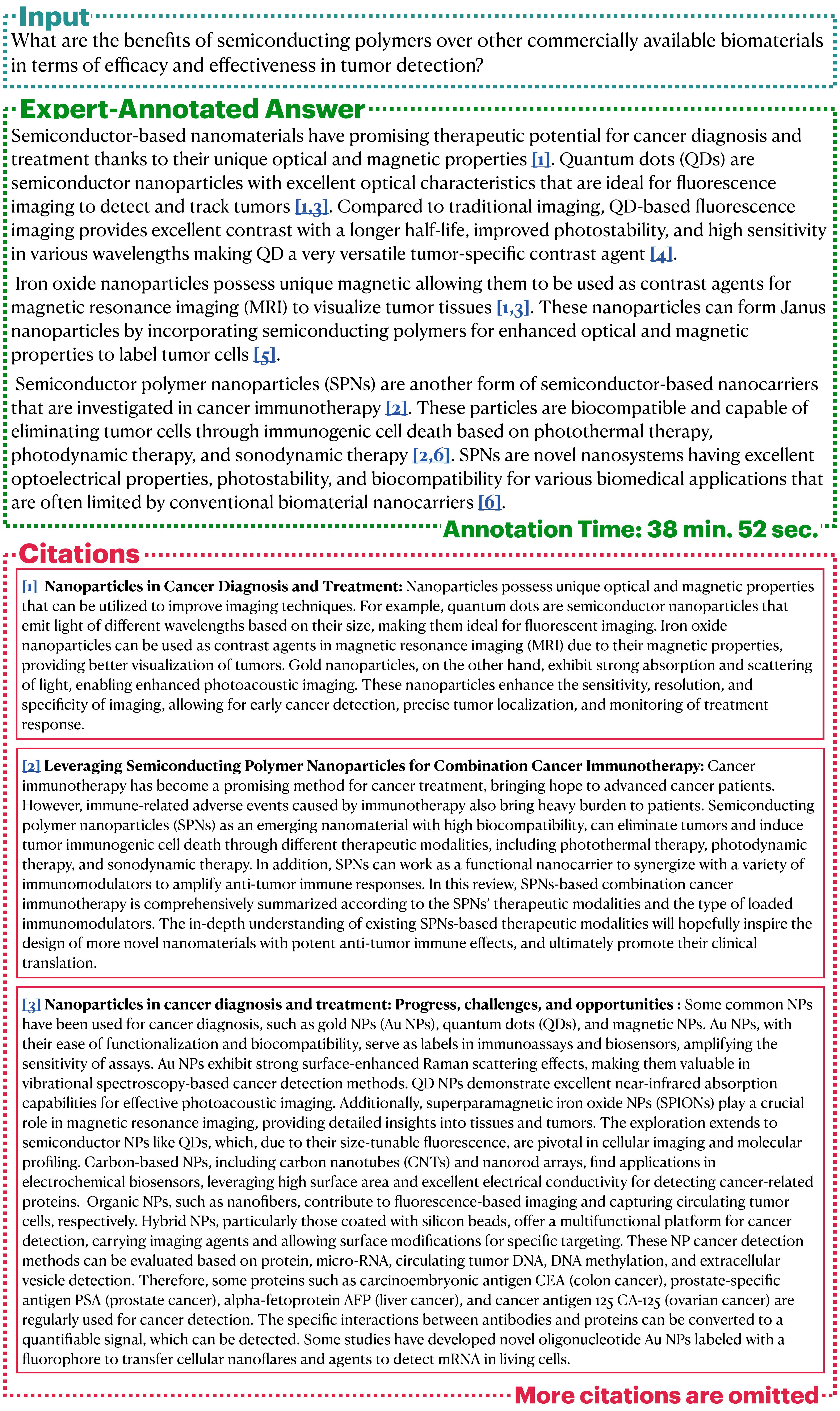}
    \caption{An example of \textsc{ScholarQA-Multi}.
    }
    \label{fig:examplse_biomed_qa_2}
\end{figure*}

\begin{figure*}[h!]
\center
    \includegraphics[width=0.9\textwidth]{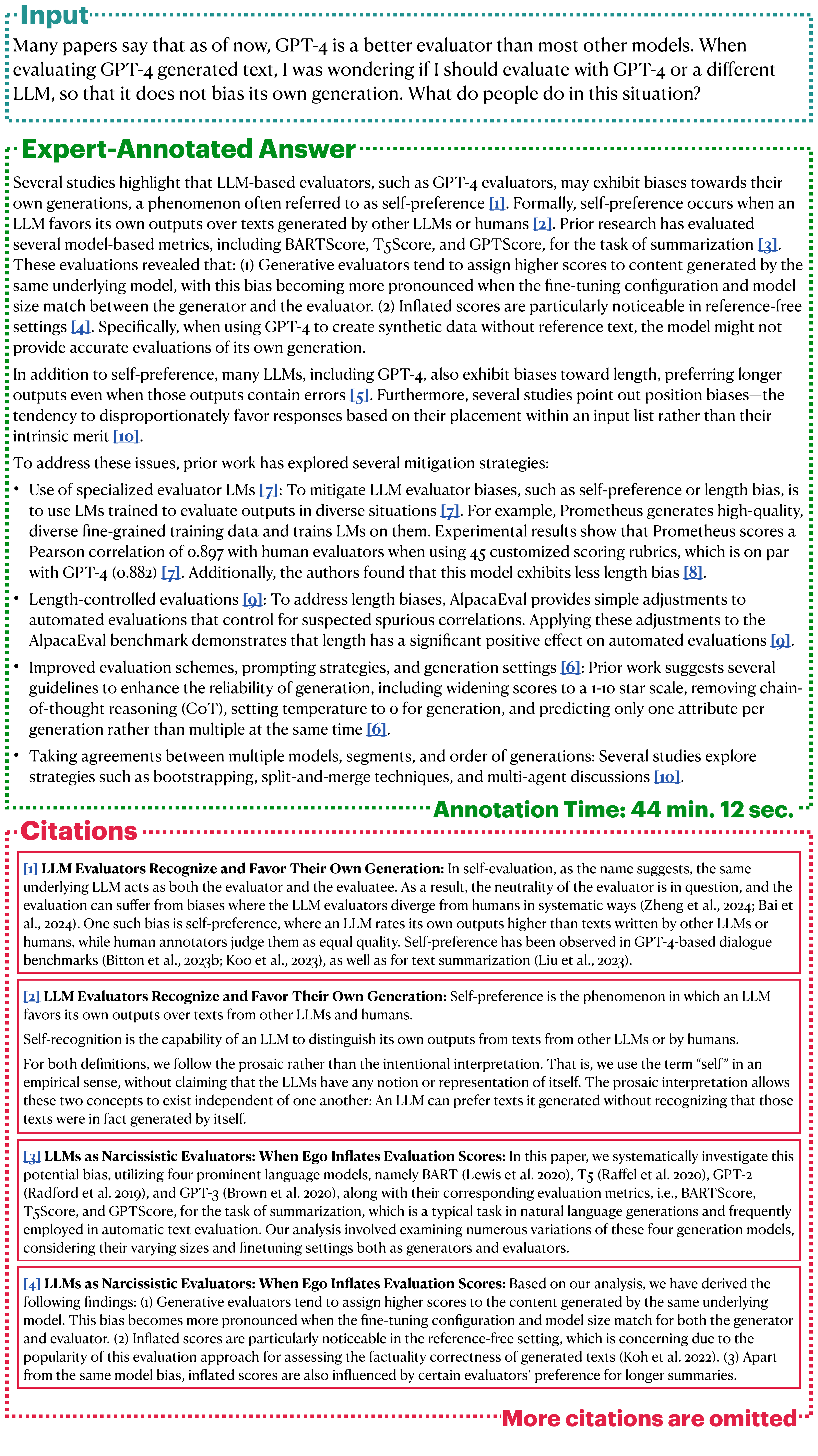}
    \caption{An example of \textsc{ScholarQA-Multi}.
    }
    \label{fig:examplse_cs_qa_1}
\end{figure*}

\begin{figure*}[h!]
\center
    \includegraphics[width=0.9\textwidth]{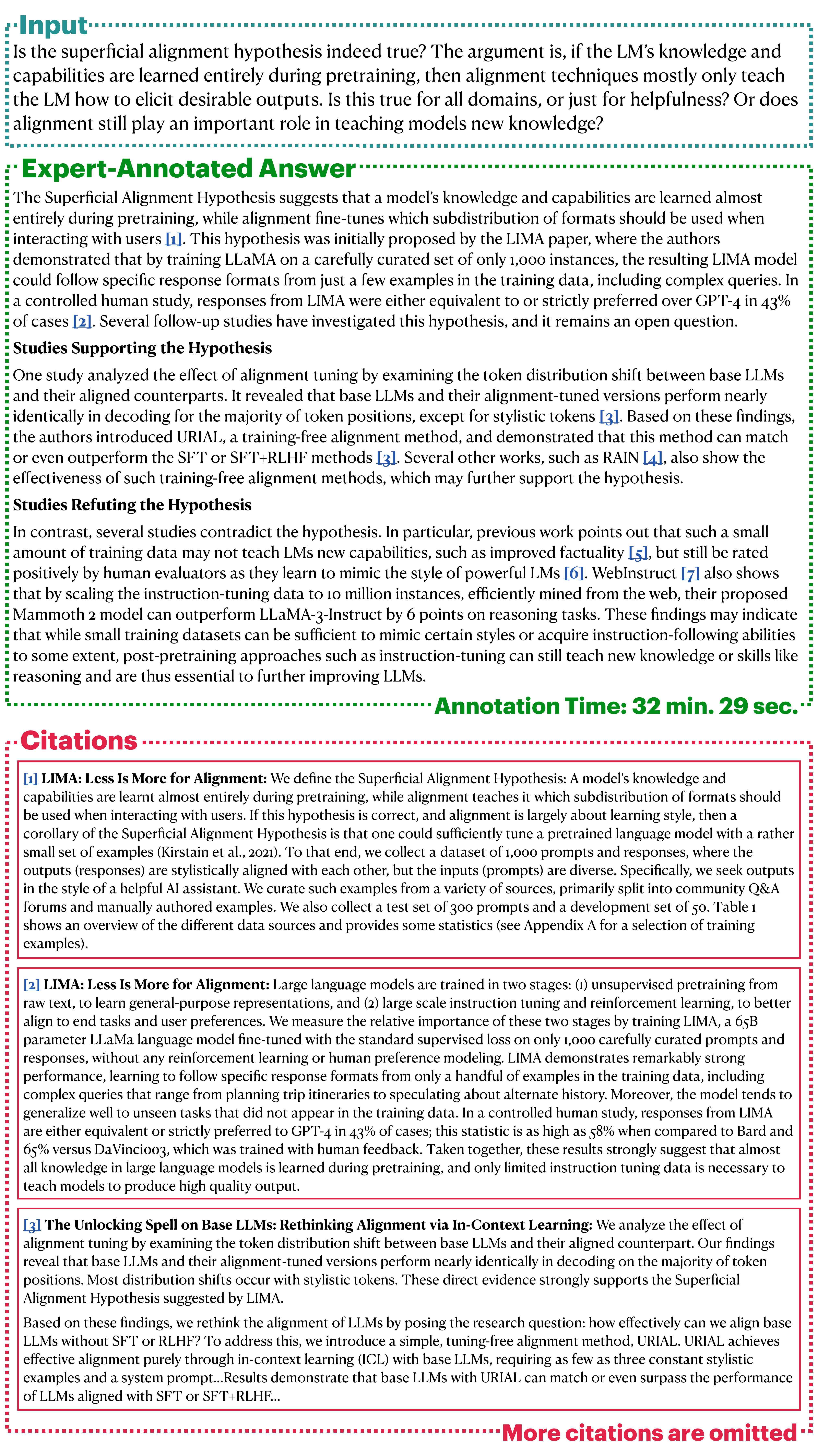}
    \caption{An example of \textsc{ScholarQA-Multi}.
    }
    \label{fig:examplse_cs_qa_2}
\end{figure*}

\begin{figure*}[h!]
\center
    \includegraphics[width=0.9\textwidth]{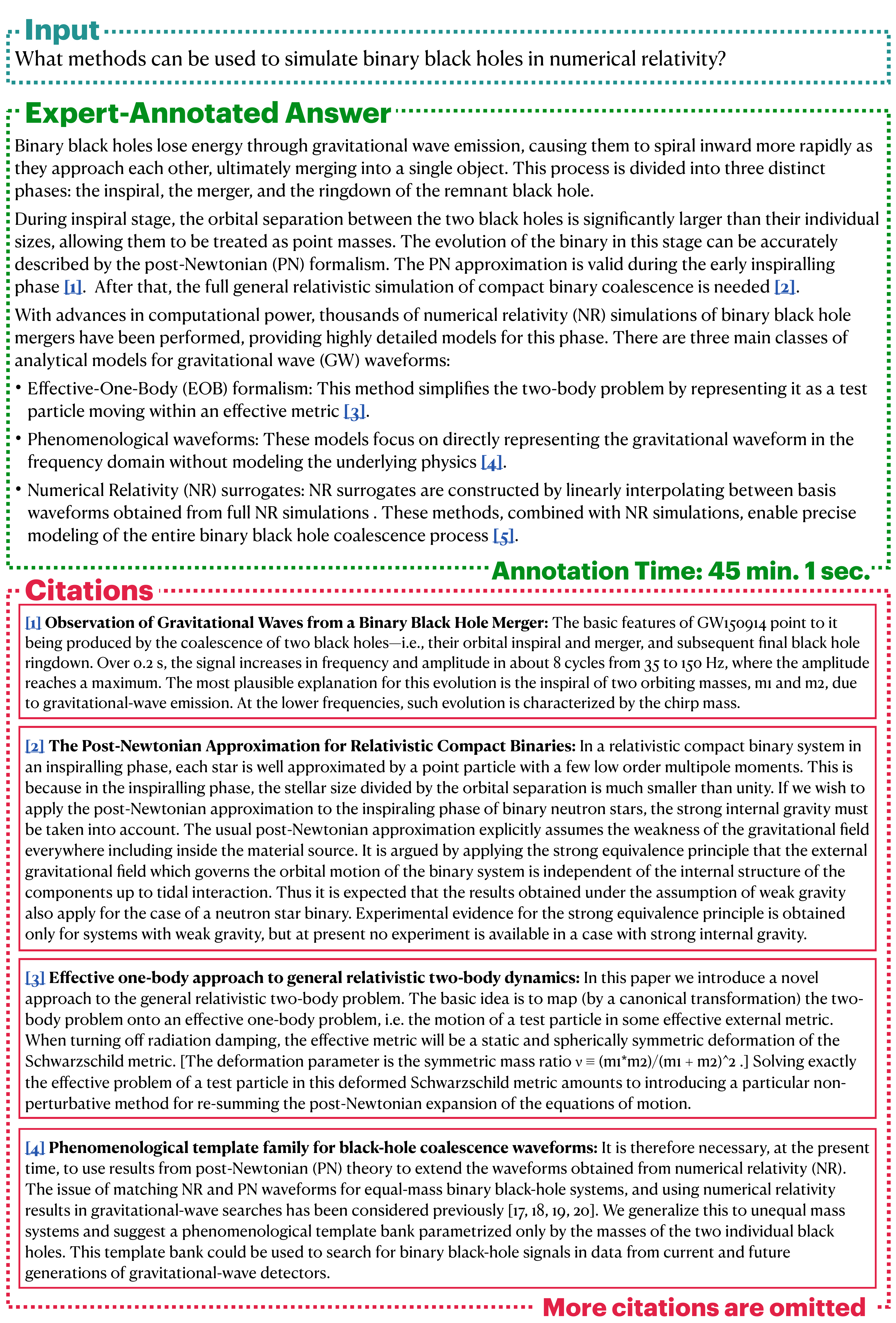}
    \caption{An example of \textsc{ScholarQA-Multi}.
    }
    \label{fig:examplse_physics_qa}
\end{figure*}

Figures~\ref{fig:examplse_biomed_qa_1}, \ref{fig:examplse_biomed_qa_2}, \ref{fig:examplse_cs_qa_1}, \ref{fig:examplse_cs_qa_2}, and \ref{fig:examplse_physics_qa} show expert-annotated examples from different domains.

\end{document}